%% file: main.tex
\documentclass[3p]{elsarticle}

\usepackage{hyperref}
\usepackage{longtable}
\usepackage{bbding}
\usepackage{amssymb}
\usepackage{amsmath}
\usepackage{float}
\usepackage{array}
\usepackage{xcolor}

\newcolumntype{P}[1]{>{\raggedright\arraybackslash}p{#1\linewidth}}

\journal{Applied Energy}

\begin{document}

\begin{frontmatter}

\title{Deep Learning in Automated Power Line Inspection: A Review}

\input{authors}

\input{abstract}

% \begin{graphicalabstract}
%     \begin{figure*}[h]
%         \centering
%         \includegraphics[width=1\linewidth]{graphical_abstract.pdf}
%     \end{figure*}
% \end{graphicalabstract}

% \begin{highlights}
%     \item Presents a comprehensive review of over 70 recent articles on automated power line inspection.
%     \item UAVs equipped with visible-light cameras are the most common platforms for data acquisition.
%     \item Other image modalities, like thermal or ultraviolet, are not widely explored.
%     \item Deep learning models like YOLO, R-CNN, and SSD, and their variants, are the most popular for component and fault detection.
%     \item Edge-cloud collaboration, multi-modal analysis, and data scarcity remain prevalent challenges.
% \end{highlights}


\end{frontmatter}

\input{sec_1_introduction}
\input{sec_2_related_works}
\input{sec_3_automated_power_line_inspection}
\input{sec_4_image_acquisition}
\input{sec_5_imaging_techniques}

\input{sec_6_datasets}
\input{sec_7_architectures}
\input{sec_8_components_detection}
\input{sec_9_fault_diagnosis}
\input{sec_10_discussion}
\input{sec_11_challenges}
\input{sec_12_conclusion}
\input{extras}

\appendix
\input{supplementary}

\bibliographystyle{elsarticle-num} 
\bibliography{references}

\end{document}

%% file: authors.tex
\author[1,2]{Md. Ahasan Atick Faisal\texorpdfstring{\corref{cor1}}{}}
\ead{atick.faisal@qu.edu.qa}

\author[2]{Imene Mecheter}
\author[2]{Yazan Qiblawey}
\author[2]{Javier Hernandez Fernandez}
\author[1]{Muhammad E. H. Chowdhury}
\author[1]{Serkan Kiranyaz}

\affiliation[1]{
    organization={Department of Electrical Engineering, Qatar University},
    city={Doha},
    postcode={2713},
    country={Qatar}
}

\affiliation[2]{
    organization={Iberdrola Innovation Middle East},
    city={Doha},
    postcode={210177},
    country={Qatar}
}

\cortext[cor1]{Corresponding Author}

%% file: abstract.tex
\begin{abstract}
In recent years, power line maintenance has seen a paradigm shift by moving towards computer vision-powered automated inspection. The utilization of an extensive collection of videos and images has become essential for maintaining the reliability, safety, and sustainability of electricity transmission. A significant focus on applying deep learning techniques for enhancing power line inspection processes has been observed in recent research. A comprehensive review of existing studies has been conducted in this paper, to aid researchers and industries in developing improved deep learning-based systems for analyzing power line data. The conventional steps of data analysis in power line inspections have been examined, and the body of current research has been systematically categorized into two main areas: the detection of components and the diagnosis of faults. A detailed summary of the diverse methods and techniques employed in these areas has been encapsulated, providing insights into their functionality and use cases. Special attention has been given to the exploration of deep learning-based methodologies for the analysis of power line inspection data, with an exposition of their fundamental principles and practical applications. Moreover, a vision for future research directions has been outlined, highlighting the need for advancements such as edge-cloud collaboration, and multi-modal analysis among others. Thus, this paper serves as a comprehensive resource for researchers delving into deep learning for power line analysis, illuminating the extent of current knowledge and the potential areas for future investigation.  
\end{abstract}

%% file: sec_1_introduction.tex
\section{Introduction}\label{sec:introduction}
The late 19th century saw Edison's light bulb and Tesla and Westinghouse's alternating current (AC) systems usher in the electrical age.  This era not only illuminated the world but also laid the foundation for the modern power delivery system, which has become a complex network of power plants, transmission lines, and distribution networks. For many decades, this complex network relied on manual inspections, which were dangerous and limited. Technological advances improved the methods, but the recent rise of computer vision and deep learning has revolutionized how power line inspections are conducted. 

A power line comprises a multitude of components, each with distinct functions, including insulators, towers, conductors, and fittings. Operating in a challenging outdoor environment, exposed to complex landforms and unpredictable weather, power line components are susceptible to frequent damage. A single faulty component, such as a conductor, or a combination of multiple damaged components, such as fittings, can trigger power outages with far-reaching consequences. These disruptions not only disrupt regional electricity supply but can also escalate to supra-regional blackouts and even catastrophic incidents, such as forest fires \cite{mitchell_power_2013}. In California, about \(10\%\) of the state's wildfires are believed to be triggered by power lines. The severity of these fires led the California Public Utilities Commission to investigate Pacific Gas \& Electric (PG\&E) power line safety practices, considering drastic measures such as breaking up the utility into smaller entities for better management and accountability \cite{noauthor_link_nodate}. The resulting economic and societal costs due to power line failures can be substantial \cite{salim_modeling_2018}. Effective power line inspection serves as the vanguard against such calamities. Its primary objective is to assess the condition of the power line components, enabling informed decisions on maintenance and replacement. A swift and accurate inspection process significantly enhances the efficiency of maintenance decision-making and, in turn, reduces the likelihood of power line failures, safeguarding the safety and reliability of the power supply to the connected load \cite{nguyen_automatic_2018}.

Nonetheless, power line inspection encounters a series of challenges, ranging from covering vast geographic areas to dealing with a diverse array of components and navigating complex natural environments. For decades, traditional inspection methods have relied on manual ground surveys and helicopter-assisted patrols \cite{liu_two_layer_2019}. These methods heavily depend on visual observations from humans, which involve significant costs, inherent risks, low operational efficiency, and long timeframes \cite{matikainen_remote_2016}. In recent years, the application of computer vision and deep learning technologies has helped in a transformative era for power line inspection \cite{yang_review_2020}. These advanced techniques have effectively decoupled the traditional inspection process into two distinct phases: data collection and data analysis. Operators can now leverage computer vision and deep learning to automatically process images and videos captured by Unmanned Aerial Vehicles (UAVs) or other means. This transition from manual labor-intensive methods to automated inspection is driven by compelling advantages, including cost-efficiency, enhanced safety, and superior operational efficiency \cite{yang_review_2020}.

Today, the maintenance of power lines is being transformed by the integration of computer vision and deep learning algorithms. These technologies enable the automation of inspection processes, offering a safer, more efficient, and cost-effective method of identifying potential issues before they escalate into major failures. By leveraging high-resolution images and real-time data analysis, utility companies can now predict maintenance needs, prevent outages, and ensure the reliable delivery of electricity to consumers. However, this transition has introduced a deluge of data. Furthermore, the conventional approach for analyzing these images and videos involves time-consuming manual efforts, which are not only costly but also fraught with potential safety hazards and may lack the necessary precision \cite{martinez_power_2018}. Consequently, there is an urgent demand for the development of automated methodologies to replace manual analysis \cite{liu_data_2020}. Over the past years, several attempts have diligently striven to devise rapid and accurate methods for the automatic evaluation of power lines from aerial imagery or land \cite{nguyen_automatic_2018}. Those attempts tried an extensive spectrum of power line components and their associated faults, primarily leveraging image processing and computer vision. Although image processing-based approaches like color \cite{reddy_condition_2013}, shape \cite{zhao_localization_2015}, or texture segmentation \cite{wu_texture_2012} have seen some success over the years, they are gradually being replaced by more advanced deep learning-based approaches. Moreover, a substantial portion of these endeavors are task-specific, concentrating on isolated components or specific fault types. 

This review paper explores the recent works on vision-based power line inspection, casting the gaze through the lens of deep learning. The paper starts with an introduction to the foundational concepts in power line inspection, encompassing various inspection methods and data sources. Subsequently, this paper provides a brief introduction to the deep learning-based techniques applied to power line inspection. Moving forward, an extensive review of research endeavors focused on the analysis of images in power line inspection has been explored. 
% {\color{blue}
The literature has been organized into two categories: component detection and fault diagnosis. Component detection research focuses on locating and identifying power line elements, either as a standalone task or as a crucial first step in fault analysis. Fault diagnosis studies, on the other hand, encompass both direct fault detection approaches and methods that build upon component detection to identify specific defects. This systematic division allows us to examine the unique challenges and solutions in each domain while highlighting their interconnected nature.
% }
The paper unveils the key features of each analytical approach, explores the nature of the datasets employed, and showcases representative quality analysis results, offering insight into the diverse capabilities of these methods across various applications. Finally, the paper presents a set of open research questions and uncharted territories awaiting future exploration. These questions encompass challenges related to data quality, the intricacies of small object detection, the application of deep learning in embedded systems, and the definition of robust evaluation baselines. Finally, this paper synthesizes the key findings and insights derived from this comprehensive analysis.

The structure of the rest of this paper is outlined as follows: Section \ref{sec:related_works} explores similar works and compares this review with the existing ones. Section \ref{sec:automated} introduces the key stages of automated power line inspection. Section \ref{sec:image_acquisition} explores various image acquisition methods and the vehicles employed for collecting power line data. In Section \ref{sec:imaging}, we examine a range of imaging techniques, discussing their advantages and limitations. Section \ref{sec:datasets} presents a concise overview of available computer vision datasets pertinent to power line inspection. Section \ref{sec:architectures} highlights the leading deep learning models and architectures employed in computer vision for tasks such as object detection and classification. Following this, Section \ref{sec:components} offers an in-depth review of research papers focused on power line component detection. Section \ref{sec:fault} scrutinizes literature dedicated to identifying power line faults through computer vision techniques. Section \ref{sec:discussion} provides a qualitative assessment of the reviewed articles and Section \ref{sec:challenges} addresses the prevalent challenges within this field and proposes potential areas for future research. The paper concludes with Section \ref{sec:conclusion}, summarizing the findings of this comprehensive review. 

%% file: sec_2_related_works.tex
\section{Related Works}\label{sec:related_works}
The application of computer vision and deep learning for power line inspection has garnered increasing attention in recent years, reflected in a growing body of survey papers. Chen et al. \cite{chen2021environment} and Yang et al. \cite{yang2020review} offer general overviews of automated power line inspection techniques, including 3D reconstruction, object detection, and inspection platforms. Their work is primarily centered on the technology aspects, such as LiDAR-based reconstruction and the classification of detection techniques. While this technical perspective is valuable, the review lacks an in-depth exploration of deep learning's impact on these technologies. Sundaram et al. \cite{sundaram2021deep} provide an overall perspective on deep learning applications in the electrical domain, touching upon power line inspection alongside other areas. However, their review is broader in scope, covering multiple electrical applications, which dilutes the focus on power line inspection. Ruszczak et al. \cite{ruszczak2023overview} focus specifically on the importance of specialized datasets for training deep learning models in power line inspection tasks. 

Several reviews delve into the use of unmanned aerial vehicles (UAVs) for inspection. Xu et al. \cite{xu2023development} provide a systematic summary of UAV platforms and image recognition techniques, while Foudeh et al. \cite{foudeh2021advanced} concentrate on UAV technologies and control strategies. Nguyen et al. \cite{nguyen_intelligent_2019, nguyen_automatic_2018} explored UAV-based power line inspection techniques in the light of deep learning and proposed a concept for an autonomous UAV-based inspection system and discuss its challenges and possibilities. The above mentioned reviews are notable for their focus on UAV-based systems, but it largely centers on the hardware and system integration aspects. Although they mention deep learning, it is not the primary focus of their analysis. Finally, Liu et al. \cite{liu_data_2020} present a comprehensive review of data analysis techniques, including deep learning methods, for power line inspection. Although the authors cover deep learning, the topic of the review is much broader due to the inclusion of image processing techniques and non machine learning-based approaches.

While these existing reviews offer valuable insights, this current review distinguishes itself through several key contributions. First, it provides a comprehensive and up-to-date analysis of deep learning applications specifically for power line inspection, encompassing both component detection and fault diagnosis. This focused exploration of deep learning techniques sets it apart from previous surveys that either covered a broader range of inspection methods or touched upon power line inspection as part of a larger survey. Second, the review systematically categorizes existing research into component detection and fault diagnosis, providing a structured understanding of the field. It summarizes various methods and techniques, offering insights into their functionality and use cases. This systematic approach facilitates easier navigation and comprehension of the current state-of-the-art. Furthermore, the review places a particular emphasis on deep learning methodologies, including their fundamental principles and practical applications in power line inspection. Finally, the review outlines future research directions, highlighting areas like data quality improvement and small object detection that require further exploration. It serves as a roadmap for future research endeavors, guiding researchers towards potential breakthroughs.

In summary, this review paper contributes significantly to the field by offering a comprehensive, systematic, and deep learning-focused analysis of automated power line inspection. It builds upon existing knowledge, provides a structured understanding of current research, and charts a course for future advancements.

%% file: sec_3_automated_power_line_inspection.tex
\section{Methodology for Automated Power Line Inspection}\label{sec:automated}
\begin{figure*}[htb]
    \centering
    \includegraphics[width=1\linewidth]{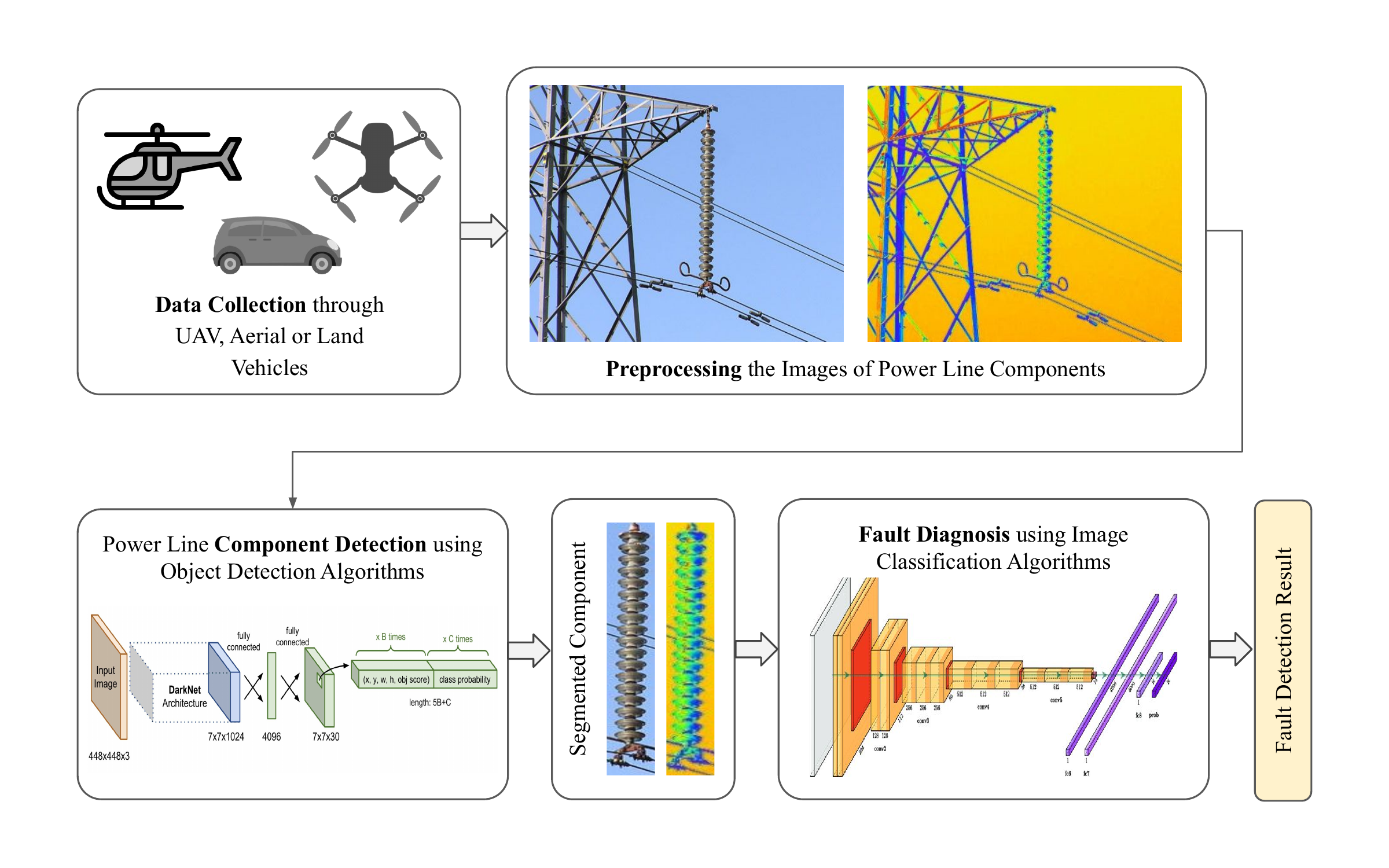}
    \caption{Block Diagram of an automated multi-modal power line inspection system.}
    \label{fig:methodology}
\end{figure*}

Automated power line inspection leverages the capabilities of computer vision and deep learning to enhance the safety, reliability, and efficiency of power transmission systems. Figure \ref{fig:methodology} shows the overview of a typical computer vision-based power line inspection process. It starts by capturing images of power line components using aerial or land vehicles sometimes using multiple imaging techniques. The images are then processed through the object detection or segmentation algorithm. Subsequently, the image classification algorithm evaluates the segmented insulators to classify them as 'good' or 'faulty', resulting in the final fault detection output.  This section provides an overview of the key steps involved in this process, from data collection to fault diagnosis. It should be noted that in many cases in the literature, the component detection part has been skipped and the deep learning model has been trained for the fault detection purpose directly. 

\subsection{Data Collection}
The basis of automated power line inspection is high-resolution images usually captured from land or aerial vehicles. The collected images provide valuable visual data that forms the basis for subsequent analysis. The data collection process for power line inspection has been discussed in more detail in Section \ref{sec:image_acquisition}.

\subsection{Preprocessing}
\begin{figure}[htb]
    \centering
    \includegraphics[width=0.5\textwidth]{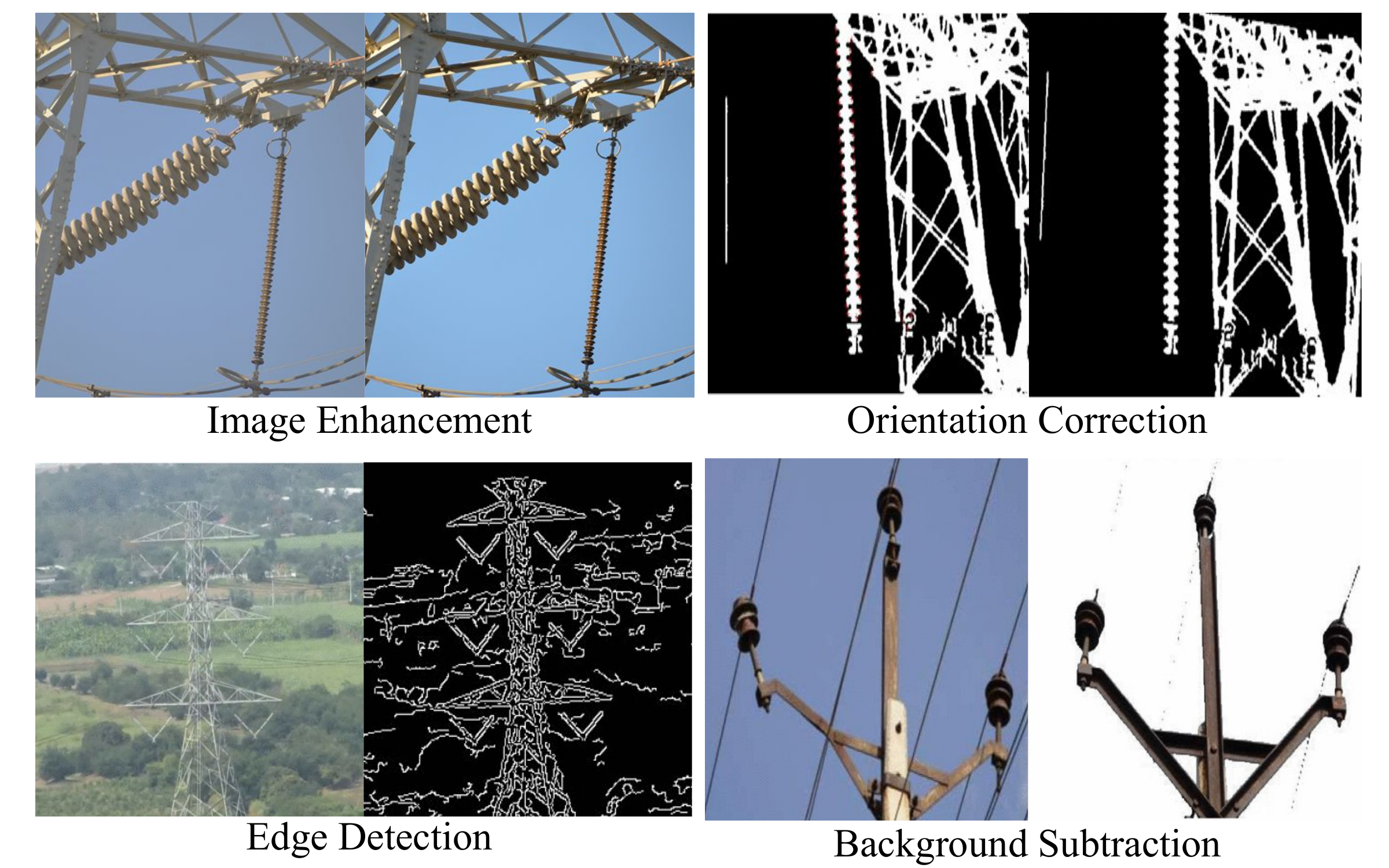}
    \caption{Example of different image enhancement techniques used for power line inspection \cite{liu_data_2020}.}
    \label{fig:image_processing}
\end{figure}

The raw images obtained during data collection often require preprocessing to improve their suitability for analysis. Preprocessing steps may include noise reduction, image enhancement, and geometric correction to account for variations in lighting conditions and perspective distortions. These enhancements ensure that the subsequent computer vision algorithms can work effectively. Figure \ref{fig:image_processing} shows visual examples of different image processing techniques such as image enhancement, edge detection, orientation correction, and color-based background subtraction. 

\subsection{Component Detection}
Component detection plays a central role in identifying and locating power line components within the power line images. Deep learning models, such as Faster R-CNN \cite{girshick_rich_2014}, (You Only Look Once) YOLO \cite{redmon_you_2016}, or (Single Shot Multibox Detector) SSD \cite{liu_ssd_2016}, are employed to detect a wide range of components, including insulators, suspension clamps, dampers, and conductors. These models can distinguish these components from the complex backgrounds typically encountered. Section \ref{sec:components} provides a comprehensive overview of the literature focusing on power line component detection.

\subsection{Fault Diagnosis}
The final phase in automated power line inspection involves fault diagnosis, where deep learning models analyze detected components or direct image inputs to identify potential issues. Deep learning models are employed to analyze the extent and severity of faults, enabling operators to prioritize maintenance and repair efforts. These models can provide valuable insights into the overall health of the power transmission system, aiding in the prevention of outages and accidents. Section \ref{sec:fault} provides a comprehensive overview of the literature focusing power line fault detection.

Automated power line inspection harnesses the capabilities of computer vision and deep learning to streamline the inspection and maintenance of power transmission systems. By automating the detection and diagnosis of components and faults, these systems contribute to safer, more reliable power delivery while minimizing downtime and operational costs.

%% file: sec_4_image_acquisition.tex
\section{Image Acquisition Platforms and Vehicles}\label{sec:image_acquisition}
The success of computer vision applications in the domain of transmission line maintenance largely depends on the quality and diversity of the images acquired. The choice of image acquisition techniques plays a vital role in determining the cost and labor associated with the operation. This section explores various methods employed for power line image acquisition.

\subsection{Unmanned Aerial Vehicles}
Unmanned Aerial Vehicles (UAVs), commonly known as drones, have emerged as a popular choice for image acquisition in power line inspection. UAVs equipped with high-resolution cameras can capture images and videos of transmission lines and their components from various angles and distances. They offer a unique combination of accessibility, maneuverability, and safety that is particularly well-suited to the challenges posed by transmission line maintenance. This versatility allows for comprehensive visual data collection, enabling the detection of defects and anomalies in power lines with greater accuracy \cite{zhang_automatic_2017}.

\subsection{Aerial Vehicles}
Aerial vehicles such as helicopters have been utilized for decades in power line inspection. Equipped with specialized cameras and imaging systems, they offer the advantage of covering long distances and providing a stable platform for capturing images of transmission lines. They are particularly suitable for inspecting high-voltage lines, where safety considerations are paramount \cite{yang_review_2020}.

\subsection{Land Vehicles}
In situations where aerial inspection may not be feasible or cost-effective, land vehicles equipped with imaging equipment are employed. These vehicles can navigate the terrain near transmission lines, capturing images and videos of the components and their surroundings. They are especially useful for inspecting power lines in areas with limited airspace accessibility.

\subsection{Fixed Camera}
Fixed cameras installed at strategic locations along power lines provide continuous monitoring. These cameras capture images at predefined intervals or when triggered by certain events. They offer a cost-effective solution for routine inspections and surveillance, although they may have limitations in terms of coverage and flexibility compared to aerial methods.

\subsection{Satellite Imaging}
Satellite imaging technology is increasingly being explored for large-scale monitoring of transmission lines. While the resolution may not be as high as that of UAVs or helicopters, satellite imagery can provide valuable data for identifying overall trends and assessing the condition of transmission networks over vast geographic areas \cite{zhou_insulator_2023}. 

% {\color{blue}
\subsection{Power Line Inspection Robots}
Power line inspection robots have emerged as a promising technology for automating the inspection and maintenance of transmission lines. These robots can be classified into climbing robots, flying robots (UAVs), and hybrid climbing-flying robots \cite{alhassan2020power}. Climbing robots can roll along the power lines and provide high-quality inspection data, but they face challenges in obstacle avoidance and deployment onto the lines. Flying robots offer faster inspection and easier obstacle avoidance but may have limitations in terms of inspection quality. Hybrid robots aim to combine the advantages of both climbing and flying robots for more effective inspection. Research in this area focuses on improving the robots' mechanical design, power systems, control algorithms, and sensing capabilities for autonomous operation \cite{chen2021environment}. For a comprehensive review of power line inspection robots, readers are referred to \cite{alhassan2020power, chen2021environment, ekren2024review}.
% }

Selecting the most suitable image acquisition method is crucial for obtaining high-quality data. The choice often depends on factors like cost, terrain, and the specific goals of the inspection. The choice of image acquisition method may also be influenced by weather conditions. Adverse weather, such as rain, snow, or fog, can impact the performance of aerial methods. Ground-based and fixed-camera systems may be preferred in such scenarios for their resilience to adverse weather conditions. Table \ref{tab:imaging_platforms} shows a general comparison between these techniques focusing on several different aspects such as cost, accuracy, coverage, and safety. 
% Figure \ref{fig:imaging_platform} shows statistics of the different imaging platforms used in the literature with UAV being the most used platform followed by fixed cameras, aerial vehicles, railway cameras and satellite.

\begin{table*}[htb]
\centering
\scriptsize
\caption{Comparison of different image acquisition techniques.}
\label{tab:imaging_platforms}
\begin{tabular}{P{0.15} P{0.05} P{0.1} P{0.07} P{0.1} P{0.07} P{0.07} P{0.15}}
\hline
Platform & Cost & Tracking* & Accuracy & Efficiency & Coverage & Safety & Usage \\
\hline
UAV & Low & Difficult & Good & Fast & Good & Safe & General Purpose \\
Land Vehicles & Low & Easy & Good & Slow & Limited & Safe & Road-side lines \\
Aerial Vehicles & High & Easy & Bad & Fast & Good & Unsafe & Hard-to-reach areas \\
Fixed Camera & High & Easy & Good & NA & Limited & Safe & For towers \\
Satellite & High & Difficult & Bad & Slow & Good & NA & Hard-to-reach areas \\
Inspection Robots & High & Easy & Good & Slow & Limited & Safe & Detailed inspection \\
\hline
\multicolumn{8}{l}{\textit{* Tracking refers to following and keeping track of the power line}}

\end{tabular}
\end{table*}

% \begin{figure}[htb]
%     \centering
%     \includegraphics[width=0.5\textwidth]{imaging_platform_updated.pdf}
%     \caption{Statistics of different imaging platforms used in literature. The figure is generated based on 64 articles. Some articles did not mention the imaging platform and hence were not included in this analysis.}
%     \label{fig:imaging_platform}
% \end{figure}

%% file: sec_5_imaging_techniques.tex
\section{Imaging Techniques}\label{sec:imaging}
The choice of imaging technique is critical for obtaining the most relevant and accurate data. Various imaging modalities are employed, each offering unique advantages and limitations. In this section, four primary image acquisition techniques: visible light images, infrared images, UV images, and x-ray images have been explored. Furthermore, LiDAR-based geospatial mapping has also been discussed briefly.

\subsection{Visible Light Imaging}
Visible light imaging is the most widely used method and is readily available through standard cameras. It is a cost-effective choice for routine inspections. Visible light images capture fine details of power line components and surrounding vegetation and other objects, which is crucial for identifying defects and assessing structural integrity. However, visibility can be compromised in adverse weather conditions, such as fog, rain, or darkness \cite{zhang_study_2023}, \cite{zhang_finet_2022}. While visible light images provide valuable surface-level data, they may not detect defects hidden within materials or components.

\subsection{Infrared Imaging (Thermography)}
Infrared imaging, or thermography, detects temperature variations in power line components, highlighting issues like overheating, loose connections, and faulty insulators \cite{singh_design_2021}. Faulty electrical components often result from internal electrical defects which can lead to over-current flow and heating of the component which can be detected from its surface temperature. Infrared imaging is not reliant on visible light and can operate effectively day and night. However, Infrared imaging requires specialized cameras that can be costly and may require trained personnel. It primarily provides information about surface temperature, and its ability to penetrate materials is limited \cite{jaffery_design_2014}.

\subsection{Ultraviolet (UV) Imaging}
UV imaging is particularly useful for detecting corona discharges - electrical discharges caused by the ionization of air surrounding high-voltage conductors - which can indicate electrical faults \cite{hu_new_2012}. UV imaging is non-destructive and can reveal hidden faults without physical contact with the power lines \cite{li_image_2019}. However, UV imaging has a limited range compared to visible light or infrared imaging, which means it may not capture an entire transmission line in a single image \cite{zang_status_2008}. Also, specialized UV cameras are required for this technique.

\subsection{X-Ray Imaging}
X-ray imaging can penetrate materials, providing detailed images of inner structures and components. It is effective at identifying hidden defects, such as corrosion and internal damage \cite{wang_internal_2023}. The use of X-ray imaging involves exposure to ionizing radiation, which requires adherence to safety protocols and limits its routine use. Also, X-ray imaging equipment is costly and requires skilled operators to acquire.

\subsection{LiDAR Imaging}
LiDAR (Light Detection and Ranging) imaging is a cutting-edge technique increasingly used in power line inspection. It employs laser light to create high-resolution, three-dimensional representations of power line infrastructure and surrounding environments. LiDAR sensors emit pulses of laser light and measure the time taken for each pulse to bounce back after hitting an object. This data is then used to construct detailed 3D models of the power lines and their immediate surroundings. LiDAR provides precise, three-dimensional information, enabling accurate mapping of power line components and detection of even minor structural anomalies. Unlike visible light imaging, LiDAR can penetrate through mild fog, rain, and other atmospheric conditions, offering more consistent results in diverse weather scenarios. It is particularly effective in assessing vegetation encroachment and potential physical obstructions near power lines, which are crucial for maintaining line safety and preventing outages \cite{guan2021uav, bergmann2024approach}.

The choice of image acquisition technique in power line inspection depends on specific inspection goals, budget, and the expected challenges. Often, a combination of imaging modalities may be used to maximize the coverage of inspection. While visible light images provide a foundational dataset for routine inspections, advanced techniques like infrared, UV, and X-ray imaging offer deeper insights into the condition of power lines. When combined with computer vision algorithms, these techniques can significantly enhance the accuracy and efficiency of fault detection and maintenance decisions \cite{li_image_2019}. Table \ref{tab:imaging_techniques} shows a comparison between these techniques in terms of cost, lighting condition, and coverage.

\begin{table*}[htb]
\scriptsize
\caption{Comparison between different imaging techniques.}
\label{tab:imaging_techniques}
\centering
\begin{tabular}{P{.1} P{.1} P{.15} P{.075} P{.425}}
\hline
Modality & Cost & Lighting & Coverage & Usage \\
\hline
Visible & Low & Requires Light & High & Structural defect, foreign object detection \\
IR & Moderate & Not   needed & High & Internal electrical defect \\
UV & Moderate & Not needed & Low & Corona discharge \\
X-Ray & High & Not needed & Low & Internal structural defect \\
LiDAR & High & Not needed & High & 3D mapping, terrain analysis, vegetation encroachment \\
\hline
\end{tabular}
\end{table*}

%% file: sec_6_datasets.tex
\section{Publicly Available Datasets}\label{sec:datasets}
Developing deep learning tools for checking power lines automatically relies on having detailed datasets with images of power line parts. These images need to be gathered and labeled by experts, which can be tricky and expensive. It often involves using UAVs or helicopters. Labeling these images is a detailed task, especially since the parts are small and spotting problems need an expert's eye. Therefore, making even a small dataset can take a lot of time and money. However, these deep-learning models need a lot of data to learn well and make accurate predictions. To deal with the lack of enough real images, some researchers have tried making artificial images by changing the background or surroundings of power line parts using computer programs \cite{tao2018detection}. The big issue is that there aren't many large datasets available for everyone to use. The datasets that are created by utility companies are not available to the public due to privacy and data protection laws and regulations. Several companies around the world are obligated to follow rules and regulations regarding data protection such
General Data Protection Regulation (GDPR) \cite{voigt2017eu}
in Europe or California Consumer Protection Act (CCPA) \cite{ccpa} in California. Those laws limit the sharing of data from Utility companies. But there are a few datasets that can be found online in Table \ref{tab:datasets}. For more information on this topic, readers are referred to a study by Ruszczak et al. \cite{ruszczak2023overview} containing a comprehensive review of the power line datasets.

\begin{table*}[htb]
\scriptsize
\caption{Summary of some publicly available power line image datasets}
\label{tab:datasets}
\begin{tabular}{P{0.3} P{0.05} P{0.05} P{0.1} P{0.125} P{0.075} P{0.05} P{0.05}}
\hline
Name & Year & Type & Component & Task & No. of Images & Ref & Download\\
\hline
Chinese Power Line Insulator Dataset (CPLID) & 2018 & RGB  & Insulator & Segmentation, Classification & 848 & \cite{tao2018detection} 
& \href{https://github.com/InsulatorData/InsulatorDataSet}{Link} \\

Powerline Dataset (Infrared-IR and Visible Light-VL & 2019 & RGB, IR  & Conductor, No Conductor & Detection & 8000 & \cite{Yetgin_2019} &  \href{https://data.mendeley.com/datasets/n6wrv4ry6v/8}{Link} \\

Transmission Towers and Power Lines (TTPLA) & 2020 & RGB & Insulator, Tower, Conductor & Segmentation & 1234 & \cite{abdelfattah2020ttpla} 
& \href{https://github.com/R3ab/ttpla_dataset}{Link} \\

Power Transmission Line Dataset & 2021 & RGB & Conductor & Classification & 1044 & \cite{t9qk_cn48_21} & \href{https://ieee-dataport.org/documents/power-transmission-line-dataset}{Link} \\

(Recognizance - 2) Power Lines Detection & 2021 & RGB & Conductor & Classification & 16078 & \cite{recognizance_2} 
& \href{https://www.kaggle.com/competitions/recognizance-2/overview}{Link} \\

STN PLAD: A Dataset for Multi-Size Power Line Assets Detection in High-Resolution UAV Images & 2021 & RGB  & Tower, Insulator, Damper & Segmentation & 2409 & \cite{9643100} 
& \href{https://ieeexplore.ieee.org/document/9643100}{Link} \\

Aerial Power Infrastructure Detection Dataset & 2023 & RGB  & Tower & Segmentation & 3956 & \cite{antonis_savva_2023_7781388} 
& \href{https://zenodo.org/records/7781388}{Link} \\

RSIn-Dataset: An UAV-Based Insulator Detection Aerial Images Dataset and Benchmark & 2023 & RGB  & Insulator & Segmentation & 1887 & \cite{drones7020125} 
& \href{https://github.com/caigouyihao/Rsin-dataset}{Link} \\
\hline
\end{tabular}
\end{table*}

%% file: sec_7_architectures.tex
\section{Deep Learning Architectures and Detection Paradigms}\label{sec:architectures}

Deep learning models have significantly advanced the field of power line inspection, offering unique advantages in terms of speed, accuracy, and the ability to handle complex scenarios \cite{sundaram_deep_2021}. Models like YOLO \cite{redmon_you_2016, redmon_yolo9000_2016, redmon_yolov3_2018, bochkovskiy_yolov4_2020, jocher_yolo_2023}, R-CNN family \cite{girshick_rich_2014, girshick_fast_2015, ren_faster_2016}, SSD \cite{liu_ssd_2016}, and transformer architectures \cite{dosovitskiy2020image, liu2021swintransformerhierarchicalvision, carion2020end} have shown remarkable performance in detecting and classifying various power line components, including insulators, dampers, pin bolts, and conductor wires \cite{sadykova2019yolo, singh_2023_interpretable, zhang_cloud_edge_2020, wei_online_2022, zhai_hybrid_2021, rong_intelligent_2021, miao_insulator_2019, nguyen_intelligent_2019, dong_improved_2023, zhang_pa_detr_2023, jain2024transfer}. These models are increasingly being deployed in hierarchical detection systems, where lightweight variants perform initial coarse screening while more sophisticated versions handle refined secondary recognition \cite{wei_online_2022}. YOLO stands out for its real-time capabilities and high frame rate \cite{li_improved_2022}, while region-based CNNs excel in precisely localizing objects within images \cite{bharati2020deep}. SSD offers a balance between speed and accuracy \cite{huang2017speed}, making it particularly suitable for edge deployment, and transformer architectures, such as ViT \cite{dosovitskiy2020image}, Swin Transformers \cite{liu2021swintransformerhierarchicalvision}, and DETRs \cite{carion2020end}, have demonstrated their effectiveness in capturing global context and handling complex scenes \cite{han2022survey, dong_improved_2023, zhang_pa_detr_2023, jain2024transfer}.

Classification algorithms, particularly those pretrained on large datasets like ImageNet \cite{5206848}, have also been employed for identifying faults and anomalies in power line images. ResNet \cite{he_2023_deep}, VGG \cite{simonyan2014very}, MobileNet \cite{howard2017mobilenets}, and EfficientNet \cite{tan2019efficientnet} have shown promising results in classifying power line components as either faulty or in good condition \cite{wei_online_2022, cao_accurate_2023, luo_ultrasmall_2023, stefenon_semi_protopnet_2022, qiu_lightweight_2023, li_improved_2022, odo_aerial_2021, li_pin_2022}. The attention mechanism \cite{vaswani2017attention} has also gained widespread popularity in recent years, enhancing the accuracy and efficiency of object detection tasks \cite{cao_accurate_2023, fan_2019_few, kong_context_2018}. 

Various computer vision tasks, such as bounding box detection, semantic segmentation, and instance segmentation have been utilized in automating the inspection of power line components. Bounding box detection is particularly useful for identifying larger components like towers, insulators, and dampers \cite{ge_birds_2022}. Semantic segmentation provides detailed component-wise masks \cite{electronics12153210, bob_semantic}, while instance segmentation excels in scenarios where components are close together or overlapping \cite{electronics12153210, bob_semantic}.

For a more detailed discussion on these deep learning models and computer vision tasks, please refer to Appendices \ref{appendix:dl_models} and \ref{appendix:cv_tasks}, respectively.

%% file: sec_8_components_detection.tex
\section{Detection of Power Line Components }\label{sec:components}
In recent years, advancements in computer vision and deep learning have revolutionized the field of power line inspection and maintenance. Power transmission lines, a critical part of our modern infrastructure, consist of various components such as insulators, conductors, fittings, and more. 
% {\color{blue}
The accurate and efficient detection of these components serves two crucial purposes in power line inspection: first, as a standalone task for inventory management and infrastructure mapping, and second, as a prerequisite step for fault diagnosis, where detected components are further analyzed for potential defects. This section focuses specifically on component detection methodologies, covering research works that either concentrate solely on locating and identifying power line components, or present novel component detection techniques that later serve as foundations for fault diagnosis systems. Section \ref{sec:fault} on fault diagnosis will then examine approaches for detecting various types of faults, both in scenarios where component detection is a preliminary step and in methods that analyze faults directly from full images.
% }

\subsection{Insulator Detection}
\begin{figure*}[htb]
    \centering
    \includegraphics[width=1\linewidth]{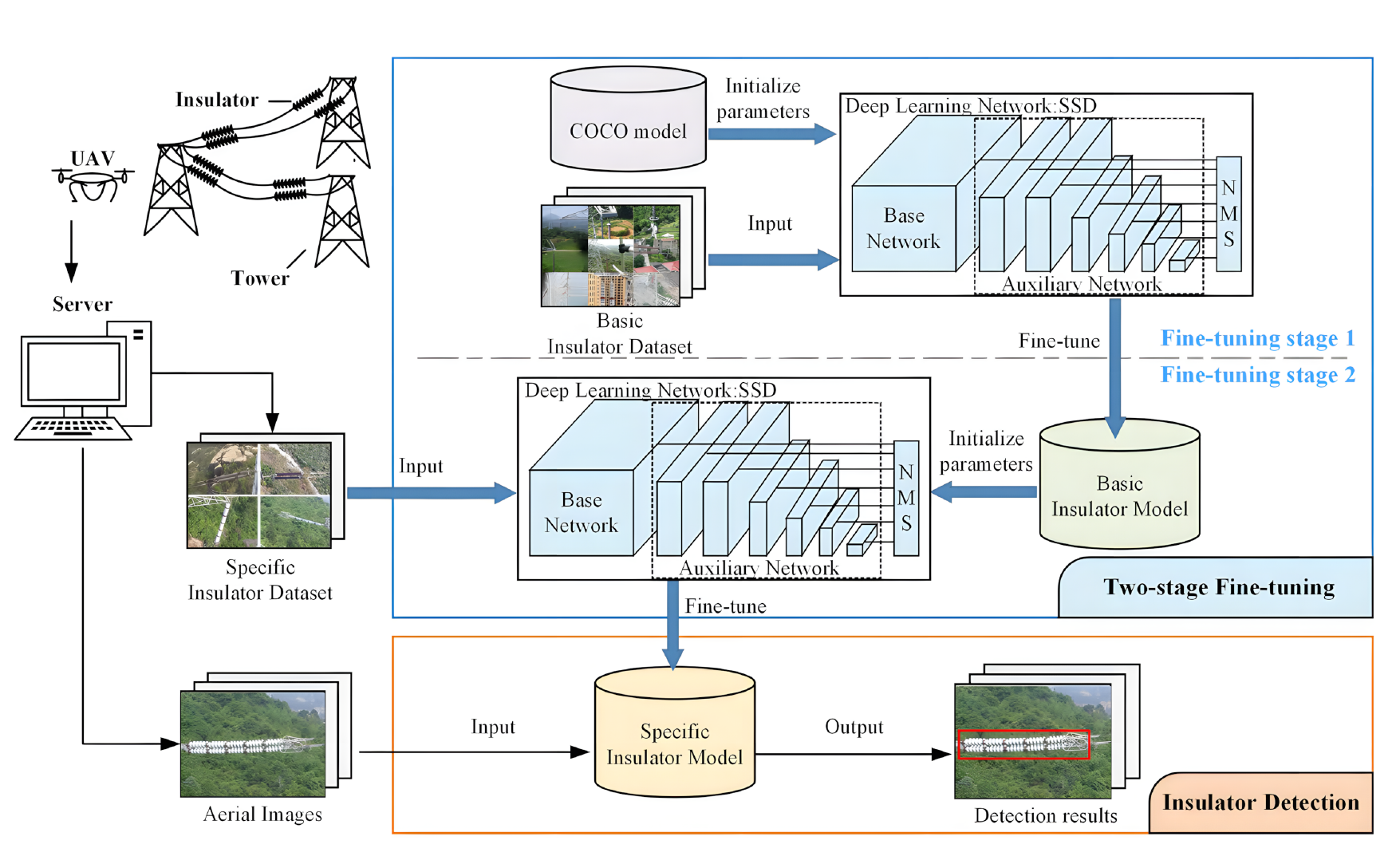}
    \caption{Overall process of the two stage insulator detection using the SSD network \cite{miao_insulator_2019}.}
    \label{fig:miao_ssd}
\end{figure*}

Among the components of transmission lines, the insulator detection technologies are the most well-researched in the literature. These technologies offer efficient and accurate solutions to identify insulators among the other components. In a 2019 work, Miao et al. \cite{miao_insulator_2019} proposed an efficient insulator detection method for aerial images, addressing the challenge of cluttered backgrounds. The approach utilizes the SSD network and employs a two-stage fine-tuning strategy to improve accuracy (Figure \ref{fig:miao_ssd}). In the first stage, a basic insulator model is fine-tuned with a diverse dataset, while the second stage fine-tunes the model for specific insulator types and scenarios. This approach demonstrates that pretraining on a generalized insulator dataset can improve the model's performance when using a smaller and more specific type of insulator dataset.

The YOLO and Faster R-CNN algorithms for object detection have been widely used for insulator detection tasks. Sadykova et al. \cite{sadykova2019yolo} proposed a cost-effective solution for automatically detecting high voltage insulators from aerial images, particularly in scenarios with uncluttered backgrounds, varying object resolutions, and lighting conditions. The approach utilizes the YOLO network and includes data augmentation to prevent overfitting, leveraging a training dataset of 56,000 image samples after augmentation. Experimental results demonstrate the accuracy of this method in locating insulators in real-time UAV-based image data. Kang et al. \cite{kang_deep_2019} proposed a two-stage deep learning system for the detection of defects on the surface of high-speed railway insulators. In the first stage, a Faster R-CNN was employed to locate the insulators within images captured by the inspection vehicle. The second stage involved a novel Deep Multitask Neural Network (DMNN) that combined a Deep Material Classifier (DMC) for segmenting the insulator from the background and a Deep Denoising Autoencoder (DDAE) for detecting anomalies (defects) on the insulator surface. The DMNN was trained in a multitask learning framework, allowing the DMC and DDAE to benefit from shared convolutional layers. The authors noted that the system might struggle to detect defects that are small or have a gray value similar to the normal insulator surface.

Any kind of occlusion such as fog, smoke, etc. can be challenging for computer-vision-based insulator detection. A 2022 work by Zhang et al. \cite{zhang_finet_2022} introduced a dataset designed for insulators and presented a benchmark model called Foggy Insulator Network (FINet), which builds upon the improved YOLOv5 framework. To enhance the dataset, the research develops and optimizes a synthetic fog algorithm, resulting in the creation of a Synthetic Foggy Insulator Dataset (SFID) containing 13,000 images. Furthermore, the YOLOv5 network is enhanced with a channel attention mechanism to form SE-YOLOv5. However, the synthetic fog generation algorithm, while effective, might not fully capture the complexity and diversity of real-world fog conditions. The evaluation of the FINet model was primarily focused on the SFID dataset, which might limit the generalizability of the results to other datasets or real-world scenarios. Table \ref{tab:insulator_detection} summarizes the relevant literature on insulator detection.

\begin{table*}[htb]
% {\color{blue}
\scriptsize
\caption{Summary of insulator detection studies in power line inspection.} \label{tab:insulator_detection} 
\begin{tabular}{P{.05} P{.1} P{.15} P{.1} P{.15} P{.1} P{.15}}

\hline
Year \& Ref & Component & Type of Detection & Imaging Platform & Dataset & Algorithm & Performance \\
\hline
2019 \cite{miao_insulator_2019} & Porcelain Insulator, Composite Insulator & Bounding Box Detection & UAV & 7605 RGB Images & SSD & Porcelain Insulator: 90.51-94.12\% Composite Insulator: 86.70-87.29\% \\

2019 \cite{sadykova2019yolo} & Glass Insulator & Bounding Box Detection & UAV & 3500 RGB Images  & YOLOv2 & Detection Accuracy: 88\% Prediction Time: 0.04s \\

2019 \cite{kang_deep_2019} & Insulator & Bounding Box Detection & High-Speed Railway & 18000 RGB Images  & F-RCNN & mAP: 99.8\% \\

2020 \cite{zhao2020image} & Insulator & Object Presence Detection & NA & 4780 Insulator and 13012 Background IR Images  & DCNN \& VLAD Coding & Detection Accuracy: 99.21\% \\

2021 \cite{zhang_defgan_2021} & Insulator & Semantic Segmentation & High-Speed Railway & 800 RGB Images  & CDSNets & IOU: 0.94 \\

2021 \cite{waleed_drone_based_2021} & Ceramic Insulator & Bounding Box Detection & UAV & 2973 RGB Images & R-CNN, SSD & Onshore mAP: 0.56-0.77; Onboard mAP: 0.24-0.27 \\

2022 \cite{wei_online_2022} & Insulator & Bounding Box Detection & NA & 8500 Insulator Images & SSD, F-RCNN  & Accuracy: SSD: 89\%, F-RCNN: 91.6\% \\

2022 \cite{zhang_finet_2022} & Insulator & Bounding Box Detection & Synthetic & 13000 Foggy RGB Images & Improved YOLOv5 & mAP@0.5:0.95: 88.3\%, F1 Score: 96.2\%  \\

2022 \cite{antwi_bekoe_deep_2022} & Insulator & Instance Segmentation & UAV & 1523 RGB Images & Mask R-CNN & mAP: 87.0\% \\

2023 \cite{zhou_insulator_2023} & Insulator & Semantic Segmentation & Satellite & 9900 RGB Images & HRNet and OHEM & F1 Score: 0.7952 \\

2023 \cite{shuang_rsin_dataset_2023} & Insulator & Bounding Box Detection & UAV & 1887 RGB Images & YOLOv4++ & mAP: 94.24\% \\

2023 \cite{singh_2023_interpretable} & Insulator & Bounding Box Detection & UAV \& Synthetic & CPLID Dataset: 848 RGB Images & YOLOv8 & mAP@[0.5:0.95]: 91.25\%   \\
\hline
\end{tabular}
% }
\end{table*}

\subsection{Detection of Power Line Fittings }
Detecting power line fittings, including components like pin bolts, dampers, and suspension clamps, presents a unique set of challenges. These fittings are relatively small in size compared to the overall transmission line structure, making their detection and classification a particularly intricate task \cite{luo_ultrasmall_2023}. This section explores the efforts and methodologies aimed at the automatic detection of these critical power line fittings, focusing on the utilization of deep learning and computer vision techniques to address the intricacies associated with their small-scale identification. Table \ref{tab:fittings_detection} summarizes the literature on power line fittings detection.

As mentioned earlier, one of the biggest challenges to overcome in power line fittings like bolts detection is their relatively small size compared to the other components. To tackle this, Luo et al. \cite{luo_ultrasmall_2023} propose a novel model (UBDDM) based on deep CNN. This model includes two modules: the ultrasmall object perception module (UOPM) for initial bolt region identification and the local bolt detection module  (LBDM) for pinpointing defects within high-resolution image blocks. The authors enhance feature extraction using ResNet-50, a hybrid attention mechanism, and multiscale feature fusion. Their method simplifies data labeling requirements while maintaining end-to-end detection capabilities. While experiments demonstrate the model's superior performance in detecting bolt defects within inspection images, the model can only draw bounding boxes of some predefined dimensions.

Transmission line fittings encapsulate a wide range of components and often require a multi-task object detection model to detect all the different types. In 2022 a study done by Zhai et al. \cite{zhai_multi_fitting_2022} proposed the Cascade Reasoning Graph Network (CRGN) which offers a novel solution to the intricate challenges posed by detecting multiple power line fittings in aerial images. CRGN uses spatial knowledge representations that capture the interrelationships among objects based on the unique characteristics of transmission line fittings. However, the model's dependency on high-resolution, close-up photographs of the fittings presents a practical challenge, as such image quality may be difficult to consistently achieve in real-world scenarios.

\begin{table*}[htb]
% {\color{blue}
\scriptsize
\caption{Summary of power line fittings detection studies.} \label{tab:fittings_detection}
\begin{tabular}{P{.05} P{.1} P{.15} P{.1} P{.1} P{.15} P{.15}}
\hline
Year \& Ref & Component & Type of Detection & Imaging Platform & Dataset & Algorithm & Performance \\
\hline
2021 \cite{zhai_hybrid_2021} & Fittings & Bounding Box Detection & UAV & 1455 RGB Images & HK R-CNN & mAP: 59.82\% - 98.27\%  \\

2022 \cite{zhang_attention_guided_2022} & Dampers and Suspension Clamps & Bounding Box Detection & UAV & 1209 RGB Images & AGMNet & mAP: 95.3\% \\

2022 \cite{zhai_multi_fitting_2022} & Fittings & Bounding Box Detection & Aerial Vehicle & 1455 Aerial Images & CRGN & mAP@0.5-0.95: 47.5\% \\

2023 \cite{huang_structural_2023} & Damper & Semantic Segmentation & UAV & 240 RGB Images & Improved GrabCut & F1 Score: 89.1-97.3\% \\
\

2023 \cite{luo_ultrasmall_2023} & Bolts & Bounding Box Detection & UAV & 1852 RGB Images & UPOM (Based on ResNet and Attention) & Recall 0.94 - 1.00 \\
\hline
\end{tabular}
% }
\end{table*}

\subsection{Conductor Detection}
Conductor detection using computer vision must contend with multiple complexities, including the conductor's slender profile, varying backgrounds, and potential occlusions. This subsection explores the state-of-the-art techniques and advancements in conductor detection using computer vision and deep learning, addressing the unique challenges associated with this critical task. Table \ref{tab:conductor_detection} summarizes the literature on conductor detection.

Power lines are often a small portion of the image, leading to class imbalance issues. To address these challenges, Yang et al. \cite{yang_vision_based_2022} proposed a novel vision-based power line detection network designed to address challenges in detecting power transmission lines in complex aerial images including varying background environments, illumination conditions, and foreground-background class imbalance, where power lines occupy a small portion of the image. The proposed network utilizes an encoder-decoder architecture (similar to UNet) to create an end-to-end power line detection system. To improve segmentation precision, it incorporates an attention block to capture global contexts and emphasize target power line regions. Additionally, an attention fusion block is introduced to enhance multi-scale feature fusion and capture richer information, mitigating issues related to local contextual feature processing and information loss caused by multiple pooling operations. However, due to the complexity of the UNet-based network combined with the attention mechanism, the proposed model is not particularly suitable for real-time applications.

While deep learning, especially U-Net and its variants, has advanced pixel-level object segmentation, limitations in processing local contextual features and information loss in deep CNNs persist. To overcome these issues, in a follow-up study, Yang et al. \cite{yang_dra_net_2023} proposed a novel dual-branch residual attention network called DRA-Net. It features a dual-branch encoder with a residual CNN (RCNN) branch and a recurrent RCNN (RRCNN) branch to capture richer semantic information. Additionally, a U-shaped noise denoising (UND) block reduces background interference, and an edge enhancement block (EEB) strengthens the network's capacity to extract useful edge features. Experimental results demonstrate DRA-Net's excellent segmentation performance, achieving a Dice coefficient of 93.26\% and a mean Intersection over Union (mIoU) of 93.19\% on the public power line dataset (PLD) and 96.40\% Dice coefficient and 96.04\% mIoU on the self-built overhead power line (OPL) dataset. However, the reported results were obtained from a relatively small training dataset of 1144 images and could be reduced when tested on a wide variety of real-world images.

\begin{table*}[htb]
% {\color{blue}
\scriptsize
\caption{Summary of power line conductor detection studies.} \label{tab:conductor_detection}
\begin{tabular}{P{.05} P{.1} P{.15} P{.1} P{.15} P{.1} P{.15}}
\hline
Year \& Ref & Component & Type of Detection & Imaging Platform & Dataset & Algorithm & Performance \\
\hline
2022 \cite{yang_vision_based_2022} & Conductor & Semantic Segmentation & UAV & 366 RGB Images  & UNet with Attention Blocks & Dice: 0.957 \\

2023 \cite{yang_dra_net_2023} & Conductor & Semantic Segmentation & UAV & PLD Dataset: 573 RGB Images; OPL Dataset: 571 RGB Images & DRA-Net  & mIOU: 93.19\% (PLD Dataset), mIOU: 96.04\% (OPL Dataset) \\
\hline
\end{tabular}
% }
\end{table*}

\subsection{Multi-Component Detection}
A typical transmission tower consists of multiple different types of components with varying shapes and dimensions. Some studies have extended focus beyond individual component detection to multi-component detection \cite{zhang_multi_scale_2020, wang_image_2019, nguyen_intelligent_2019}. In a 2024 study, Dong et al. \cite{dong2024transmission} proposed a meta learning-based model to address the challenge of detecting key components and defects in transmission lines using aerial images, particularly when dealing with limited sample sizes for certain categories. The model incorporated a region-aware fusion (RAF) module to capture spatial relationships between support and query images, enabling effective matching and identification of objects. Additionally, cascade RAF heads were employed to progressively refine bounding box proposals and improve detection accuracy. The model was trained using a two-stage fine-tuning strategy, leveraging a larger dataset of base classes to enhance the detection of novel classes with fewer samples. Experimental results demonstrated the model's superior performance compared to traditional deep learning and few-shot object detection models. The authors acknowledged potential limitations in their work. The model's performance might be affected by variations in image quality and environmental conditions encountered in real-world power transmission line inspections. The reliance on a pre-defined set of base classes and novel classes might limit the model's adaptability to new or unexpected object categories. The authors suggested exploring the incorporation of online learning or active learning techniques to enable the model to continuously learn and adapt to new scenarios.

Self-supervised pretraining has been used effectively in other domains to tackle the challenge of minimal or non-existent annotated dataset. Liu et al. \cite{liu2023tower} introduced Tower Masking MIM (TM-MIM), a self-supervised pretraining method designed to enhance power line component detection in aerial images, particularly in scenarios with limited labeled data. By employing a novel masking strategy that focuses on the tower-conductor region and a Siamese architecture with dual reconstruction branches, the model learns to capture discriminative features and global representations from unlabeled data. The incorporation of knowledge distillation further enhances the model's ability to retain general knowledge from pretrained models while acquiring domain-specific knowledge. The authors identified potential areas for future work. The masking strategy's reliance on tower presence in images might necessitate the development of techniques to handle images without towers. Additionally, exploring the use of varied-grained masks for different feature hierarchies could further enhance the model's ability to detect components of varying scales. The authors also suggested extending the TM-MIM approach to other object detection frameworks, such as YOLO, to broaden its applicability. Table \ref{tab:multicomponent_detection} summarizes the literature on power line multi-component detection.

\begin{table*}[htb]
% {\color{blue}
\scriptsize
\caption{Summary of multi-component detection studies in power line inspection.} \label{tab:multicomponent_detection} 
\begin{tabular}{P{.05} P{.1} P{.15} P{.1} P{.15} P{.1} P{.15}}

\hline
Year \& Ref & Component & Type of Detection & Imaging Platform & Dataset & Algorithm & Performance \\
\hline
2019 \cite{chen_research_2019} & Insulator and Damper & Bounding Box Detection & UAV & 4416 Insulator \& 4352 Damper Images & YOLOv3 & Accuracy: 95.84\% \\

2019 \cite{wang_image_2019} & Tower Components & Bounding Box Detection & NA & 11600 RGB Images & F-RCNN & - \\

2019 \cite{nguyen_intelligent_2019} & Tower Components & Bounding Box Detection & Aerial Vehicle & 123151 RGB Images & SSD & mAP: 0.67 \\

2020 \cite{zhang_multi_scale_2020} & Tower Components & Bounding Box Detection & UAV & City A: 2016 RGB Images; City B: 3960 RGB Images & Enhanced F-RCNN & City A Dataset: mAP: 52.9\%, City B Dataset: mAP: 45.3\% \\

2021 \cite{odo_aerial_2021} & Insulator and Bolts & Bounding Box Detection & Aerial Vehicle & 1830 RGB Images  & Mask R-CNN and RetinaNet & Precision: 96.7\% (Insulators), 97.9\% (Bolts) \\

2023 \cite{liu2023tower} & Tower Components & Bounding Box Detection & UAV & PLCD Dataset: 1000 Images & TM-MIM Network & mAP@0.5: 87.7\% \\

2023 \cite{shi2024lskf} & Towers & Bounding Box Detection & Satellite & Duke University Dataset: 2740 Images & Improved YOLO-based Network & mAP@0.5: 77.47\% \\

2024 \cite{dong2024transmission} & Tower Components & Bounding Box Detection & UAV & 9017 RGB Images & Meta Learning & mAP@0.5: 64.6\% \\
\hline

\end{tabular}
% }
\end{table*}

% {\color{blue}
This section has demonstrated the remarkable progress in automating power line component detection through computer vision and deep learning techniques. The reviewed literature showcases significant achievements across various components: from insulator detection achieving mAP rates of up to 91.25\% using YOLOv8, to conductor segmentation reaching Dice coefficients of 96.40\% with innovative architectures like DRA-Net, and multi-component detection systems attaining mAP rates of 87.7\% through self-supervised learning approaches. These advances have been enabled by architectural innovations such as attention mechanisms, multi-scale feature fusion, and hybrid networks combining CNN backbones with specialized modules. However, several domain-specific challenges persist. The detection of small-scale components like pin-bolts and dampers, which often occupy minimal pixels in aerial imagery, remains particularly challenging, with performance metrics for fitting detection typically lower than those for larger components. Additionally, real-world deployment faces obstacles such as varying imaging conditions, complex backgrounds, and class imbalance issues where critical components occupy only a small portion of the image. The limited size and diversity of publicly available datasets—with most studies utilizing fewer than 5,000 images—continues to constrain model development and benchmarking. Recent promising directions include meta-learning approaches for few-shot detection, self-supervised pretraining methods like TM-MIM for leveraging unlabeled data, and specialized architectures incorporating domain knowledge about component spatial relationships. These challenges and emerging solutions in power line component detection will be explored more comprehensively in Section \ref{sec:challenges}, along with their implications for future research and practical applications.
% }

%% file: sec_9_fault_diagnosis.tex
\section{Power Line Fault Diagnosis}\label{sec:fault}
\begin{figure*}[htb]
    \centering
    \includegraphics[width=1\linewidth]{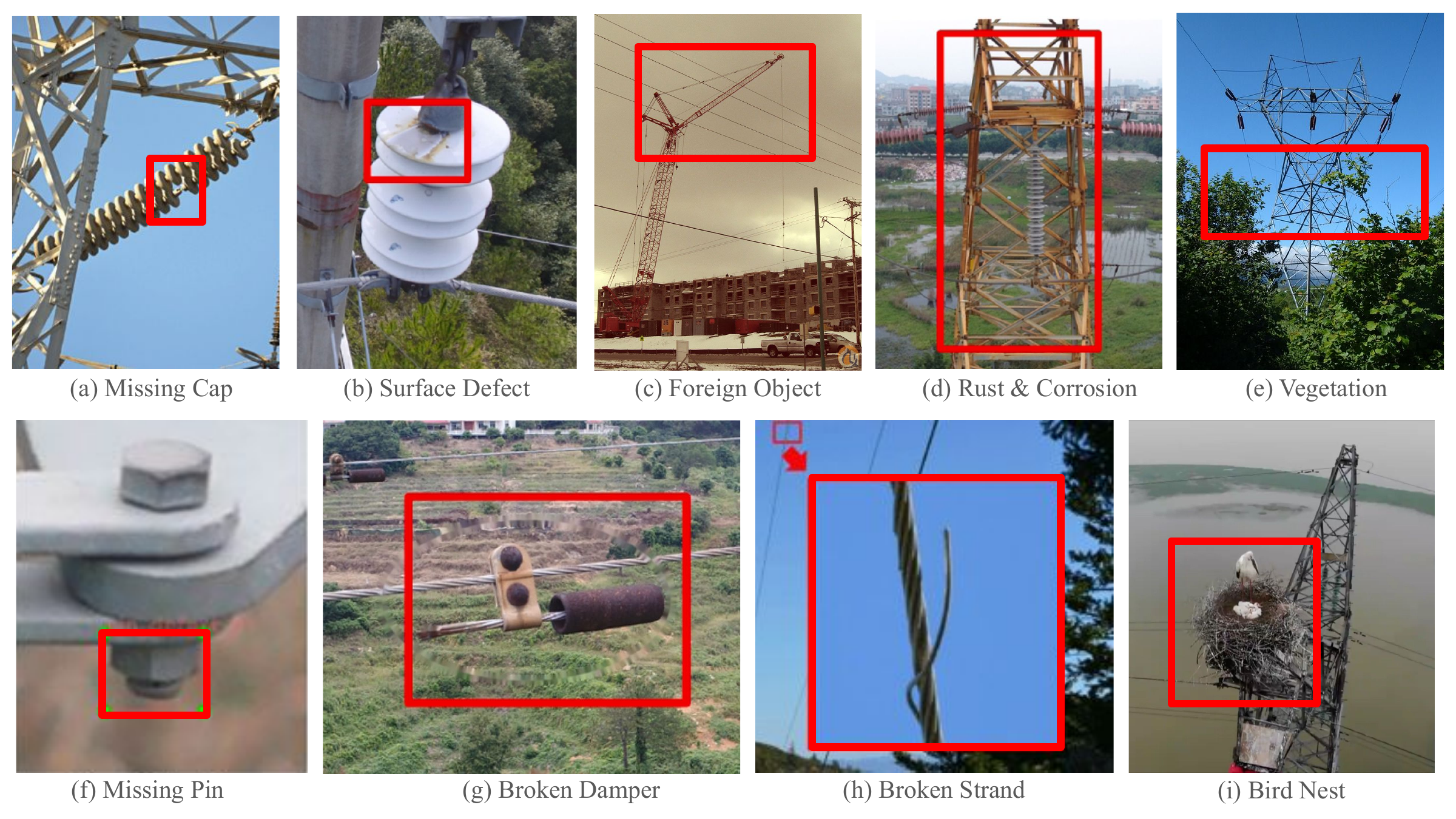}
    \caption{Different types of power line faults \cite{liu_data_2020}.}
    \label{fig:different_faults}
\end{figure*}

Visual inspection, coupled with cutting-edge computer vision techniques has emerged as a powerful and efficient means of identifying and diagnosing a spectrum of faults. From insulator defects to conductor issues, and tower anomalies to grounding problems, the ability of computer vision to meticulously assess power line components is transforming maintenance and reliability in the electrical grid. This section delves into the application of computer vision and deep learning for the diagnosis of various faults (Figure \ref{fig:different_faults}) within power lines and associated equipment, underlining its potential to enhance the resilience and performance of critical power transmission infrastructure.

\subsection{Insulator Fault Detection}
Insulators are critical components of power lines, ensuring that high-voltage currents are safely transported without coming into contact with the towers or poles that support them. However, insulators continuously endure environmental stresses and mechanical wear, which can lead to various types of faults. The following sections discuss some of the most common types of insulator faults and review significant research addressing these issues.

\begin{table*}[htb]
% {\color{blue}
\scriptsize
\caption{Summary of insulator surface fault detection studies.}
\label{tab:insulator_surface_faults} 
\begin{tabular}{P{.05} P{.1} P{.1} P{.1} P{.1} P{.15} P{.1} P{.1}}  

\hline  
Year \& Ref & Component & Type of Detection & Type of Fault & Imaging Platform & Dataset & Algorithm & Performance \\ 
\hline 

2017 \cite{liu_discrimination_2017} & Porcelain Insulators & Classification & Deterioration Damage & Still Camera & 700 IR Images & CNN & Accuracy: 93\% \\

2019 \cite{kang_deep_2019} & Insulators & Classification & Surface Defects & High Speed Railway & 18000 RGB Images & DDAE (Based on CNN AutoEncoder) & F1-Score: 0.95 \\

2019 \cite{sadykova2019yolo} & Glass Insulator & Classification & Surface Defects & UAV & 3500 RGB Images & CNN & Accuracy: 87\% \\

2020 \cite{ibrahim_application_2020} & Insulators & Classification & 3 Types of Surface Defects & Still Camera & 1240 RGB Images & CNN & Accuracy: 89.5\% \\

2020 \cite{mussina_multi_modal_2020} & Glass Insulators & Classification & 4 Types of Surface Defects & UAV & 5000 RGB Images & FCN & Accuracy: 99.76\% \\

2021 \cite{waleed_drone_based_2021} & Ceramic Insulators & Bounding Box Detection & Insulator Surface \& Structural Defects & UAV & 2973 RGB Insulator Images & R-CNN, SVM, CNN, SSD Fusion & Onshore mAP: 0.56-0.77; Onboard mAP: 0.24-0.27 \\

2022 \cite{stefenon_classification_2022} & Insulators & Classification & Kaolin Defects & Still Camera & 600 RGB Images & ANN & Accuracy: 97.50\% \\

2023 \cite{roy_accurate_2023} & Insulators & Classification & 4 Types of Surface Defects & Still Camera & 1000 RGB Images & CNN & Accuracy: 97.5\% \\
\hline

\end{tabular}
% }
\end{table*}

\subsubsection{Surface Defect Detection}
Surface defects in insulators encompass a range of anomalies that affect the outermost layer of the insulator. This category includes issues such as surface contamination, cracking, flashover marks, arcing damage, and chipping. Recent research works have made significant strides in the detection and classification of these surface defects, utilizing a variety of computer vision techniques and machine learning algorithms. Roy et al. \cite{roy_accurate_2023} developed a deep learning framework incorporating AlexNet, VGG16, and ResNet50 models to detect sand, ash, and mud contaminations on the insulator surface. However, this approach requires images of the insulators taken at close proximity and is limited to classification and cannot detect contaminated areas. Mussina et al. \cite{mussina_multi_modal_2020} introduced a Fusion Convolutional Network (FCN) for the real-time monitoring of insulators using UAVs, addressing challenges like varying resolutions and unclear surfaces. It combines a CNN with a binary ANN to form a multi-modal information fusion system that improves contamination classification accuracy on insulators to 99.76\% by using image data and leakage current values. However, the proposed model was trained on a relatively small dataset of only 250 images per class and it requires close-up images of the insulator with uniform background. These limitations can affect the model's capability to generalize on real-world situations. Table \ref{tab:insulator_surface_faults} summarizes the literature on insulator surface faults.

\subsubsection{Structural Defect Detection}
Structural defects refer to issues that affect the internal composition and mechanical strength of the insulator. They include complete breakage, missing insulator caps, and material degradation. Table \ref{tab:insulator_structural_faults} summarizes the literature on insulator structural faults. The following goes over some of the notable works on structural defect detection.
 
In a 2022 work, \cite{antwi_bekoe_deep_2022} proposed an attention-based end-to-end framework that combines object detection and instance segmentation at the pixel level. Although trained on a relatively small dataset of 1523 images, this is one of the few works that targeted pixel-level segmentation and achieved great results. By integrating a three-branch attention structure into the backbone network, the proposed model achieved a significant improvement in detection performance, surpassing the state-of-the-art instance mask prediction models while maintaining computational efficiency. In a similar work by Wang et al. \cite{wang_detection_2020}, the authors proposed an insulator defect detection method that leverages an improved ResNeSt network and an enhanced Region Proposal Network (RPN). The improvements to ResNeSt were aimed at refining the detection of insulator defects, particularly in aerial images where insulators might be small and have low resolution. The authors acknowledge that their method might struggle to detect certain types of defects, such as breaks on specific insulator types, due to their subtle visual characteristics. The computational efficiency of the proposed method, while improved compared to some baseline models, might still be a concern for real-time applications on resource-constrained devices.

Multi-task networks can be used to detect different types of insulator faults as separate classes. In a recent article Fu et al. \cite{fu_small_sized_2023} presented the I2D-Net, a deep learning-based method for detecting small-sized defects in overhead transmission line insulators, particularly missing caps. The I2D-Net enhanced the Faster R-CNN object detection framework with three key components: the Three-path Feature Fusion Network (TFFN) to improve feature extraction from shallow layers, the enhanced Receptive Field Attention (RFA+) block to adapt to different-scale defects, and the Context Perception Module (CPM) to enhance defect localization. The authors acknowledged that while the I2D-Net achieved high accuracy, it came with a slight increase in inference time compared to the baseline Faster R-CNN + FPN model. Liu et al. \cite{liu_box_point_2021} proposed another approach that utilizes a deep CNN with parallel branches to locate fault regions and estimate insulator endpoints. The method offers high accuracy and robustness, outperforming previous approaches like Faster R-CNN, SSD, and cascading networks. Zhang et al. \cite{zhang_defgan_2021} took a different approach where they used a generative network including a denoising autoencoder, discriminator, and classifier to detect defects. The method comprises two stages: first, insulator extraction using cascaded deep segmentation networks (CDSNets); second, defect detection within sampled patches using the proposed DefGAN, scoring defects based on classifier anomaly probability and denoising autoencoder reconstruction error. Although the proposed model can detect mild deformities, it is sensitive to the noise in the image.

\begin{figure}[htb]
    \centering
    \includegraphics[width=0.5\textwidth]{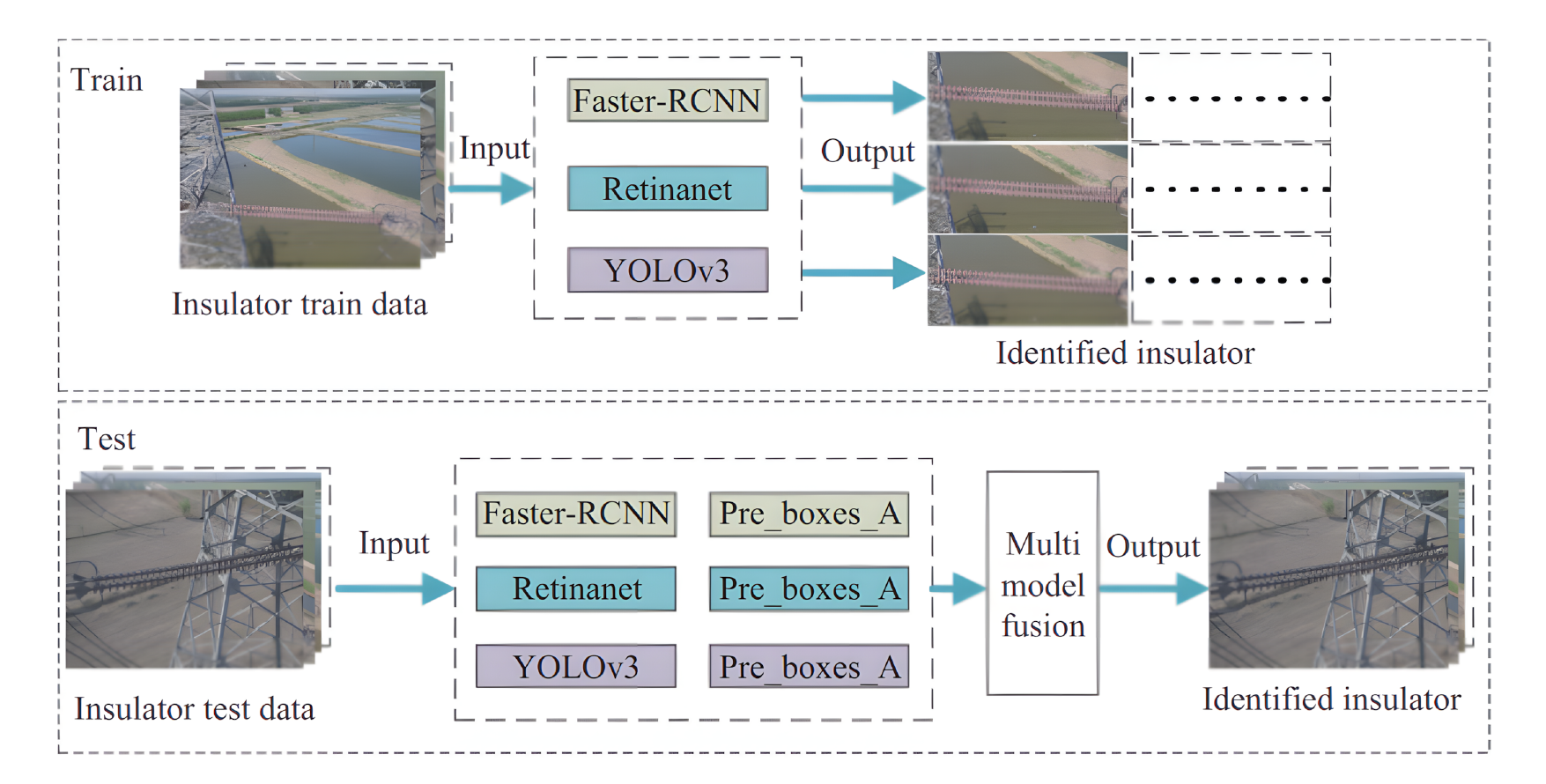}
    \caption{Simplified diagram of multi-model fusion network for detecting insulators \cite{wei_online_2022}.}
    \label{fig:wei_multimodel}
\end{figure}

One of the primary causes of insulator failures is the self-explosion of caps and several research works have been done to detect these faults. Cao et al. \cite{cao_accurate_2023} proposed an improved image augmentation method for the detection of self-detonation defects in aerial images of insulators. The method incorporated edge detection to enhance the shape features of insulators, which were then used to guide the augmentation process. The authors utilized the YOLOv3 model for insulator detection and an improved ResNet-18 model for defect classification. The proposed method was evaluated on a dataset of aerial images and showed superior performance compared to baseline models and other augmentation techniques. However, the authors acknowledged the need for further evaluation on larger and more diverse datasets, including images captured under different weather conditions. Additionally, the proposed method relied on edge detection as prior knowledge, which might not be sufficient for detecting more complex or subtle defects. In another work, Wei et al. \cite{wei_online_2022} introduced a fault detection scheme for insulator self-explosion, leveraging edge computing and deep learning to address issues with traditional centralized cloud computing. It employs a lightweight SSD target recognition network at the edge, replacing VGG16 with MobileNets to reduce computation. In the cloud, three distinct detection algorithms (Faster-RCNN, Retinanet, YOLOv3) are used to identify insulator self-explosion areas, and a novel multimodel fusion algorithm (Figure \ref{fig:wei_multimodel}) enables overhead transmission line fault detection. Results demonstrate effective data reduction, achieving an average cloud recognition accuracy of 95.75\%, with a modest increase in edge device power consumption.

% {\color{blue}
{\scriptsize
\begin{longtable}{P{.05} P{.1} P{.1} P{.1} P{.1} P{.15} P{.1} P{.1}} \caption{Summary of insulator structural fault detection studies.}
\label{tab:insulator_structural_faults} \\
\hline  

Year \& Ref & Component & Type of Detection & Type of Fault & Imaging Platform & Dataset & Algorithm & Performance \\ 
\hline 

2019 \cite{jiang_insulator_2019} & Glass and Porcelain Insulators & Bounding Box Detection & Insulator Structural Defects & UAV & 485 RGB Images & SSD & Precision: 91.23\%, Recall: 93.69\% \\

2020 \cite{tao_detection_2020} & Insulators & Bounding Box Detection & Structural Defects & UAV & 1956 RGB Images & Cascaded DNN & F1 Score: 93.3\%, Speed: 135ms/Image \\

2020 \cite{wang_detection_2020} & Insulators & Bounding Box Detection & Insulator Structural Defect & NA & 2560 RGB Images & Improved RetinaNet & Accuracy: 98.38\% \\

2021 \cite{liu_box_point_2021} & Insulators & Bounding Box Detection & Structural Faults & Aerial Vehicle & 969 RGB Images & Box Point Detector & mAP@0.5: 94.0\% \\

2021 \cite{zhang_insudet_2021} & Insulators & Bounding Box Detection & Missing Cap & UAV \& Synthetic & CPLID Dataset: 848 RGB Images, Pascal VOC Dataset: 5011 RGB Images & Improved YOLOv3 & Pascal VOC: mAP@0.75: 45.90\% CPLID: mAP@0.75: 64.05\% \\

2021 \cite{zhang_defgan_2021} & Insulators & Classification & Insulator Structural Defect & High Speed Railway Catenary & 800 RGB Images & GAN & F1-Score: 0.95 \\

2022 \cite{antwi_bekoe_deep_2022} & Insulators & Instance Segmentation & Insulator Defects & UAV & 1523 RGB Images & Mask R-CNN & AP: 89.4\% \\

2022 \cite{hao_insulator_2022} & Insulators & Bounding Box Detection & Bunch Drop Defect & UAV \& Synthetic & CPLID Dataset: 848 RGB Images & Improved YOLOv4-ResNest & mAP: 95.63\% \\

2022 \cite{wei_online_2022} & Insulators & Bounding Box Detection & Self-Explosion Defect & UAV & 8500 RGB Images & F-RCNN, RetinaNet, YOLOv3 Fusion & Precision: 99.04\%, Recall: 93.69\% \\

2023 \cite{cao_accurate_2023} & Glass Insulators & Bounding Box Detection \& Classification & Self-Explosion Defects & UAV & 8463 RGB Images & YOLOv3 \& Improved ResNet-18 & F1-Score: 86.25\% \\

2023 \cite{fu_small_sized_2023} & Insulators & Bounding Box Detection & 5 Types of Structural Defects & UAV \& Synthetic & CPLID Dataset: 848 RGB Images & I2D-Net (Based on F-RCNN) & mAP: 89.6\% \\

2023 \cite{hao2023pkamnet} & Insulator & Bounding Box Detection & Structural Defect & Aerial Vehicle & 900 RGB Images & PKAMNet & mAP@0.5: 95.5\% \\

2023 \cite{singh_2023_interpretable} & Insulators & Classification & Structural Defects & UAV & 848 RGB Images & Ps-ProtoPNet & F1-Score: 0.996 \\

2023 \cite{zhang_multi_objects_2023} & Insulators & Bounding Box Detection & Self-Explosion Defect & UAV \& Synthetic & CPLID Dataset: 848 RGB Images & GhostNet-YOLOv4 & mAP: 99.50\% \\

2024 \cite{jain2024transfer} & Insulator & Bounding Box Detection & Structural Defects & UAV & 5939 RGB Images & YOLO-v5 and DETR & mAP: 98\% \\

2024 \cite{wang2024mci} & Insulator & Bounding Box Detection & Structural Defect & UAV and Synthetic & CPLID Dataset: 848 RGB Images & Improved YOLO-based Network & mAP@0.5: 85.8\% \\
\hline

\end{longtable}
}
% }

\subsection{Detection of Conductor Faults}
Transmission line conductors are the lifelines of the electrical power grid, carrying electrical energy over long distances. Conductor defects, such as aging, corrosion, internal damage, and contamination, pose significant threats to the reliable and safe operation of power transmission systems. This subsection explores the challenges associated with conductor defect detection and highlights the use of computer vision technologies to enhance the efficiency and accuracy of inspections. Table \ref{tab:conductor_faults} summarizes the literature on conductor faults.

\begin{table*}[htb]
% {\color{blue}
\scriptsize
\caption{Summary of conductor fault detection studies.}
\label{tab:conductor_faults}
\begin{tabular}{P{.05} P{.1} P{.1} P{.1} P{.1} P{.15} P{.1} P{.1}}    

\hline  
Year \& Ref & Component & Type of Detection & Type of Fault & Imaging Platform & Dataset & Algorithm & Performance \\ 
\hline 

2019 \cite{li_image_2019} & Oil Transformer and Conductor & Classification & Internal Defects & Still Camera & 12000 UV, IR and Visible Images & Capsule Network & Qualitative Assessment \\

2021 \cite{rong_intelligent_2021} & Conductor & Bounding Box Detection & Vegetation & Tower Mounted Camera & 70 RGB Images & Cascaded Network (Based on F-RCNN) & Accuracy: 95\% \\

2022 \cite{yi_intelligent_2022} & Conductor & Classification & Aging Defect & NA & 5200 RGB Images & Improved AlexNet & Avg. MAE: 3.80 \\

2023 \cite{wang_internal_2023} & Conductors & Bounding Box Detection & 4 Types of Composite Core Defects & Inspection Robot & 2500 X-Ray Images & CenterNet-NDT (Based on ResNet50) & mAP: 90.60\% \\

2024 \cite{bergmann2024approach} & Conductor & Semantic Segmentation & Vegetation & Land & 851 LiDAR Images & UNet-based Network & Precision: Above 90\% \\
\hline

\end{tabular}
% }
\end{table*}

External defects of conductor wires can be detected using visible light images. These types of defects can be due to aging, structural damages or due to foreign objects which are described in more detail in the “Detection of Foreign Objects” subsection. Aging defects of conductors are one of the most common for aluminum conductors.  In a 2022 study, Yi et al. \cite{yi_intelligent_2022} introduced a novel approach to quantitatively diagnose the aging of conductor surfaces in smart high-voltage electricity grids. The model consists of an AlexNet-based deep convolution network and a specialized loss function that incorporates Gaussian label distribution. By reframing the conductor morphology aging problem as a multiclassification challenge, the model leverages a weakly labeled training dataset and a carefully designed loss function, combining entropy loss, cross-entropy loss, and Kullback–Leibler divergence loss. Comparative analysis with four commonly used CNN-based classifiers demonstrates superior performance on the collected dataset. However, the proposed model is suitable only when super-close-up high resolution conductor images are available.

Internal defects of the conductors can be challenging to detect using visible light images. To address this, Wang et al. \cite{wang_internal_2023} presented a novel automatic detection system for identifying internal defects in overhead aluminum conductor composite core (ACCC) transmission lines. The system utilizes an X-ray inspection robot equipped with a nondestructive testing (NDT) system to capture X-ray images of ACCC wires. The proposed system employs an anchor-free object detection model called CenterNet-NDT, which incorporates specialized modules like spatial pyramid pooling-cross stage partial convolution (SPPCSPC), polarized self-attention (PSA), and a weighted bidirectional feature pyramid network (SOFPN). Although CenterNet-NDT achieves a high mAP of 90.60\% on the IN-ACCC dataset, the instrumentation and maintenance required for this method can be challenging. 

\subsection{Fault Detection for Fittings: Pin-Bolts, Dampers, Suspension Clamps}
Power line fittings such as pin bolts, dampers, and suspension clamps are integral components of transmission and distribution systems. These components are tiny compared to the rest of the transmission line tower and they occupy a very small area in the images involving only a few pixels. As a result, accurately detecting faults in these components requires a high level of image processing and powerful deep-learning algorithms. Table \ref{tab:fittings_faults} summarizes the literature on power line fittings fault detection.

\begin{table*}[htb]
% {\color{blue}
\scriptsize
\caption{Summary of fault detection studies for power line fittings.}
\label{tab:fittings_faults}
\begin{tabular}{P{.05} P{.1} P{.1} P{.1} P{.1} P{.15} P{.1} P{.1}}    

\hline  
Year \& Ref & Component & Type of Detection & Type of Fault & Imaging Platform & Dataset & Algorithm & Performance \\ 
\hline 

2020 \cite{zhao_detection_2020} & Bolts & Bounding Box Detection & Missing Pin Defect & Tower Mounted Cameras & 1840 RGB Images & AVSCNet & mAR: 0.876 \\

2021 \cite{xiao_detection_2021} & Bolts & Bounding Box Detection & Pin Defects & Aerial Vehicle & 1500 RGB Images & Improved MTCNN & mAP: 94.76\% \\

2022 \cite{li_pin_2022} & Bolts & Bounding Box Detection & 2 Types of Pin Defects & UAV & 482 RGB Images & EfficientDet-D7 & mAP: 54.3\%, mAR: 63.4\% \\

2022 \cite{zhang_attention_guided_2022} & Dampers and Suspension Clamps & Bounding Box Detection & Rust Defects and Abnormal Conditions & UAV & 1209 RGB Images & AGMNet & Rust: mAP: 75.4\%, Abnormal: mAP: 92.7\% \\

2022 \cite{zhao_new_2022} & Bolts & Classification & 5 Types of Bolt Defects & UAV & 1944 RGB Images & VFPKNet & Accuracy: 83.29\% \\

2023 \cite{huang_structural_2023} & Damper & Classification & Structural Defects & UAV & 240 Aerial Images & Improved GrabCut & Accuracy: 95.76\% \\

2023 \cite{jiao2023defective} & Bolts & Bounding Box Detection & Structural Defect & UAV & 32094 Images & MARF-CCN Network & mAP: 87.16\% \\

2023 \cite{luo_ultrasmall_2023} & Bolts & Bounding Box Detection & Structural Defects & UAV & 1852 RGB Images & UBDDM (Based on ResNet-50 \& Attention) & mAP: 78.6\% \\

2023 \cite{song_deformable_2023} & Fittings & Bounding Box Detection & Rust Defect & NA & TLCF Dataset: 700 RGB Images & Deformable YOLOX & mAP@0.5: 0.857, mAP@0.5:0.95: 0.533 \\

2023 \cite{zhang_pa_detr_2023} & Bolts & Classification & 4 Types of Bolt Defects & UAV & VIBD Dataset: 8972 Bolt Instances in RGB Images & PA-DETR (Based on ResNet50, FPN and Attention) & mAP: 81.9\% \\

2023 \cite{zhang2023dsa} & Dampers & Bounding Box Detection & Structural Defect & UAV & 490 RGB Images & DSA-Net & mAP@0.5: 0.935 \\
\hline

\end{tabular}
% }
\end{table*}

Zhao et al. \cite{zhao_new_2022} proposed RFBD (Recognition of Bolt Faults using multilabel image recognition) that comprises two networks: VFSKnet leverages professional posterior knowledge to model relationships between bolt defect labels, while VFPKnet extracts structural features to capture fine-grained details. After combining and weighting these subnetworks, RFBD achieves label-level accuracy of 93.91\% and image-level accuracy of 83.29\% in detecting 5 types of bolt defects. However, the proposed solution is not end-to-end and requires properly segmented images of bolts. In a similar work, Zhang et al. \cite{zhang_pa_detr_2023} addressed the challenge of visually indistinguishable bolt defects in transmission lines by proposing an end-to-end detection method based on transmission line knowledge reasoning. It employs the DETR (End-to-End Object Detection with Transformers ) \cite{carion2020end} model augmented with a dilated encoder module to capture multiscale features. Additionally, a Transmission Line Image Relative Position Encoding (TL-iRPE) is designed to infer bolt position knowledge. The approach includes a bolt attributes classifier and a bolt defects classifier, combining position and attributes knowledge to enhance defect detection accuracy.

One of the most prominent faults that occurs often is the missing pin of bolts. Due to its tiny nature, it can be very challenging to detect these faults. Zhao et al. \cite{zhao_detection_2020} proposed the AVSCNet model to address the challenge of detecting pin-missing defects in bolts on transmission lines using aerial images. The model tackled the issue of varying visual shapes of bolts due to different camera angles by employing an automatic visual shape clustering method. It also incorporated feature enhancement, feature fusion, and expanded region-of-interest feature extraction to improve the detection of small-scale defects in complex aerial images. However, the model's performance might be affected by the distributional differences in aerial images captured from diverse transmission line structures and environments.

Structural anomalies and rusting of the dampers and suspension clamps can compromise the transmission line system causing failure. Zhang et al. \cite{zhang_attention_guided_2022} introduced the AGMNet, an attention-guided multitask convolutional neural network designed for the detection of power line parts in aerial images. The network incorporated a region attention mechanism to enhance the feature representation of objects, a refinable Region Proposal Network (RPN) to improve proposal quality, and a multitask learning framework to simultaneously detect power line fittings (dampers and suspension clamps), identify their rust levels, and detect abnormal conditions. However, the authors acknowledged the need for further evaluation on larger and more diverse datasets. Additionally, the distinction between different rust levels was found to be challenging, leading to less accurate rust level identification. The authors suggested further research to improve the accuracy of rust level detection.  

\subsection{Fault Recognition on Towers}
Towers are the sturdy backbone that supports high-voltage conductors, ensuring the reliable delivery of electricity over long distances. Traditional methods for fault detection in transmission lines have limitations, such as susceptibility to noise and transient fluctuations. To address these challenges, Wang et al. \cite{wang_image_2019} proposed a novel approach for fault zone detection that emphasizes fine-grained categorization and quality-awareness. The primary objective is to identify the most distinguishing image patches for classification. The key components of their method involve extracting features from line currents using a Fast R-CNN-based image sample decomposition, with a quality module selecting the most informative regions. These extracted features are then utilized in a Support Vector Machine (SVM) \cite{cortes1995support} for fault recognition. In a similar work, Liang et al. \cite{liang_detection_2020} created four medium-sized datasets for training component detection and classification models. They also employed data augmentation techniques to balance the imbalanced classes. Furthermore, the authors proposed a multi-stage component detection and classification approach using the SSD network and deep ResNet to detect small components and faults. The results demonstrate that the proposed system is both fast and accurate in detecting common faults in tower components, such as missing top caps, cracks in poles and cross arms, woodpecker damage on poles, and rot damage on cross arms. Table \ref{tab:tower_faults} summarizes the literature on tower faults.

\begin{table*}[htb]
% {\color{blue}
\scriptsize
\caption{Summary of tower fault detection studies.}
\label{tab:tower_faults}
\begin{tabular}{P{.05} P{.1} P{.1} P{.1} P{.1} P{.15} P{.1} P{.1}}   

\hline  
Year \& Ref & Component & Type of Detection & Type of Fault & Imaging Platform & Dataset & Algorithm & Performance \\ 
\hline 

2019 \cite{nguyen_intelligent_2019} & Tower Components & Classification & Structural Defects & UAV & 123151 RGB Images & ResNet & F1-Score: 77.96\% \\

2019 \cite{wang_image_2019} & Tower Components & Classification & Line Faults & NA & 11600 RGB Images & SVM & Accuracy: 96.73\% \\

2020 \cite{liang_detection_2020} & Tower Components & Bounding Box Detection & 8 Types of Structural Defects & UAV & WIRE\_10 Dataset: 8000 RGB Images & F-RCNN & mAP: 91.11\% \\

2021 \cite{odo_aerial_2021} & Tower Components & Classification & Surface and Structural Defects & Helicopter & Insulators: 25804 RGB Images, Bolts: 94,619 RGB Images & EfficientNetB0 & ROC: 0.94 (insulator), ROC: 0.98 (Bolts) \\

2022 \cite{liu2022component} & Tower Components & Bounding Box Detection & Structural Defect & UAV & 1309 RGB Images & Graph Neural Network & mAP@0.5: 89.1\% \\

2022 \cite{stefenon_classification_2022} & Tower Components & Classification & Structural Defects & Still Camera & 240 RGB Images & Inception v3 & F1-Score: 84.50\% \\

2022 \cite{stefenon_semi_protopnet_2022} & Tower Components & Classification & Structual Defects & Still Camera & 240 Images & Semi-ProtoPNet (Based on VGG-19) & Accuracy: 97.22\% \\

2023 \cite{liu2023fault} & Tower Components & Classification & Structural Defect & Aerial Vehicle & TLAD Dataset: 4811 RGB Images & NMHEM Model & F1 Score: 80.5\% \\

2023 \cite{yi2023pstl} & Tower Components & Bounding Box Detection & Structural Defect & UAV & 956 RGB Images & PSTL-Net & mAP: 0.848 \\

2024 \cite{zhong2024visual} & Substation and Tower Components & Classification and Bounding Box Detection & Structural Defect & UAV and Land & Substation Dataset: 2000 Images; UAV Dataset: 5048 Images; CPLID Dataset 848 Images & Federated Learning & mAP: .990 \\
\hline

\end{tabular}
% }
\end{table*}

\subsection{Detection of Foreign Objects}
Foreign object-related anomalies and faults in power lines encompass a multitude of potential hazards that can pose significant threats to the reliable operation of electrical transmission systems. 
% {\color{blue}
These hazards can range from inadvertent interference caused by construction equipment and vehicles to more natural occurrences like fires, bird nests, and overgrown vegetation among others. 
% } 
The presence of foreign objects in proximity to power line components can lead to various issues, including interruptions in electrical supply, damage to critical equipment, and heightened safety concerns for both utility personnel and the public. Table \ref{tab:foreign_object_detection} summarizes the literature on foreign object faults.

\begin{table*}[htb]
% {\color{blue}
\scriptsize
\caption{Summary of foreign object detection studies in power line systems.}
\label{tab:foreign_object_detection}
    
\begin{tabular}{P{.05} P{.1} P{.1} P{.1} P{.1} P{.1} P{.1} P{.15}}
\hline  
Year \& Ref & Component & Type of Detection & Type of Fault & Imaging Platform & Dataset & Algorithm & Performance \\ 
\hline 

2020 \cite{zhang_cloud_edge_2020} & Tower, Conductor & Bounding Box Detection & 7 Types of Foreign Objects & NA & 926 \& 2000 RGB Images & YOLOv4 & - \\

2020 \cite{zhu_deep_2020} & Tower, Conductor & Bounding Box Detection & Foreign Objects & Tower Mounted Camera & 8000 RGB Images & DBF-NN & mAP: 88.1\% \\

2022 \cite{ge_birds_2022} & Tower & Bounding Box Detection & Bird Nest & Aerial Platform & 3695 RGB Images & YOLOv5 & mAP: 83.4\%, FPS: 85.32 \\

2022 \cite{li_improved_2022} & Tower, Conductor & Bounding Box Detection & 4 Types of Foreign Objects & NA & 35000 RGB Images & YOLOv3-MobileNetv2 & mAP: 0.832, FPS: 60 \\

2023 \cite{bi_yolox_2023} & Tower, Insulator & Bounding Box Detection & Bird Nest & UAV \& Synthetic & Bird Nest: 2864 Images, CPLID Dataset: 848 RGB Images & YOLOX++ & mAP: 86.8\% \\

2023 \cite{qiu_lightweight_2023} & Tower & Bounding Box Detection & Foreign Objects & UAV \& Manual & 1232 RGB Images & YOLOv4 with Attention & mAP: 96.71\%, FPS: 45 \\

2023 \cite{yu_foreign_2023} & Tower, Conductor & Bounding Box Detection & 6 Types of Foreign Objects & UAV & 8803 RGB Images & Improved YOLOv7 & mAP: 92.2\%, FPS: 19 \\

2023 \cite{zhang_edge_2023} & Conductor & Bounding Box Detection & Cranes & Tower Mounted Camera & 4000 RGB Images & Edge VIP (Based on YOLOv5s) & mAP: 50.60\% \\
\hline

\end{tabular}
% }
\end{table*}

Construction equipment and vehicles operating near power lines can accidentally make contact with electrical components, causing short circuits, equipment failure, and potentially even electrical fires. Similarly, fires in the vicinity of power lines can result from various sources, including wildfires, discarded cigarette butts, or even arcing caused by faulty equipment. Zhang et al. \cite{zhang_edge_2023} presented an edge-based framework for power transmission line abnormal target detection, focusing on overcoming resource limitations and improving model performance. To mitigate the lack of labeled data, deep semi-supervised learning was introduced, which can refine the decision boundary by learning from unlabeled samples. To achieve this, the framework starts with an initial model trained on a small amount of labeled data and then updates itself using unlabeled data. In another work, Zhang et al. \cite{zhang_cloud_edge_2020} introduced a framework that combines cloud and edge computing with deep learning techniques. Initially, a YOLOv4 model is trained in the cloud server for abnormal object detection. This trained model is then deployed to edge servers for real-time detection of abnormal objects in captured pictures. To address the limited initial data samples of only 926 images, enhancement techniques are used to increase the number of pictures, and real-time data streams are employed for incremental learning.

\begin{figure*}[htb]
    \centering
    \includegraphics[width=1\linewidth]{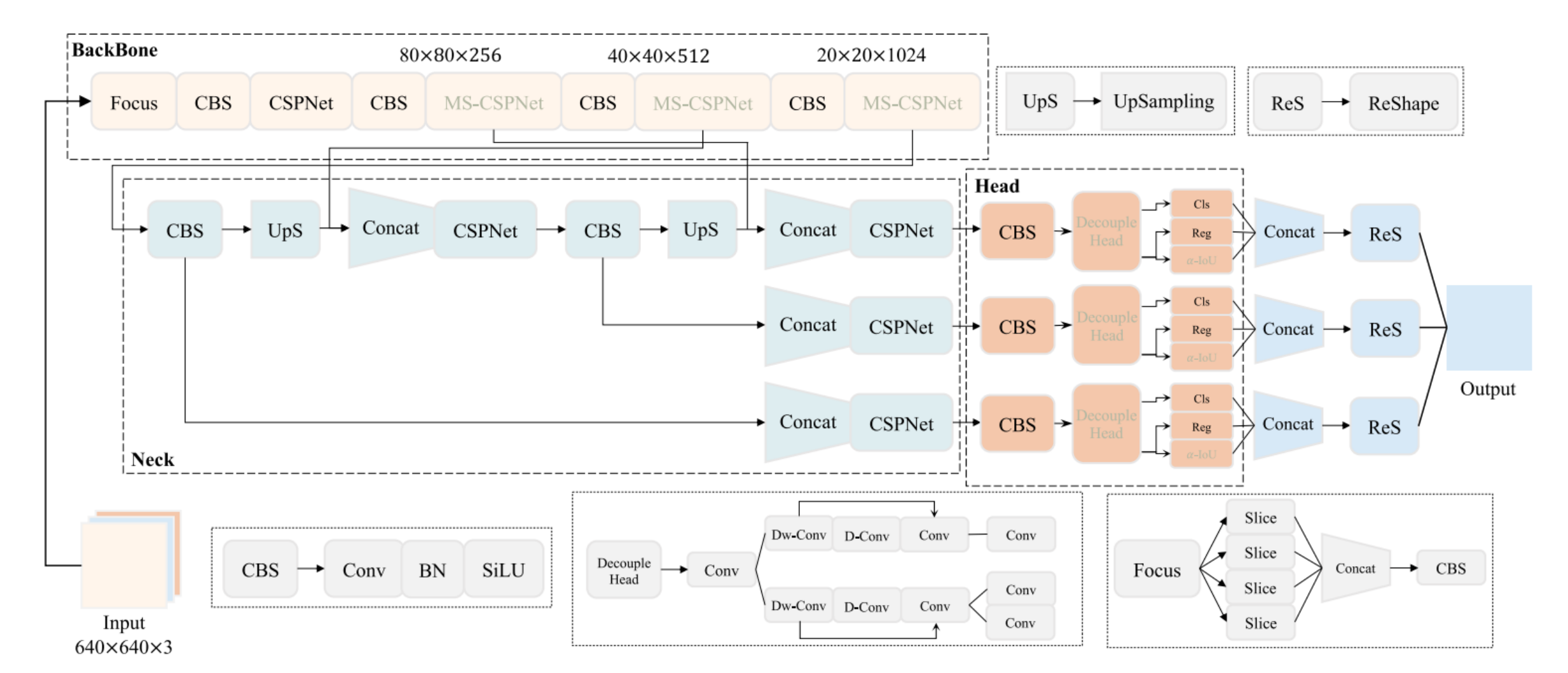}
    \caption{The simplified network architecture of the proposed YOLOX++ network \cite{bi_yolox_2023}.}
    \label{fig:yolox}
\end{figure*}

Bird nests, while seemingly innocuous, can also present challenges for power lines. Nests built on or near power line components can lead to electrical faults if they bridge connections or create conductive pathways. Ge et al. \cite{ge_birds_2022} proposed a bird's nest defect recognition method using YOLOv5, aiming to address the safety concerns posed by bird nests on power transmission towers. The method employs a YOLOv5- based architecture, comprising a backbone network, Feature Pyramid Network (FPN) \cite{lin2017feature}, and YOLO head, and undergoes multiple rounds of training with a constructed bird's nest defect database and transmission line model. Results demonstrate that the YOLOv5 model achieves an 83.40\% recognition rate for bird's nests while maintaining a high FPS rate of 85.32. In a similar work, Bi et al. \cite{bi_yolox_2023} presented a novel target detection model called YOLOX++, which is built upon the YOLOX \cite{ge2021yolox} architecture to enhance the detection of abnormal targets in transmission lines. It introduces a multiscale cross-stage partial network (MS-CSPNet) to fuse multiscale features and expand the receptive field of the target, improving channel combination (Figure \ref{fig:yolox}). Depth-wise and dilated convolutions are added to the object decoupling head to capture long-range dependencies of objects in feature maps. Additionally, the alpha loss function ($\alpha$-IoU) is incorporated to optimize small object localization. Experimental results demonstrate that YOLOX++ achieves detection accuracies of 86.8\% for high-voltage-tower bird nests and 96.60\% for power line insulators, outperforming the YOLOX model. On the PASCAL VOC dataset \cite{everingham2010pascal}, YOLOX++ exhibits a 9.30\% improvement in AP50 and a 5\% improvement in APS compared to YOLOX, showcasing its enhanced robustness for small target detection.

Another common cause of power line faults is vegetation, which, if left unmanaged, can grow into power lines, potentially causing short circuits, outages, or even wildfires during dry conditions. Rong et al. \cite{rong_intelligent_2021} proposed an intelligent detection framework for real-time monitoring of vegetation encroachment on power lines. The framework utilized binocular vision sensors mounted on transmission towers to capture images, which were then processed locally. The framework employed Faster R-CNN for vegetation detection, Hough transform for power line detection, and an advanced stereovision algorithm for 3D reconstruction. The advanced stereovision algorithm incorporated calibration optimization and world coordinate system transformation to improve accuracy in large-scale scenes. The authors acknowledged that the accuracy of the proposed method might be affected by complex terrain and weather conditions, which could impact image quality and object detection. Additionally, the computational efficiency of the framework was not explicitly addressed, which could be a concern for real-time monitoring applications. 

% {\color{blue}
To conclude, this section has demonstrated the transformative impact of computer vision and deep learning in revolutionizing power line fault diagnosis. Through the integration of advanced neural network architectures and multi-task models, significant progress has been made across multiple domains: from detecting small insulator defects with accuracies exceeding 97\%, to identifying structural anomalies in towers with mAP rates above 90\%, and recognizing foreign objects with real-time processing capabilities of up to 85 FPS. However, several critical challenges persist. The scarcity of large-scale, publicly available datasets remains a significant bottleneck, with most studies limited to small, proprietary datasets of under 5,000 images. The detection of miniature components like pin-bolts and dampers, which often occupy only a few pixels in aerial imagery, continues to challenge even state-of-the-art models. Environmental factors such as varying lighting conditions, weather effects, and complex backgrounds further complicate reliable fault detection. To address these limitations, promising research directions include: leveraging generative adversarial networks (GANs) for synthetic data augmentation, developing semi-supervised learning approaches to maximize the utility of unlabeled data, exploring attention-based architectures specifically optimized for small object detection, and investigating multi-task learning frameworks that can simultaneously handle different types of faults. The integration of edge computing solutions, as demonstrated in several studies, also shows promise in enabling real-time fault detection while managing computational constraints. These challenges and future research directions regarding power line fault diagnosis are explored in greater detail in Section \ref{sec:challenges}.
% }

%% file: sec_10_discussion.tex
\section{Discussion}\label{sec:discussion}

In this section, we present a comprehensive overview of the research articles reviewed in this study. Additionally, we provide a qualitative assessment of these articles. Initially, a set of criteria has been established for evaluating the papers, ensuring these criteria adequately reflect the core content of each article without favoring a particular subset. This involved a meticulous review of the articles and consultation with multiple co-authors to mitigate individual biases. Following this procedure, several criteria were identified for our qualitative assessment, including the use of large datasets (sample size greater than 5000), publicly available datasets, publication of accompanying code, fault detection across multiple components, employment of image modalities beyond visible light, application of advanced image processing techniques beyond resizing and cropping, use of synthetic and augmented data, focus on small components such as bolts, localization of specific components or faults, provision of performance metrics, acknowledgment of limitations, and suitability for real-time deployment. Based on these criteria, we conducted a thorough evaluation of all selected papers, and the findings are summarized in Table \ref{tab:quality_assesment}.

{\scriptsize
\begin{longtable}{| l | l | l | p{0.25cm} p{0.25cm} p{0.25cm} p{0.25cm} p{0.25cm} p{0.25cm} p{0.25cm} p{0.25cm} p{0.25cm} p{0.25cm} p{0.25cm} p{0.25cm} p{0.5cm} |}
\caption{Assessment of the reviewed literature}\label{tab:quality_assesment} \\
\hline
\rotatebox{90}{Number} & \rotatebox{90}{Authors} & \rotatebox{90}{Year of Publication} & \rotatebox{90}{Dataset Size $>$ 5000} & \rotatebox{90}{Dataset Availability} & \rotatebox{90}{Code Availability} & \rotatebox{90}{Multi-Component} & \rotatebox{90}{Other Image Modalities} & \rotatebox{90}{Image Processing} & \rotatebox{90}{Synthetic Data} & \rotatebox{90}{Data Augmentation} & \rotatebox{90}{Small Object} & \rotatebox{90}{Fault Localization} & \rotatebox{90}{Performance Metrics} & \rotatebox{90}{Limitations} & \rotatebox{90}{Real-Time} \\
\hline
1 & Liu et al. \cite{liu_discrimination_2017} & 2017 & $\times$ & $\times$ & $\times$ & $\times$ & \checkmark & $\times$ & $\times$ & \checkmark & $\times$ & $\times$ & \checkmark & $\times$ & $\times$ \\
2 & Tao et al. \cite{tao2018detection} & 2018 & $\times$ & \checkmark & $\times$ & $\times$ & $\times$ & \checkmark & \checkmark & \checkmark & $\times$ & \checkmark & \checkmark & $\times$ & \checkmark \\
3 & Kang et al. \cite{kang_deep_2019} & 2019 & \checkmark & $\times$ & $\times$ & $\times$ & $\times$ & $\times$ & \checkmark & $\times$ & $\times$ & \checkmark & \checkmark & $\times$ & $\times$ \\
4 & Li et al. \cite{li_image_2019} & 2019 & \checkmark & $\times$ & $\times$ & \checkmark & \checkmark & \checkmark & $\times$ & $\times$ & $\times$ & $\times$ & $\times$ & $\times$ & $\times$ \\
5 & Miao et al. \cite{miao_insulator_2019} & 2019 & \checkmark & $\times$ & $\times$ & $\times$ & $\times$ & $\times$ & $\times$ & \checkmark & $\times$ & \checkmark & \checkmark & $\times$ & \checkmark \\
6 & Chen et al. \cite{chen_research_2019} & 2019 & \checkmark & $\times$ & $\times$ & \checkmark & $\times$ & \checkmark & $\times$ & \checkmark & $\times$ & \checkmark & \checkmark & $\times$ & \checkmark \\
7 & Jiang et al. \cite{jiang_insulator_2019} & 2019 & $\times$ & $\times$ & $\times$ & $\times$ & $\times$ & $\times$ & $\times$ & \checkmark & $\times$ & \checkmark & \checkmark & \checkmark & $\times$ \\
8 & Nguyen et al. \cite{nguyen_intelligent_2019} & 2019 & \checkmark & $\times$ & $\times$ & \checkmark & $\times$ & $\times$ & $\times$ & \checkmark & $\times$ & \checkmark & \checkmark & $\times$ & $\times$ \\
9 & Sadykova et al. \cite{sadykova2019yolo} & 2019 & $\times$ & $\times$ & $\times$ & $\times$ & $\times$ & $\times$ & $\times$ & \checkmark & $\times$ & \checkmark & \checkmark & $\times$ & \checkmark \\
10 & Wang et al. \cite{wang_image_2019} & 2019 & \checkmark & $\times$ & $\times$ & \checkmark & $\times$ & $\times$ & $\times$ & $\times$ & $\times$ & $\times$ & $\times$ & $\times$ & $\times$ \\
11 & Sadykova et al. \cite{sadykova2019yolo} & 2019 & $\times$ & $\times$ & $\times$ & $\times$ & $\times$ & $\times$ & $\times$ & \checkmark & $\times$ & $\times$ & \checkmark & $\times$ & \checkmark \\
12 & Ibrahim et al. \cite{ibrahim_application_2020} & 2020 & $\times$ & $\times$ & $\times$ & $\times$ & $\times$ & \checkmark & $\times$ & $\times$ & $\times$ & $\times$ & \checkmark & $\times$ & $\times$ \\
13 & Mussina et al. \cite{mussina_multi_modal_2020} & 2020 & \checkmark & $\times$ & $\times$ & $\times$ & $\times$ & $\times$ & $\times$ & \checkmark & $\times$ & $\times$ & \checkmark & $\times$ & \checkmark \\
14 & Wang et al. \cite{wang_detection_2020} & 2020 & $\times$ & \checkmark & $\times$ & $\times$ & $\times$ & $\times$ & $\times$ & $\times$ & $\times$ & \checkmark & \checkmark & \checkmark & \checkmark \\
15 & Zhang et al. \cite{zhang_multi_scale_2020} & 2020 & \checkmark & $\times$ & $\times$ & \checkmark & $\times$ & $\times$ & \checkmark & $\times$ & \checkmark & \checkmark & \checkmark & $\times$ & $\times$ \\
16 & Zhang et al. \cite{zhang_cloud_edge_2020} & 2020 & $\times$ & $\times$ & $\times$ & \checkmark & $\times$ & $\times$ & $\times$ & \checkmark & $\times$ & \checkmark & \checkmark & $\times$ & \checkmark \\
17 & Zhao et al. \cite{zhao_detection_2020} & 2020 & $\times$ & $\times$ & $\times$ & $\times$ & $\times$ & $\times$ & $\times$ & \checkmark & \checkmark & \checkmark & \checkmark & \checkmark & $\times$ \\
18 & Zhao et al. \cite{zhao2020image} & 2020 & \checkmark & \checkmark & $\times$ & $\times$ & \checkmark & $\times$ & $\times$ & $\times$ & $\times$ & $\times$ & \checkmark & $\times$ & $\times$ \\
19 & Zhu et al. \cite{zhu_deep_2020} & 2020 & \checkmark & $\times$ & $\times$ & \checkmark & $\times$ & $\times$ & $\times$ & \checkmark & $\times$ & \checkmark & \checkmark & \checkmark & \checkmark \\
20 & Odo et al. \cite{odo_aerial_2021} & 2021 & $\times$ & $\times$ & $\times$ & \checkmark & $\times$ & $\times$ & $\times$ & \checkmark & \checkmark & \checkmark & \checkmark & \checkmark & $\times$ \\
21 & Singh et al. \cite{singh_design_2021} & 2021 & $\times$ & $\times$ & $\times$ & $\times$ & \checkmark & \checkmark & $\times$ & $\times$ & $\times$ & $\times$ & \checkmark & $\times$ & $\times$ \\
22 & Waleed et al. \cite{waleed_drone_based_2021} & 2021 & $\times$ & $\times$ & $\times$ & $\times$ & $\times$ & $\times$ & $\times$ & $\times$ & $\times$ & \checkmark & \checkmark & \checkmark & \checkmark \\
23 & Xiao et al. \cite{xiao_detection_2021} & 2021 & $\times$ & $\times$ & $\times$ & $\times$ & $\times$ & $\times$ & $\times$ & \checkmark & \checkmark & \checkmark & \checkmark & $\times$ & $\times$ \\
24 & Zhang et al. \cite{zhang_defgan_2021} & 2021 & $\times$ & $\times$ & $\times$ & $\times$ & $\times$ & $\times$ & \checkmark & \checkmark & $\times$ & \checkmark & \checkmark & \checkmark & $\times$ \\
25 & Zhai et al. \cite{zhai_hybrid_2021} & 2021 & $\times$ & $\times$ & $\times$ &
\checkmark & $\times$ & $\times$ & $\times$ & \checkmark & \checkmark & \checkmark & \checkmark & $\times$ & $\times$ \\
26 & Rong et al. \cite{rong_intelligent_2021} & 2021 & $\times$ & $\times$ & $\times$ & $\times$ & $\times$ & $\times$ & $\times$ & $\times$ & $\times$ & \checkmark & \checkmark & $\times$ & $\times$ \\
27 & Zhang et al. \cite{zhang_defgan_2021} & 2021 & $\times$ & \checkmark & $\times$ & $\times$ & $\times$ & $\times$ & \checkmark & \checkmark & $\times$ & \checkmark & \checkmark & $\times$ & $\times$ \\
28 & Antwi-Bekoe et al. \cite{antwi_bekoe_deep_2022} & 2022 & $\times$ & $\times$ & $\times$ & $\times$ & $\times$ & $\times$ & $\times$ & \checkmark & $\times$ & \checkmark & \checkmark & \checkmark & $\times$ \\
29 & Ge et al. \cite{ge_birds_2022} & 2022 & $\times$ & $\times$ & $\times$ & $\times$ & $\times$ & $\times$ & $\times$ & \checkmark & $\times$ & \checkmark & \checkmark & $\times$ & \checkmark \\
30 & Hao et al. \cite{hao_insulator_2022} & 2022 & $\times$ & \checkmark & $\times$ & $\times$ & $\times$ & $\times$ & \checkmark & \checkmark & $\times$ & \checkmark & \checkmark & $\times$ & \checkmark \\
31 & Hunag et al. \cite{huang_structural_2023} & 2022 & $\times$ & $\times$ & $\times$ & $\times$ & $\times$ & \checkmark & $\times$ & \checkmark & \checkmark & \checkmark & \checkmark & $\times$ & \checkmark \\
32 & Li et al. \cite{li_improved_2022} & 2022& \checkmark & $\times$ & $\times$ & \checkmark & $\times$ & $\times$ & $\times$ & \checkmark & $\times$ & \checkmark & \checkmark & $\times$ & \checkmark \\
33 & Li et al. \cite{li_pin_2022} & 2022 & $\times$ & $\times$ & $\times$ & $\times$ & $\times$ & $\times$ & $\times$ & $\times$ & \checkmark & \checkmark & \checkmark & \checkmark & $\times$ \\
34 & Qiu et al. \cite{qiu_lightweight_2023} & 2022 & $\times$ & $\times$ & $\times$ & \checkmark & $\times$ & \checkmark & $\times$ & \checkmark & $\times$ & \checkmark & \checkmark & \checkmark & \checkmark \\
35 & Stefenon et al. \cite{stefenon_classification_2022} & 2022 & $\times$ & \checkmark & $\times$ & \checkmark & $\times$ & \checkmark & $\times$ & \checkmark & $\times$ & $\times$ & \checkmark & $\times$ & $\times$ \\
36 & Stefenon et al. \cite{stefenon_semi_protopnet_2022} & 2022 & $\times$ & \checkmark & $\times$ & \checkmark & $\times$ & $\times$ & $\times$ & \checkmark & $\times$ & $\times$ & \checkmark & \checkmark & $\times$ \\
37 & Wei et al. \cite{wei_online_2022} & 2022 & \checkmark & $\times$ & $\times$ & $\times$ & $\times$ & $\times$ & $\times$ & \checkmark & $\times$ & \checkmark & \checkmark & $\times$ & $\times$ \\
38 & Yang et al. \cite{yang_vision_based_2022} & 2022 & $\times$ & $\times$ & $\times$ & $\times$ & $\times$ & $\times$ & $\times$ & $\times$ & $\times$ & \checkmark & $\times$ & $\times$ & $\times$ \\
39 & Yi et al. \cite{yi_intelligent_2022} & 2022 & \checkmark & $\times$ & $\times$ & $\times$ & $\times$ & $\times$ & $\times$ & \checkmark & $\times$ & $\times$ & \checkmark & $\times$ & $\times$ \\
40 & Zhai et al. \cite{zhai_multi_fitting_2022} & 2022 & $\times$ & $\times$ & $\times$ & \checkmark & $\times$ & $\times$ & $\times$ & \checkmark & \checkmark & \checkmark & \checkmark & $\times$ & $\times$ \\
41 & Zhang et al. \cite{zhang_attention_guided_2022} & 2022 & \checkmark & \checkmark & \checkmark & $\times$ & $\times$ & $\times$ & \checkmark & \checkmark & $\times$ & \checkmark & \checkmark & $\times$ & \checkmark \\
42 & Zhang et al. \cite{zhang_attention_guided_2022} & 2022 & $\times$ & $\times$ & $\times$ & \checkmark & $\times$ & $\times$ & $\times$ & $\times$ & \checkmark & \checkmark & \checkmark & $\times$ & $\times$ \\
43 & Zhao et al. \cite{zhao_new_2022} & 2022 & $\times$ & $\times$ & $\times$ & $\times$ & $\times$ & $\times$ & $\times$ & $\times$ & \checkmark & $\times$ & \checkmark & $\times$ & $\times$ \\
44 & Bi et al. \cite{bi_yolox_2023} & 2023 & $\times$ & \checkmark & $\times$ & \checkmark & $\times$ & $\times$ & \checkmark & \checkmark & $\times$ & \checkmark & \checkmark & \checkmark & $\times$ \\
45 & Cao et al. \cite{cao_accurate_2023} & 2023 & \checkmark & $\times$ & $\times$ & $\times$ & $\times$ & \checkmark & $\times$ & \checkmark & $\times$ & \checkmark & \checkmark & $\times$ & $\times$ \\
46 & Dong et al. \cite{dong_improved_2023} & 2023 & $\times$ & $\times$ & $\times$ & \checkmark & $\times$ & $\times$ & $\times$ & \checkmark & \checkmark & \checkmark & \checkmark & $\times$ & $\times$ \\
47 & Fu et al. \cite{fu_small_sized_2023} & 2023 & $\times$ & \checkmark & $\times$ & $\times$ & $\times$ & $\times$ & \checkmark & \checkmark & $\times$ & \checkmark & \checkmark & $\times$ & $\times$ \\
48 & Luo et al. \cite{luo_ultrasmall_2023} & 2023 & $\times$ & $\times$ & $\times$ & $\times$ & $\times$ & $\times$ & $\times$ & $\times$ & \checkmark & \checkmark & \checkmark & $\times$ & $\times$ \\
49 & Roy et al. \cite{roy_accurate_2023} & 2023 & $\times$ & $\times$ & $\times$ & $\times$ & $\times$ & \checkmark & $\times$ & $\times$ & $\times$ & $\times$ & \checkmark & $\times$ & \checkmark \\
50 & Shuang et al. \cite{shuang_rsin_dataset_2023} & 2023 & $\times$ & \checkmark & $\times$ & $\times$ & $\times$ & $\times$ & $\times$ & $\times$ & $\times$ & \checkmark & \checkmark & \checkmark & \checkmark \\
51 & Singh et al. \cite{singh_2023_interpretable} & 2023 & $\times$ & \checkmark & $\times$ & $\times$ & $\times$ & $\times$ & $\times$ & \checkmark & $\times$ & \checkmark & \checkmark & $\times$ & $\times$ \\
52 & Song et al. \cite{song_deformable_2023} & 2023 & $\times$ & $\times$ & $\times$ & \checkmark & $\times$ & $\times$ & $\times$ & \checkmark & \checkmark & \checkmark & \checkmark & $\times$ & \checkmark \\
53 & Wang et al. \cite{wang_internal_2023} & 2023 & $\times$ & $\times$ & $\times$ & $\times$ & \checkmark & \checkmark & $\times$ & \checkmark & $\times$ & \checkmark & \checkmark & \checkmark & $\times$ \\
54 & Yang et al. \cite{yang_dra_net_2023} & 2023 & $\times$ & \checkmark & $\times$ & $\times$ & $\times$ & $\times$ & $\times$ & $\times$ & $\times$ & \checkmark & \checkmark & $\times$ & $\times$ \\
55 & Yu et al. \cite{yu_foreign_2023} & 2023 & \checkmark & $\times$ & $\times$ & \checkmark & $\times$ & $\times$ & $\times$ & \checkmark & $\times$ & \checkmark & \checkmark & \checkmark & \checkmark \\
56 & Zhang et al. \cite{zhang_edge_2023} & 2023 & $\times$ & $\times$ & $\times$ & $\times$ & $\times$ & $\times$ & $\times$ & \checkmark & $\times$ & \checkmark & \checkmark & $\times$ & $\times$ \\
57 & Zhang et al. \cite{zhang_multi_objects_2023} & 2023 & $\times$ & \checkmark & $\times$ & $\times$ & $\times$ & $\times$ & \checkmark & \checkmark & $\times$ & \checkmark & \checkmark & $\times$ & \checkmark \\
58 & Zhang et al. \cite{zhang_pa_detr_2023} & 2023 & \checkmark & $\times$ & $\times$ & $\times$ & $\times$ & $\times$ & $\times$ & $\times$ & \checkmark & $\times$ & \checkmark & $\times$ & \checkmark \\
59 & Zhou et al. \cite{zhou_insulator_2023} & 2023 & \checkmark & $\times$ & \checkmark & $\times$ & $\times$ & $\times$ & $\times$ & \checkmark & \checkmark & \checkmark & \checkmark & $\times$ & $\times$ \\
60 & Zhang et al. \cite{zhang_pa_detr_2023} & 2023 & \checkmark & $\times$ & $\times$ & $\times$ & $\times$ & $\times$ & $\times$ & $\times$ & \checkmark & $\times$ & \checkmark & $\times$ & \checkmark \\
61 & Hao et al. \cite{hao2023pkamnet} & 2023 & $\times$ & $\times$ & $\times$ & $\times$ & $\times$ & \checkmark & $\times$ & \checkmark & $\times$ & \checkmark & \checkmark & \checkmark & $\times$ \\
62 & Liu et al. \cite{liu2022component} & 2023 & $\times$ & $\times$ & \checkmark & \checkmark & $\times$ & $\times$ & $\times$ & \checkmark & \checkmark & \checkmark & \checkmark & $\times$ & $\times$ \\
63 & Liu et al. \cite{liu2023fault} & 2023 & $\times$ & $\times$ & $\times$ & \checkmark & $\times$ & $\times$ & $\times$ & $\times$ & $\times$ & $\times$ & \checkmark & $\times$ & $\times$ \\
64 & Jiao et al. \cite{jiao2023defective} & 2023 & \checkmark & $\times$ & $\times$ & $\times$ & $\times$ & $\times$ & $\times$ & \checkmark & \checkmark & \checkmark & \checkmark & \checkmark & $\times$ \\
65 & Zhong et al. \cite{zhong2024visual} & 2024 & \checkmark & $\times$ & $\times$ & \checkmark & $\times$ & $\times$ & \checkmark & $\times$ & $\times$ & \checkmark & \checkmark & \checkmark & $\times$ \\
66 & Zhang et al. \cite{zhang2023dsa} & 2024 & $\times$ & $\times$ & $\times$ & $\times$ & $\times$ & $\times$ & $\times$ & \checkmark & \checkmark & \checkmark & \checkmark & \checkmark & \checkmark \\
67 & Yi et al. \cite{yi2023pstl} & 2024 & $\times$ & \checkmark & \checkmark & \checkmark & $\times$ & $\times$ & $\times$ & $\times$ & \checkmark & \checkmark & \checkmark & \checkmark & $\times$ \\
68 & Wang et al. \cite{wang2024mci} & 2024 & $\times$ & \checkmark & \checkmark & $\times$ & $\times$ & $\times$ & \checkmark & \checkmark & $\times$ & \checkmark & \checkmark & \checkmark & $\times$ \\
69 & Shi et al. \cite{shi2024lskf} & 2024 & $\times$ & \checkmark & $\times$ & $\times$ & $\times$ & $\times$ & $\times$ & \checkmark & \checkmark & $\times$ & \checkmark & \checkmark & $\times$ \\
70 & Liu et al. \cite{liu2023tower} & 2024 & $\times$ & $\times$ & \checkmark & \checkmark & $\times$ & $\times$ & \checkmark & \checkmark & \checkmark & \checkmark & \checkmark & $\times$ & $\times$ \\
71 & Jain et al. \cite{jain2024transfer} & 2024 & \checkmark & $\times$ & $\times$ & $\times$ & $\times$ & \checkmark & $\times$ & \checkmark & $\times$ & \checkmark & \checkmark & $\times$ & $\times$ \\
72 & Dong et al. \cite{dong2024transmission} & 2024 & \checkmark & $\times$ & $\times$ & \checkmark & $\times$ & $\times$ & $\times$ & $\times$ & \checkmark & \checkmark & \checkmark & \checkmark & $\times$ \\
73 & Bergmann et al. \cite{bergmann2024approach} & 2024 & $\times$ & $\times$ & $\times$ & $\times$ & \checkmark & \checkmark & \checkmark & $\times$ & $\times$ & \checkmark & \checkmark & \checkmark & $\times$ \\
\hline 
& & & $30\%$ & $23\%$ & $8\%$ & $34\%$ & $8\%$ & $19\%$ & $19\%$ & $67\%$ & $30\%$ & $78\%$ & $96\%$ & $33\%$ & $33\%$ \\
\hline
\end{longtable}
}

Table \ref{tab:quality_assesment} reveals several noteworthy insights regarding the articles reviewed. The scarcity of publicly available datasets is evident, with only 23\% of the papers utilizing them, while the majority rely on privately generated datasets. This issue has been discussed in Section \ref{sec:datasets}, “Publicly Available Datasets”. Moreover, large datasets are seldom used, with only 30\% of the articles meeting our 5000-sample threshold. To address this, numerous studies have incorporated synthetic or augmented data, with image augmentation being particularly prevalent, utilized in 67\% of the studies.

The publication of source code alongside machine learning or deep learning papers is highly beneficial for enabling readers to replicate algorithms and reproduce results, provided the dataset is also available. Although not always possible, sharing the source code enhances the credibility and impact of the research. However, among the articles reviewed, only 6 out of 73 published their source code.

Detecting faults across various components presents significant challenges due to the diverse nature of components and faults, such as the noticeable size difference between power line insulators and bolts \cite{nguyen_intelligent_2019, liang_detection_2020}. Despite these challenges, 34\% of the papers successfully employed multi-task learning techniques to address this issue, demonstrating effective results and highlighting further research opportunities in generalizing across more component types, faults, and environments. The detection of small objects remains particularly challenging due to their size relative to other components, often requiring sophisticated algorithms with considerable potential for advancement \cite{odo_aerial_2021}.

Unlike periodic inspections, real-time systems offer continuous surveillance, promptly identifying and addressing emergent issues. UAVs equipped with real-time algorithms can rapidly cover extensive areas, delivering immediate data to operators and aiding in the localization of faults, thus reducing the time and labor costs associated with manual inspections \cite{zhang_cloud_edge_2020}. Real-time systems demand algorithms capable of operating at speeds typically around 30 FPS or more on low-powered edge devices \cite{miao_insulator_2019}. Approximately 33\% of the reviewed studies have tackled this challenge by developing performant algorithms suitable for real-time deployment, often relying on rapid object detection algorithms like YOLO \cite{sadykova2019yolo} and SSD \cite{miao_insulator_2019}.

\begin{figure*}[!ht]
    \centering
    \includegraphics[width=0.9\linewidth]{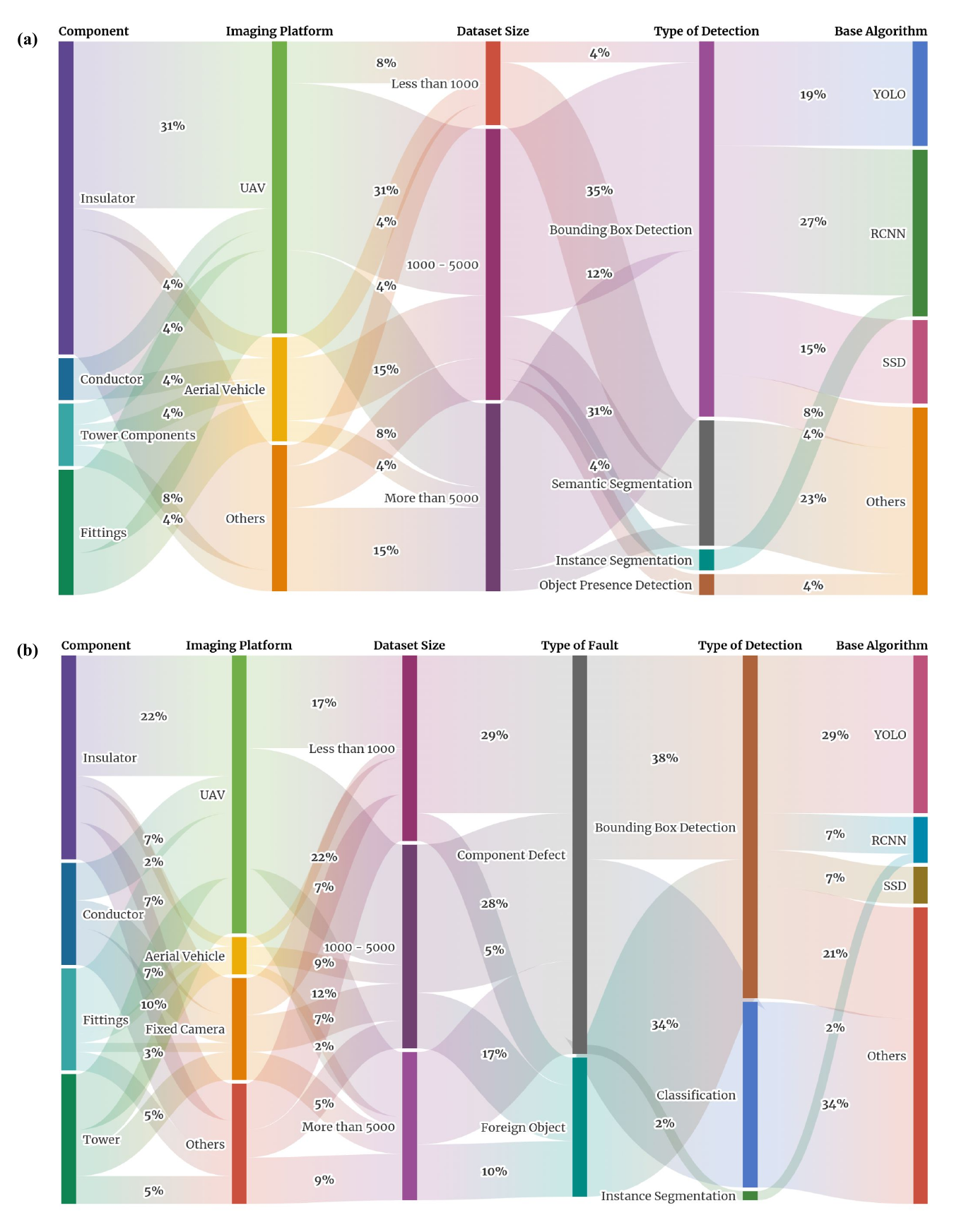}
    \caption{(a) The inter-relation between the different aspects in power line component detection. (b) The inter-relation between the different aspects in power line fault diagnosis.}
    \label{fig:sankey}
\end{figure*}

Automated power line inspection is a multi-step process where each step influences subsequent ones. For simplicity and clarity, it can be broken down into these key aspects: the component of interest, the choice of imaging platform, the size of the dataset, the type of detection, and the selected algorithm. When the process involves fault diagnosis, the type of fault should also be considered. By combining these aspects into a flow diagram, with the target component as the starting point and algorithm selection as the endpoint, we can establish a pattern for how decisions are made. With this context in mind, we carefully examined the reviewed literature to construct the diagrams in Figure \ref{fig:sankey}. Figure \ref{fig:sankey}(a) illustrates the decision-making process for automated component detection in power line infrastructure. The diagram shows that most research focused on insulators, utilized UAVs as the imaging platform, used datasets of 1000-5000 samples, and employed bounding box detection. Algorithm choices are balanced among YOLO, RCNN, SSD, or custom architectures. Notably, all works targeting semantic segmentation proposed their own custom network architecture. Similarly, Figure \ref{fig:sankey}(b) reveals the decision-making process for power line fault diagnosis. Again, a large portion of the works targeted insulators, used UAVs as the imaging platform, and collected datasets of 1000-5000 samples. Component fault detection is more common in the literature than foreign object detection. While YOLO, RCNN, and SSD have been used on multiple occasions for fault detection, most works propose custom network architectures.

%% file: sec_11_challenges.tex
\section{Current Challenges and Future Directions}\label{sec:challenges}
In the rapidly evolving field of automated power line inspection, significant advancements have been made. However, several critical challenges still persist that need to be addressed to further advance the state of the art. This section outlines these challenges and proposes potential future research directions.

% {\color{blue}
\subsection{Edge-Cloud Deployment Challenges}
The deployment of deep learning models for power line inspection faces significant architectural challenges in both cloud-centric and edge-based approaches. Traditional cloud computing, while offering substantial computational resources, suffers from high latency, excessive bandwidth consumption, and significant communication costs when processing the massive amount of visual data generated by inspection devices \cite{wang2020convergence, wei_online_2022}. Edge computing attempts to address these limitations by bringing computation closer to data sources, but edge devices typically lack the computational capacity to run sophisticated deep learning models effectively \cite{shi2016edge}. This limitation is particularly critical in power line inspection, where models must detect subtle defects and anomalies with high accuracy, creating a fundamental tension between model complexity and computational efficiency.

Edge-cloud fusion architectures present a promising direction to address these challenges by combining the strengths of both approaches. This fusion has led to the emergence of effective two-stage detection systems, where lightweight models at the edge perform initial coarse screening, followed by refined secondary recognition using more sophisticated models in the cloud \cite{wei_online_2022}. In this hierarchical framework, edge devices can efficiently filter and pre-process data, while cloud resources handle detailed analysis, significantly reducing data transmission while maintaining high accuracy \cite{wei_online_2022, zhang_edge_2023, zhang_cloud_edge_2020}. Future research should focus on several key areas to advance this fusion approach: developing more efficient model compression techniques for edge deployment \cite{howard2017mobilenets}, improving communication protocols for edge-cloud interaction \cite{shi2016edge}, creating adaptive frameworks for dynamic resource allocation \cite{wang2020convergence}, and investigating federated learning \cite{mcmahan2017communication} approaches for collaborative model training. These advancements could enable more efficient and reliable power line inspection systems while maintaining the accuracy needed for critical infrastructure monitoring.

\subsection{Multimodal Imaging and Fusion}
Power line inspection research has predominantly relied on optical imaging, with our comprehensive review revealing that only around 8\% of published works utilize other imaging modalities. This heavy dependence on optical imaging persists despite the known limitations of visible spectrum cameras in various environmental conditions and their inability to detect certain types of faults. Limited studies exploring alternative modalities demonstrated the complementary capabilities of infrared and ultraviolet imaging for detecting corona discharge and heating associated with leakage current flow in composite insulators \cite{singh_design_2021, liu_discrimination_2017, li_image_2019}. Their research highlighted how different imaging modalities can provide unique insights - with IR imaging excelling at detecting heat distribution patterns from current leakage, while UV imaging proved effective for visualizing corona discharge phenomena. However, such multimodal approaches remain vastly underutilized in power line inspection literature, despite their proven effectiveness in other domains such as medical imaging, remote sensing, and defense applications \cite{karim2023current, meher2019survey}.

The future of power line inspection could benefit significantly from greater adoption of multimodal imaging approaches. Recent developments in image fusion, as outlined in the comprehensive review by Karim et al. \cite{karim2023current}, demonstrate how combining multiple imaging modalities can provide more comprehensive information about the real world than any single modality alone. While optical images excel at providing high spatial resolution and clear texture details, other modalities like infrared can detect thermal anomalies, and ultraviolet can reveal corona discharge patterns \cite{shen2017research}. By integrating multiple modalities through advanced fusion techniques - ranging from conventional transform-based methods to emerging deep learning architectures \cite{ma2019fusiongan, han2020electrical} - future inspection systems could achieve more robust fault detection capabilities. 
% }

\subsection{Lack of Data Availability}
The scarcity of publicly available datasets remains a significant challenge in deploying deep learning for power line inspection \cite{song2020analysis}. Power line components and scenarios require vast, varied datasets for effective training. While researchers often create custom datasets \cite{chen_research_2019}, data protection regulations frequently prevent public sharing, as discussed in Section \ref{sec:datasets}.

Several approaches show promise in addressing this challenge. Synthetic data generation using GANs \cite{goodfellow2020generative} or Denoising Diffusion Models can create diverse, realistic power line images. For example, a recent study \cite{YE2024100250} enhanced Cycle-GAN \cite{zhu2017unpaired} with attention mechanisms to generate insulator defect images, demonstrating significant improvements in sample quality.

% {\color{blue}
Self-supervised pretraining \cite{raina2007self} offers another solution by leveraging unlabeled data to learn useful representations before fine-tuning on smaller labeled datasets. Additionally, few-shot learning approaches \cite{wang2020generalizing}, particularly meta-learning techniques \cite{finn2017model}, enable models to learn from limited examples. Combining these methods with transfer learning could effectively address the data scarcity challenge while maintaining robust performance in real-world applications.
% }

\subsection{Data Annotation}

The problem of data annotation presents a significant challenge in power line inspection, particularly given the complexity of power line components and the subtleties of potential faults. While many excellent annotation tools exist \cite{dwyer2024roboflow, HumanSignal, labelstudio}, the process remains time-consuming and labor-intensive, requiring substantial human intervention for verification. 
% {\color{blue}
The emergence of large foundation models like SAM (Segment Anything Model) \cite{kirillov2023segment} offers new opportunities for streamlining this process through zero-shot segmentation capabilities, though these models require significant computational resources and careful adaptation to the power line domain. The integration of such models into existing annotation workflows necessitates consideration of domain-specific fine-tuning strategies to ensure reliable performance in power line inspection contexts.
% }

Several promising approaches are being developed to address these challenges. Weakly supervised learning techniques \cite{zhou2018brief} have shown potential in reducing labeling requirements, as demonstrated by Choi et al. \cite{choi2021weakly} in their two-stage power line detection algorithm. The combination of these approaches with foundation models and specialized annotation tools could create more effective hybrid systems that leverage both general semantic understanding and domain-specific expertise. Additionally, self-supervised pretraining \cite{raina2007self} can significantly reduce the dependency on labeled datasets by equipping models with a deep understanding of structural and contextual features inherent in power line images, thereby streamlining the annotation process while maintaining accuracy and reliability.

\subsection{Very Small Components Detection}
Detecting small components like fittings, bolts, and fractures in power lines presents unique challenges due to their low resolution in images. These components, while critical for structural integrity, are often indistinguishable from complex backgrounds \cite{xiao_detection_2021}. The prevalent use of UAVs introduces additional challenges: image blur from drone motion and inability to capture close-up images due to high voltage risks. Consequently, small components often occupy only a few pixels in the captured images \cite{zhai_hybrid_2021}.

Recent advances offer promising solutions. Transformer-based approaches \cite{dong_improved_2023} have shown success in reducing errors in small and occluded object detection through lightweight self-attention modules. Image super-resolution techniques \cite{ledig2017photo} enhance resolution by inferring missing details and refining textures. Future research could focus on combining high-resolution imaging technology with advanced deep learning models, while improving multi-scale detection strategies and feature extraction methods for small objects \cite{hu2018small}.

% {\color{blue}
\subsection{Anomaly Detection for Unknown Defects}
Traditional supervised approaches often fail to identify novel power line defects absent from training data. Recent semi-supervised and unsupervised learning methods \cite{defard2021padim, batzner2024efficientad} offer solutions by learning normal patterns and detecting deviations.

Autoencoder-based methods with auxiliary anomaly localization enable end-to-end defect detection using only normal samples \cite{tsai2021autoencoder, sun2023semisupervised}. Feature embedding approaches using pre-trained networks reduce noise interference during reconstruction \cite{roth2022towards}. Memory-based methods that compare normal feature representations have improved detection of subtle anomalies \cite{yang2023memseg}, while combined reconstruction and discriminative training approaches enhance defect localization accuracy \cite{zavrtanik2021draem}. Future research should focus on developing efficient architectures suitable for edge deployment while maintaining detection accuracy.
% }

%% file: sec_12_conclusion.tex
\section{Conclusion}\label{sec:conclusion}
This review examines the evolution of vision-based power line inspection through deep learning technologies. We explored various imaging platforms and techniques, from UAVs to X-Ray imaging, analyzing their effectiveness for different inspection tasks. The review detailed how deep learning algorithms like YOLO, R-CNN, and SSD have advanced component detection and fault diagnosis, with particular success in insulator inspection. Current trends show promising developments in edge-cloud fusion architectures and two-stage detection approaches, balancing computational efficiency with accuracy. However, challenges persist in data availability, annotation efficiency and unknown defect detection among others. Solutions are emerging through synthetic data generation, few-shot learning, and semi-automatic labeling techniques. Looking forward, the integration of edge-cloud computing, multi-modal fusion, and novel learning approaches suggests a trajectory toward more resilient and adaptive power line maintenance systems. These advancements promise to enhance the reliability and efficiency of power infrastructure inspection while reducing operational costs and safety risks.

%% file: extras.tex
\section*{Declaration of Competing Interest}
Authors have no competing of interest to declare.

\section*{Acknowledgments}
This publication is supported by Iberdrola S.A. as part of its innovation department research activities. Its contents are solely the responsibility of the authors and do not necessarily represent the official views of the Iberdrola Group. The open access publication cost is covered by the Qatar National Library.

\section*{Declaration of Generative AI and AI-assisted Technologies in the Writing Process}
During the preparation of this work the authors used Google Gemini in order to improve readability and language. After using this tool, the authors reviewed and edited the content as needed and take full responsibility for the content of the publication.

\section*{Author Contributions}
\textbf{Md. Ahasan Atick Faisal}: Conceptualization, Data curation, Formal analysis, Writing - original draft. \textbf{Imene Mecheter}: Conceptualization, Writing - review \& editing. \textbf{Yazan Qiblawey}: Conceptualization, Writing - review \& editing. \textbf{Javier Hernandez Fernandez}: Project administration, Writing - review \& editing. \textbf{Muhammad E. H. Chowdhury}: Supervision, Validation, Writing - review \& editing. \textbf{Serkan Kiranyaz}: Supervision, Validation, Writing - review \& editing.

%% file: supplementary.tex
% {\color{blue}
\section{Deep Learning Models for Power Line Inspection}\label{appendix:dl_models}
Historically, power line inspections relied heavily on manual labor and traditional image processing techniques, which often proved time-consuming and prone to human error \cite{sundaram_deep_2021}. With the adoption of deep learning models, a profound shift has occurred. These models have showcased their ability to learn intricate patterns and structures within images, allowing for the precise identification and localization of various power line components. This transformation is not merely theoretical; it has translated into tangible benefits for power grid operators and maintenance teams \cite{sundaram_deep_2021}. This, in turn, enhances the reliability of power transmission infrastructure and mitigates the risks of unexpected outages and associated economic and safety implications. In this section, key deep learning techniques that have found application in power line component detection and fault diagnosis have been explored. The following sections delve into their advantages and use cases, shedding light on how these techniques are pushing the boundaries of what is achievable in the field of power line maintenance.

\subsection{You Only Look Once (YOLO)}
You Only Look Once (YOLO) \cite{redmon_you_2016} is a revolutionary real-time object detection system that has gained widespread recognition in computer vision applications. It stands out for its ability to swiftly process images and directly predict bounding boxes and class probabilities in a single evaluation. YOLO's efficiency and accuracy make it a compelling choice for power line component detection.

\begin{figure*}[htb]
    \centering
    \includegraphics[width=1\linewidth]{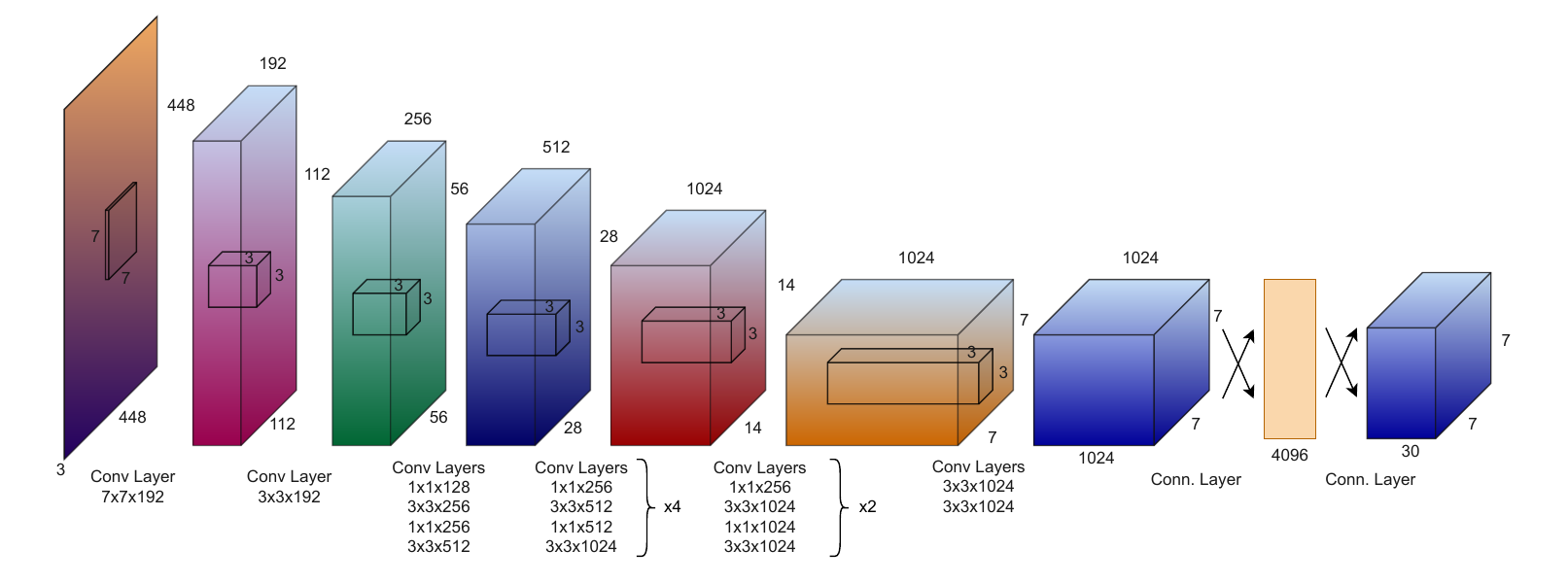}
    \caption{The architecture of the original YOLO network \cite{redmon_you_2016}.}
    \label{fig:yolo}
\end{figure*}

The original YOLO model (Figure \ref{fig:yolo}) introduced the concept of end-to-end object detection in real-time. It divides an image into a grid and predicts bounding boxes and class probabilities for objects within each grid cell. YOLOv2 \cite{redmon_yolo9000_2016} brought improvements in accuracy and flexibility by employing anchor boxes and multi-scale detection. It was also trained on a broader dataset, allowing it to detect a wide range of objects. YOLOv3 \cite{redmon_yolov3_2018} further enhanced the model's accuracy by utilizing a three-stage detection process and the addition of more anchor boxes. YOLOv4 \cite{bochkovskiy_yolov4_2020} introduced several architectural improvements, including the integration of the CSPDarknet53 backbone, PANet, and SAM block \cite{bochkovskiy_yolov4_2020}. These enhancements resulted in better performance in complex scenarios and more accurate component detection. The recently proposed YOLOv8 is built on top of the previous YOLO versions and designed to be faster, and more accurate \cite{jocher_yolo_2023}. 

In power line inspection, YOLO is utilized to identify and classify various components, such as insulators, dampers, pin bolts, conductor wires, and fittings \cite{sadykova2019yolo, singh_2023_interpretable, zhang_cloud_edge_2020}. Its speed and real-time capabilities are particularly advantageous when inspecting extensive stretches of power transmission infrastructure. YOLO's speed is one of its defining features. It operates at a significantly high frame rate, often exceeding real-time requirements \cite{li_improved_2022}. This speed advantage is particularly valuable in power line inspections, where rapid assessments of extensive infrastructure can be vital. The ability to process images quickly allows for the efficient identification of components, even in cases of frequent data acquisition through aerial surveys. It's worth noting that YOLO's real-time performance might require a sufficiently powerful hardware setup \cite{ultralyticsFrequentlyAsked}, but the trade-off between accuracy and processing speed can often be optimized according to specific project requirements.

\subsection{Region-Based CNNs (R-CNN, Fast R-CNN, Faster R-CNN)}
\begin{figure*}[htb]
    \centering
    \includegraphics[width=1\linewidth]{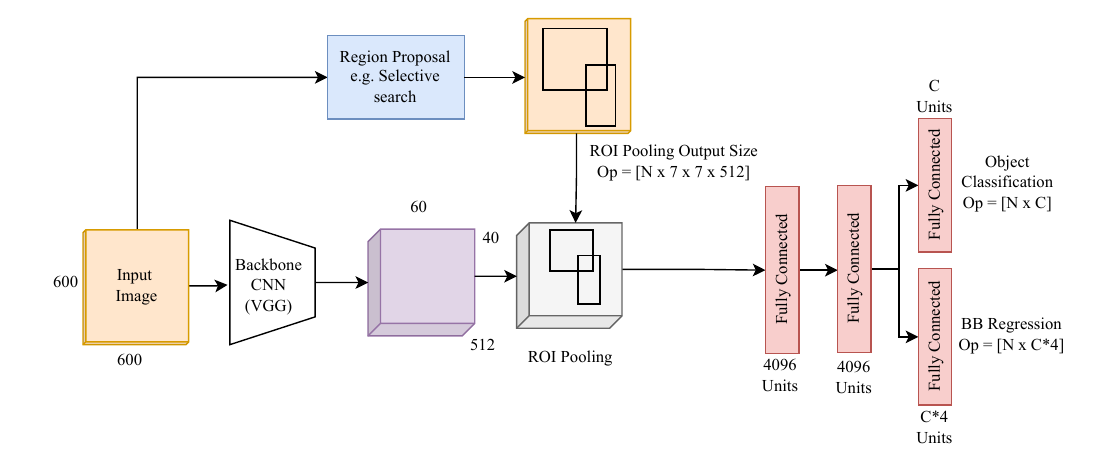}
    \caption{Simplified architecture of the Fast R-CNN Network.}
    \label{fig:rcnn}
\end{figure*}

Region-Based Convolutional Neural Networks (CNNs) \cite{girshick_rich_2014} represent a family of object detection models that focuses on detection accuracy while compromising on speed and complexity. This family includes R-CNN \cite{girshick_rich_2014}, Fast R-CNN \cite{girshick_fast_2015}, and Faster R-CNN \cite{ren_faster_2016}, each building upon the other to improve efficiency and accuracy in power line component detection. R-CNN slides an image window, extracts features for each window, and then classifies and refines bounding boxes for potential objects within those windows. Fast R-CNN (Figure \ref{fig:rcnn}) improves on R-CNN by processing the entire image at once with a single CNN to extract features, making it significantly faster. Faster R-CNN further refines the model by introducing the Region Proposal Network (RPN) \cite{ren_faster_2016} for generating region proposals. This innovation results in a more streamlined and faster detection process. These models have been applied in power line component detection to locate and classify insulators \cite{wei_online_2022}, dampers \cite{zhai_hybrid_2021}, pin bolts \cite{zhai_hybrid_2021}, conductor wires \cite{rong_intelligent_2021}, and other elements.

Region-based CNNs excel in precisely localizing objects within images, making them suitable for power line component identification. Fast R-CNN and Faster R-CNN integrate region proposal and feature extraction steps, enhancing processing efficiency. However, Training and fine-tuning region-based CNNs require substantial computational resources and a large labeled dataset. These models may need hardware acceleration for real-time performance \cite{bharati2020deep}.

\subsection{Single Shot MultiBox Detectors (SSD)}
Single Shot Detectors (SSD) \cite{liu_ssd_2016} is another object detection algorithm that combines high-speed processing with robust detection capabilities. SSD is designed for real-time object detection, eliminating the need for a separate region proposal step and streamlining all computations into a single network. SSD is employed to rapidly identify and classify various components, such as insulators \cite{miao_insulator_2019}, fittings \cite{nguyen_intelligent_2019}, conductor wires \cite{nguyen_intelligent_2019}, and other crucial infrastructure elements. 

\begin{figure*}[htb]
    \centering
    \includegraphics[width=1\linewidth]{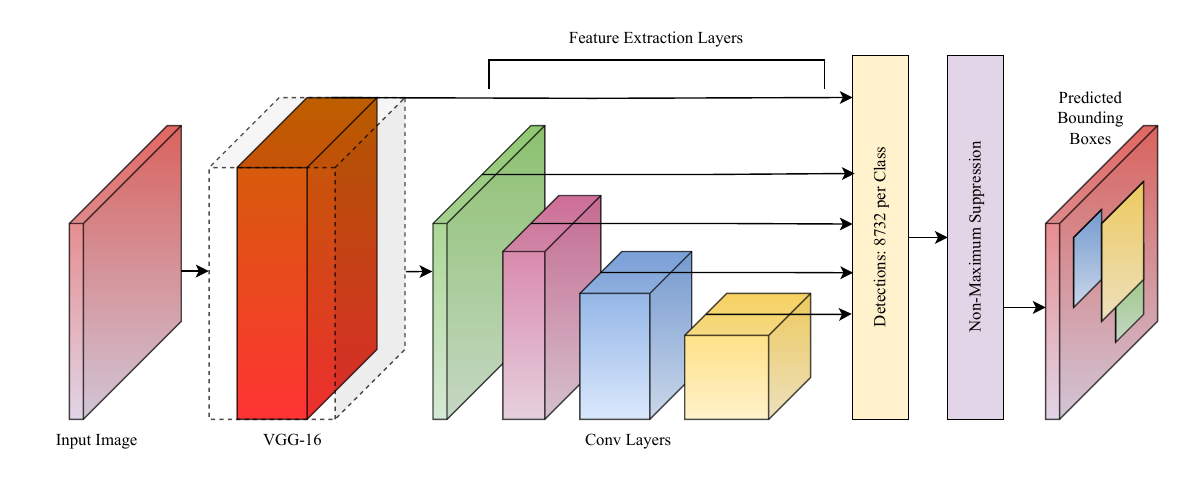}
    \caption{Simplified architecture of the Single Shot Multibox Detector (SSD) Network.}
    \label{fig:ssd}
\end{figure*}

SSD's primary advantage is its ability to process images rapidly while keeping a high enough accuracy \cite{huang2017speed}. This characteristic is essential for real-time inspections, particularly in scenarios where assessments of extensive power line infrastructure are required. SSD's architecture (Figure \ref{fig:ssd}) integrates all detection computations into a single network, eliminating the need for multiple stages, which simplifies implementation and results in efficient performance. While SSD offers impressive speed and efficiency, its performance may be compromised compared to models like Faster-RCNN \cite{huang2017speed}. Balancing speed and accuracy demands may require optimization for specific hardware configurations and project constraints.

\subsection{Transformer Architectures}
Transformer architectures have revolutionized various domains in deep learning, originally emerging as a powerful tool in natural language processing. Introduced by Vaswani et al. \cite{vaswani2017attention}, the transformer model is based on self-attention mechanisms, enabling it to capture long-range dependencies within data. Unlike traditional convolutional neural networks (CNNs), which are limited by their localized receptive fields, transformers excel in modeling global context, making them highly effective for complex tasks in computer vision, including power line inspection.

\subsubsection{Vision Transformers (ViT)}

Vision Transformers (ViT) \cite{dosovitskiy2020image} marked a paradigm shift by applying the transformer architecture directly to image data. ViTs divide an image into a sequence of patches, each treated similarly to tokens in a language model. These patches are then processed through multiple layers of self-attention, allowing the model to learn intricate relationships between different parts of the image. Although we could not find any research work on power line inspection that utilizes ViTs, they have potential in tasks requiring detailed analysis of visual data. Their ability to capture global information makes them particularly suitable for identifying subtle anomalies in power line components, such as bolt defects, micro-cracks in insulators or structural defect of conductor wires \cite{han2022survey}. However, ViTs require a very large dataset to train on to get rid of the inductive bias \cite{dosovitskiy2020image}. \\

\subsubsection{Swin Transformers}

Swin Transformers \cite{liu2021swintransformerhierarchicalvision}, or Shifted Window Transformers, build on the concept of ViTs by introducing a hierarchical structure that allows the model to operate at multiple scales. This architecture divides the image into non-overlapping windows and applies self-attention within each window. To capture cross-window information, the windows are shifted between layers, enabling the model to build a more comprehensive understanding of the image. The multi-scale feature representation of Swin Transformers is advantageous, especially in scenarios where defects or components vary in size. Swin Transformers can effectively manage high-resolution images, making them ideal for detecting and localizing faults in expansive power transmission networks, where both small and large defects need to be identified with precision \cite{dong_improved_2023}. \\

\subsubsection{Detection Transformers (DETRs)}

\begin{figure*}[htb]
    \centering
    \includegraphics[width=1\linewidth]{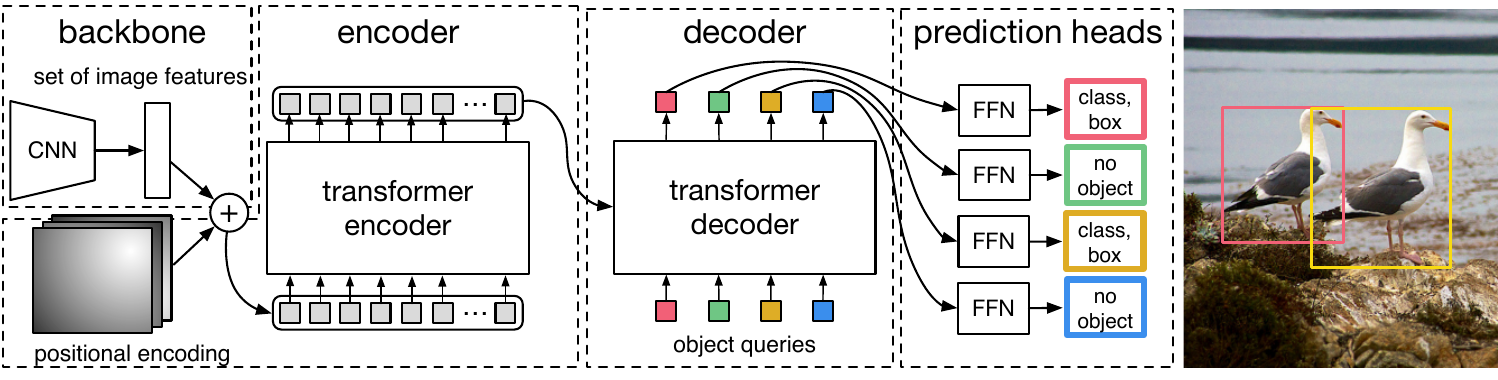}
    \caption{The simplified block diagram of the original DETR network \cite{carion2020end}}
    \label{fig:detr}
\end{figure*}

Detection Transformers (DETRs) \cite{carion2020end} integrate the transformer architecture into object detection tasks, offering an end-to-end approach that simplifies the traditional detection pipeline. The architecture of the DETR network is shown in Figure \ref{fig:detr}. DETRs eliminate the need for anchor boxes and region proposals, which are common in conventional object detection models. Instead, they leverage the transformer’s attention mechanism to directly predict object bounding boxes and class labels. Their ability to model complex interactions between objects within an image enhances detection accuracy, particularly in cluttered or complex scenes typical of power line infrastructure. Additionally, DETRs are robust to variations in object scale and orientation, which are common challenges in aerial imagery used for inspecting power lines \cite{zhang_pa_detr_2023, jain2024transfer}. \\

\subsection{Classification Algorithms}
Classification algorithms, particularly those pretrained on large datasets like ImageNet \cite{5206848}, have demonstrated remarkable capabilities in recognizing diverse patterns and anomalies in power line components. These algorithms excel in scenarios where pinpointing the exact location of a fault is unnecessary, and the primary goal is simply to determine whether a fault exists. When provided with an image of power line infrastructure or a specific segment of a power line component, these algorithms are capable of classifying the image as either faulty or in good condition. Key algorithms in this domain include ResNet, VGG, MobileNet and EfficientNet each bringing unique strengths to the table. 
 
\textbf{ResNet (Residual Networks)} \cite{he_2023_deep}, a pivotal model in deep learning, introduced the concept of residual learning to ease the training of very deep networks. It employs "skip connections" to jump over some layers, effectively addressing the vanishing gradient problem. In power line inspection, ResNet's ability to learn from a vast depth of layers makes it exceptionally good at recognizing complex patterns, crucial for identifying subtle anomalies in power lines \cite{wei_online_2022, cao_accurate_2023, luo_ultrasmall_2023}.

The \textbf{VGG (Visual Geometry Group)} \cite{simonyan2014very} network stands out for its simplicity and depth, with a uniform architecture that stacks convolutional layers directly on top of each other. This design, while computationally intensive, offers excellent feature extraction capabilities. For power line inspections, VGG's depth helps in capturing intricate details necessary for accurate component classification and fault detection \cite{stefenon_semi_protopnet_2022}. However, its performance may be compromised compared to other state-of-the-art models that came out in recent years \cite{Team}.

\textbf{MobileNet} \cite{howard2017mobilenets} architectures are designed for efficiency, making them ideal for use in mobile and edge computing scenarios. Their streamlined design, based on depthwise separable convolutions \cite{chollet2017xception}, allows for reduced computational load while maintaining high accuracy. In power line inspection, especially those conducted via drones or handheld devices, MobileNet's lightweight nature enables rapid, on-site processing of images for real-time analysis \cite{wei_online_2022, qiu_lightweight_2023, li_improved_2022}.

\textbf{EfficientNet} \cite{tan2019efficientnet} represents a new scaling method for neural networks, which uniformly scales all dimensions of depth, width, and resolution with a set of fixed scaling coefficients. This balanced scaling results in a network that achieves state-of-the-art accuracy with a lower computational cost \cite{Team}. In power line inspection, EfficientNet can be particularly useful for processing high-resolution images effectively, allowing for detailed and accurate identification of line defects and deterioration \cite{odo_aerial_2021, li_pin_2022}.

The above-mentioned object detection and classification networks are often mixed with each other to design powerful networks that often have superior performance to the original networks \cite{tao_detection_2020}. Notably in recent years, the attention mechanism \cite{vaswani2017attention} and its utilization in segmentation and classification algorithms has gained widespread popularity. Its initial application was by the Google DeepMind team in 2014, where they integrated an attention module into an RNN model for image classification \cite{dosovitskiy2020image}. The potency of self-attention networks, particularly in capturing long-distance dependencies and contextual information, has made them a staple in machine vision tasks like image segmentation and classification. Cao et al. \cite{cao_accurate_2023} innovatively used attention-guided multipath features to reconcile the contradictory needs between feature map resolution and the receptive field for high-resolution inputs. Furthermore, the introduction of the attention-RPN module by Fan et al. \cite{fan_2019_few} for small sample target detection, and the combination of global attention and local restructuring by Kong et al. \cite{kong_context_2018}, exemplify the versatility of attention mechanisms. These adaptations enable the collection of task-oriented features across different spatial locations and scales, harnessing both global and local contexts to enhance the accuracy and efficiency of object detection. 

\section{Computer Vision Tasks in Power Line Inspection}\label{appendix:cv_tasks}
\begin{figure}[htb]
    \centering
    \includegraphics[width=1\linewidth]{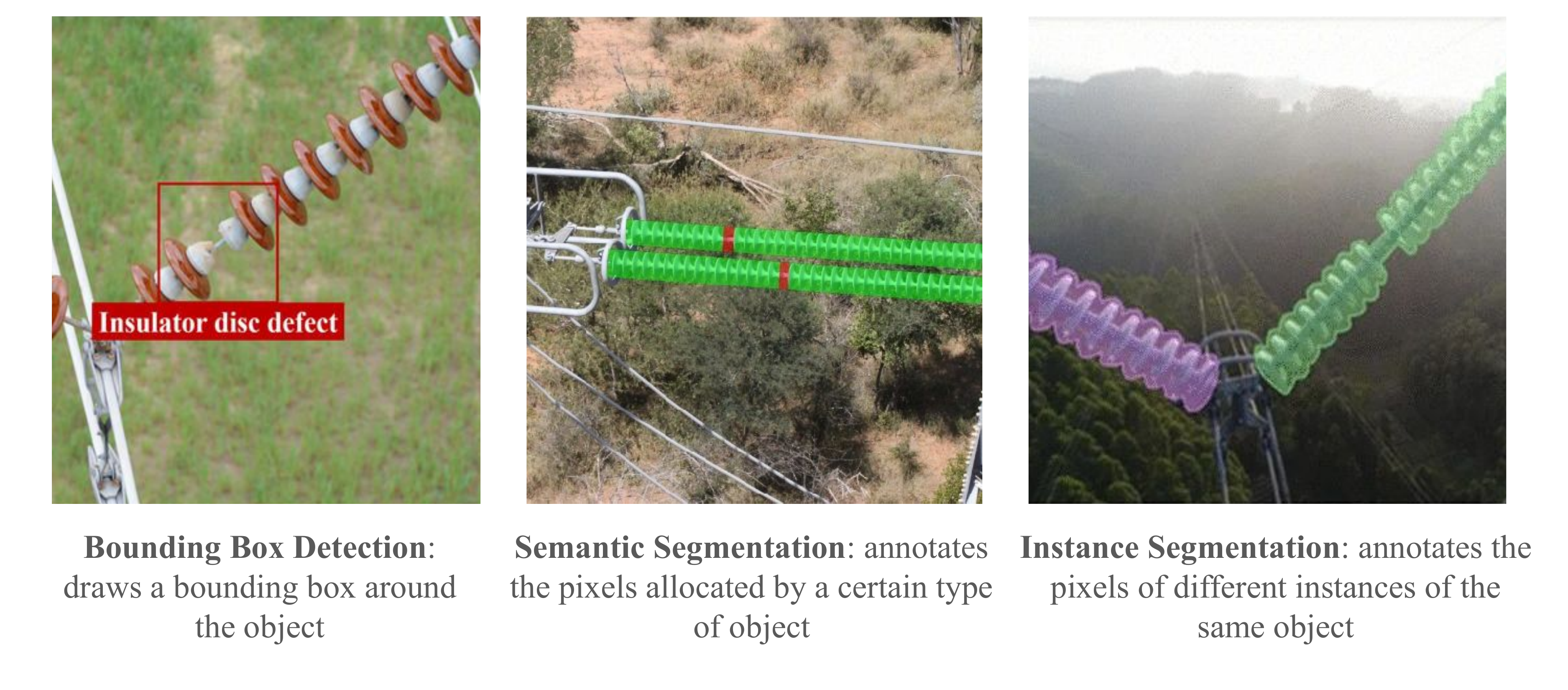}
    \caption{Comparison between different image segmentation techniques \cite{electronics12153210, bob_semantic}.}
    \label{fig:cv_tasks}
\end{figure}

Various object detection techniques are employed in computer vision to automate this task, each with its strengths and application scenarios. This section discusses key methods like bounding box detection, semantic segmentation, and instance segmentation. A visual comparison between these techniques has been shown in Figure \ref{fig:cv_tasks}.

\subsection{Bounding Box Detection}
Bounding Box Detection is a primary object detection method where a box is drawn around each object of interest in an image, marking its location and extent. It is straightforward to implement and computationally less demanding. This method is well-suited for real-time applications due to its relatively fast processing time often reaching over 80 Frames Per Second (FPS) \cite{ge_birds_2022}. Bounding box detection is typically used to identify and locate larger power line components such as towers, insulators, and dampers.

\subsection{Semantic Segmentation}
Semantic segmentation involves the partitioning of an image into segments, where each pixel is classified into a predefined category. This method is capable of producing detailed component-wise masks, which are beneficial for understanding the scene at a pixel level. It provides a precise outline of the components, which is crucial for assessing their condition. Semantic segmentation allows for a comprehensive analysis of the scene by understanding the relationship between different components. This technique is particularly useful for distinguishing between different types of insulators, conductor wires, and vegetation encroachment.

\subsection{Instance Segmentation}
Instance segmentation goes a step beyond semantic segmentation by not only separating the background from the foreground but also differentiating between individual objects of the same class. It is capable of identifying and delineating each instance of multiple objects of the same type. This method excels in scenarios where components are close together or overlapping. Instance segmentation is essential when dealing with dense power line components, such as closely spaced insulators or bundled conductor wires, to assess each component separately.

% }

%% file: main.bbl
\begin{thebibliography}{100}
\expandafter\ifx\csname url\endcsname\relax
  \def\url#1{\texttt{#1}}\fi
\expandafter\ifx\csname urlprefix\endcsname\relax\def\urlprefix{URL }\fi
\expandafter\ifx\csname href\endcsname\relax
  \def\href#1#2{#2} \def\path#1{#1}\fi

\bibitem{mitchell_power_2013}
J.~W. Mitchell, Power line failures and catastrophic wildfires under extreme
  weather conditions, Engineering Failure Analysis 35 (2013) 726--735.
\newblock \href {https://doi.org/10.1016/j.engfailanal.2013.07.006}
  {\path{doi:10.1016/j.engfailanal.2013.07.006}}.

\bibitem{noauthor_link_nodate}
The {Link} {Between} {Power} {Lines} and {Wildfires} (2018).

\bibitem{salim_modeling_2018}
N.~A. Salim, M.~M. Othman, J.~Jasni, I.~Musirin, M.~S. Serwan, Modeling and
  evaluating the customer interruption cost due to dynamic electrical power and
  energy failure, International Journal of Electrical Power \& Energy Systems
  103 (2018) 603--610.
\newblock \href {https://doi.org/10.1016/j.ijepes.2018.06.033}
  {\path{doi:10.1016/j.ijepes.2018.06.033}}.

\bibitem{nguyen_automatic_2018}
V.~N. Nguyen, R.~Jenssen, D.~Roverso, Automatic autonomous vision-based power
  line inspection: {A} review of current status and the potential role of deep
  learning, International Journal of Electrical Power \& Energy Systems 99
  (2018) 107--120.
\newblock \href {https://doi.org/10.1016/j.ijepes.2017.12.016}
  {\path{doi:10.1016/j.ijepes.2017.12.016}}.

\bibitem{liu_two_layer_2019}
Y.~Liu, J.~Shi, Z.~Liu, J.~Huang, T.~Zhou, Two-{Layer} {Routing} for
  {High}-{Voltage} {Powerline} {Inspection} by {Cooperated} {Ground} {Vehicle}
  and {Drone}, Energies 12~(7) (2019) 1385, number: 7 Publisher:
  Multidisciplinary Digital Publishing Institute.
\newblock \href {https://doi.org/10.3390/en12071385}
  {\path{doi:10.3390/en12071385}}.

\bibitem{matikainen_remote_2016}
L.~Matikainen, M.~Lehtomäki, E.~Ahokas, J.~Hyyppä, M.~Karjalainen,
  A.~Jaakkola, A.~Kukko, T.~Heinonen, Remote sensing methods for power line
  corridor surveys, ISPRS Journal of Photogrammetry and Remote Sensing 119
  (2016) 10--31.
\newblock \href {https://doi.org/10.1016/j.isprsjprs.2016.04.011}
  {\path{doi:10.1016/j.isprsjprs.2016.04.011}}.

\bibitem{yang_review_2020}
L.~Yang, J.~Fan, Y.~Liu, E.~Li, J.~Peng, Z.~Liang, A {Review} on
  {State}-of-the-{Art} {Power} {Line} {Inspection} {Techniques}, IEEE
  Transactions on Instrumentation and Measurement 69~(12) (2020) 9350--9365.
\newblock \href {https://doi.org/10.1109/TIM.2020.3031194}
  {\path{doi:10.1109/TIM.2020.3031194}}.

\bibitem{martinez_power_2018}
C.~Martinez, C.~Sampedro, A.~Chauhan, J.~F. Collumeau, P.~Campoy, The {Power}
  {Line} {Inspection} {Software} ({PoLIS}): {A} versatile system for automating
  power line inspection, Engineering Applications of Artificial Intelligence 71
  (2018) 293--314.
\newblock \href {https://doi.org/10.1016/j.engappai.2018.02.008}
  {\path{doi:10.1016/j.engappai.2018.02.008}}.

\bibitem{liu_data_2020}
X.~Liu, X.~Miao, H.~Jiang, J.~Chen, Data analysis in visual power line
  inspection: {An} in-depth review of deep learning for component detection and
  fault diagnosis, Annual Reviews in Control 50 (2020) 253--277.
\newblock \href {https://doi.org/10.1016/j.arcontrol.2020.09.002}
  {\path{doi:10.1016/j.arcontrol.2020.09.002}}.

\bibitem{reddy_condition_2013}
M.~J.~B. Reddy, K.~C. B, D.~K. Mohanta, Condition monitoring of 11 {kV}
  distribution system insulators incorporating complex imagery using combined
  {DOST}-{SVM} approach, IEEE Transactions on Dielectrics and Electrical
  Insulation 20~(2) (2013) 664--674.
\newblock \href {https://doi.org/10.1109/TDEI.2013.6508770}
  {\path{doi:10.1109/TDEI.2013.6508770}}.

\bibitem{zhao_localization_2015}
Z.~Zhao, N.~Liu, L.~Wang, Localization of multiple insulators by orientation
  angle detection and binary shape prior knowledge, IEEE Transactions on
  Dielectrics and Electrical Insulation 22~(6) (2015) 3421--3428.
\newblock \href {https://doi.org/10.1109/TDEI.2015.004741}
  {\path{doi:10.1109/TDEI.2015.004741}}.

\bibitem{wu_texture_2012}
Q.~Wu, J.~An, B.~Lin, A {Texture} {Segmentation} {Algorithm} {Based} on {PCA}
  and {Global} {Minimization} {Active} {Contour} {Model} for {Aerial}
  {Insulator} {Images}, IEEE Journal of Selected Topics in Applied Earth
  Observations and Remote Sensing 5~(5) (2012) 1509--1518.
\newblock \href {https://doi.org/10.1109/JSTARS.2012.2197672}
  {\path{doi:10.1109/JSTARS.2012.2197672}}.

\bibitem{chen2021environment}
M.~Chen, Y.~Tian, S.~Xing, Z.~Li, E.~Li, Z.~Liang, R.~Guo, Environment
  perception technologies for power transmission line inspection robots,
  Journal of Sensors 2021~(1) (2021) 5559231.

\bibitem{yang2020review}
L.~Yang, J.~Fan, Y.~Liu, E.~Li, J.~Peng, Z.~Liang, A review on state-of-the-art
  power line inspection techniques, IEEE Transactions on Instrumentation and
  Measurement 69~(12) (2020) 9350--9365.

\bibitem{sundaram2021deep}
K.~M. Sundaram, A.~Hussain, P.~Sanjeevikumar, J.~B. Holm-Nielsen, V.~K.
  Kaliappan, B.~K. Santhoshi, Deep learning for fault diagnostics in bearings,
  insulators, pv panels, power lines, and electric vehicle applications—the
  state-of-the-art approaches, IEEE Access 9 (2021) 41246--41260.

\bibitem{ruszczak2023overview}
B.~Ruszczak, P.~Michalski, M.~Tomaszewski, Overview of image datasets for deep
  learning applications in diagnostics of power infrastructure, Sensors 23~(16)
  (2023) 7171.

\bibitem{xu2023development}
B.~Xu, Y.~Zhao, T.~Wang, Q.~Chen, Development of power transmission line
  detection technology based on unmanned aerial vehicle image vision, SN
  Applied Sciences 5~(3) (2023) 72.

\bibitem{foudeh2021advanced}
H.~A. Foudeh, P.~C.-K. Luk, J.~F. Whidborne, An advanced unmanned aerial
  vehicle (uav) approach via learning-based control for overhead power line
  monitoring: A comprehensive review, IEEE Access 9 (2021) 130410--130433.

\bibitem{nguyen_intelligent_2019}
V.~N. Nguyen, R.~Jenssen, D.~Roverso, Intelligent {Monitoring} and {Inspection}
  of {Power} {Line} {Components} {Powered} by {UAVs} and {Deep} {Learning},
  IEEE Power and Energy Technology Systems Journal 6~(1) (2019) 11--21.
\newblock \href {https://doi.org/10.1109/JPETS.2018.2881429}
  {\path{doi:10.1109/JPETS.2018.2881429}}.

\bibitem{girshick_rich_2014}
R.~Girshick, J.~Donahue, T.~Darrell, J.~Malik, Rich feature hierarchies for
  accurate object detection and semantic segmentation (Oct. 2014).
\newblock \href {https://doi.org/10.48550/arXiv.1311.2524}
  {\path{doi:10.48550/arXiv.1311.2524}}.

\bibitem{redmon_you_2016}
J.~Redmon, S.~Divvala, R.~Girshick, A.~Farhadi, You {Only} {Look} {Once}:
  {Unified}, {Real}-{Time} {Object} {Detection} (May 2016).
\newblock \href {https://doi.org/10.48550/arXiv.1506.02640}
  {\path{doi:10.48550/arXiv.1506.02640}}.

\bibitem{liu_ssd_2016}
W.~Liu, D.~Anguelov, D.~Erhan, C.~Szegedy, S.~Reed, C.-Y. Fu, A.~C. Berg, Ssd:
  Single shot multibox detector, in: Computer Vision--ECCV 2016: 14th European
  Conference, Amsterdam, The Netherlands, October 11--14, 2016, Proceedings,
  Part I 14, Springer, 2016, pp. 21--37.

\bibitem{zhang_automatic_2017}
Y.~Zhang, X.~Yuan, W.~Li, S.~Chen, Automatic {Power} {Line} {Inspection}
  {Using} {UAV} {Images}, Remote Sensing 9~(8) (2017) 824, number: 8 Publisher:
  Multidisciplinary Digital Publishing Institute.
\newblock \href {https://doi.org/10.3390/rs9080824}
  {\path{doi:10.3390/rs9080824}}.

\bibitem{zhou_insulator_2023}
F.~Zhou, W.~Jin, Z.~Zheng, F.~Mou, Z.~Li, Y.~Ma, B.~Wei, S.~Huang, Q.~Wang,
  Insulator {Detection} for {High}-{Resolution} {Satellite} {Images} {Based} on
  {Deep} {Learning}, IEEE Geoscience and Remote Sensing Letters 20 (2023) 1--5.
\newblock \href {https://doi.org/10.1109/LGRS.2023.3251372}
  {\path{doi:10.1109/LGRS.2023.3251372}}.

\bibitem{alhassan2020power}
A.~B. Alhassan, X.~Zhang, H.~Shen, H.~Xu, Power transmission line inspection
  robots: A review, trends and challenges for future research, International
  Journal of Electrical Power \& Energy Systems 118 (2020) 105862.

\bibitem{ekren2024review}
N.~Ekren, Z.~Karag{\"o}z, M.~{\c{S}}ahin, A review of line suspended inspection
  robots for power transmission lines, Journal of Electrical Engineering \&
  Technology 19~(4) (2024) 2549--2583.

\bibitem{zhang_study_2023}
M.~Zhang, Z.~Song, J.~Yang, M.~Gao, Y.~Hu, C.~Yuan, Z.~Jiang, W.~Cheng, Study
  on the enhancement method of online monitoring image of dense fog environment
  with power lines in smart city, Frontiers in Neurorobotics 16 (2023).

\bibitem{zhang_finet_2022}
Z.-D. Zhang, B.~Zhang, Z.-C. Lan, H.-C. Liu, D.-Y. Li, L.~Pei, W.-X. Yu,
  {FINet}: {An} {Insulator} {Dataset} and {Detection} {Benchmark} {Based} on
  {Synthetic} {Fog} and {Improved} {YOLOv5}, IEEE Transactions on
  Instrumentation and Measurement 71 (2022) 1--8.
\newblock \href {https://doi.org/10.1109/TIM.2022.3194909}
  {\path{doi:10.1109/TIM.2022.3194909}}.

\bibitem{singh_design_2021}
L.~Singh, A.~Alam, K.~V. Kumar, D.~Kumar, P.~Kumar, Z.~A. Jaffery, Design of
  thermal imaging-based health condition monitoring and early fault detection
  technique for porcelain insulators using {Machine} learning, Environmental
  Technology \& Innovation 24 (2021) 102000.
\newblock \href {https://doi.org/10.1016/j.eti.2021.102000}
  {\path{doi:10.1016/j.eti.2021.102000}}.

\bibitem{jaffery_design_2014}
Z.~A. Jaffery, A.~K. Dubey, Design of early fault detection technique for
  electrical assets using infrared thermograms, International Journal of
  Electrical Power \& Energy Systems 63 (2014) 753--759.
\newblock \href {https://doi.org/10.1016/j.ijepes.2014.06.049}
  {\path{doi:10.1016/j.ijepes.2014.06.049}}.

\bibitem{hu_new_2012}
B.~Hu, L.-X. Ma, S.-J. Yuan, B.~Yang, New corona ultraviolet detection system
  and fault location method, in: 2012 {China} {International} {Conference} on
  {Electricity} {Distribution}, 2012, pp. 1--4.
\newblock \href {https://doi.org/10.1109/CICED.2012.6508475}
  {\path{doi:10.1109/CICED.2012.6508475}}.

\bibitem{li_image_2019}
Y.~Li, F.~Yu, Q.~Cai, K.~Yuan, R.~Wan, X.~Li, M.~Qian, P.~Liu, J.~Guo, J.~Yu,
  T.~Zheng, H.~Yan, P.~Hou, Y.~Feng, S.~Wang, L.~Ding, Image fusion of fault
  detection in power system based on deep learning, Cluster Computing 22~(4)
  (2019) 9435--9443.
\newblock \href {https://doi.org/10.1007/s10586-018-2264-2}
  {\path{doi:10.1007/s10586-018-2264-2}}.

\bibitem{zang_status_2008}
C.~Zang, J.~He, Y.~Xiaogen, B.~Chen, H.~Lei, J.~Zhenglong, Z.~Xinjie, Status
  and application foreground of ultraviolet technology on fault detection of
  power devices, in: 2008 {International} {Conference} on {Condition}
  {Monitoring} and {Diagnosis}, 2008, pp. 122--125.
\newblock \href {https://doi.org/10.1109/CMD.2008.4580245}
  {\path{doi:10.1109/CMD.2008.4580245}}.

\bibitem{wang_internal_2023}
F.~Wang, G.~Song, J.~Mao, Y.~Li, Z.~Ji, D.~Chen, A.~Song, Internal {Defect}
  {Detection} of {Overhead} {Aluminum} {Conductor} {Composite} {Core}
  {Transmission} {Lines} {With} an {Inspection} {Robot} and {Computer}
  {Vision}, IEEE Transactions on Instrumentation and Measurement 72 (2023).
\newblock \href {https://doi.org/10.1109/TIM.2023.3265104}
  {\path{doi:10.1109/TIM.2023.3265104}}.

\bibitem{guan2021uav}
H.~Guan, X.~Sun, Y.~Su, T.~Hu, H.~Wang, H.~Wang, C.~Peng, Q.~Guo, Uav-lidar
  aids automatic intelligent powerline inspection, International Journal of
  Electrical Power \& Energy Systems 130 (2021) 106987.

\bibitem{bergmann2024approach}
M.~A. Bergmann, L.~F.~R. Moreira, B.~Krohling, T.~L. Silveira, C.~R. Jung,
  J.~Tang, M.~V. Feitosa, R.~L.~B. Gomes, B.~N. Soares, An approach based on
  lidar and spherical images for automated vegetation inspection in urban power
  distribution lines, IEEE Access (2024).

\bibitem{tao2018detection}
X.~Tao, D.~Zhang, Z.~Wang, X.~Liu, H.~Zhang, D.~Xu, Detection of power line
  insulator defects using aerial images analyzed with convolutional neural
  networks, IEEE transactions on systems, man, and cybernetics: systems 50~(4)
  (2018) 1486--1498.

\bibitem{voigt2017eu}
P.~Voigt, A.~Von~dem Bussche, The eu general data protection regulation (gdpr),
  A Practical Guide, 1st Ed., Cham: Springer International Publishing
  10~(3152676) (2017) 10--5555.

\bibitem{ccpa}
\href{https://oag.ca.gov/privacy/ccpa}{California consumer privacy act (ccpa)}
  (2024).
\newline\urlprefix\url{https://oag.ca.gov/privacy/ccpa}

\bibitem{Yetgin_2019}
Yetgin, \href{https://data.mendeley.com/datasets/n6wrv4ry6v/8}{Powerline image
  dataset (infrared-ir and visible light-vl)} (2019).
\newline\urlprefix\url{https://data.mendeley.com/datasets/n6wrv4ry6v/8}

\bibitem{abdelfattah2020ttpla}
R.~Abdelfattah, X.~Wang, S.~Wang, Ttpla: An aerial-image dataset for detection
  and segmentation of transmission towers and power lines, in: Proceedings of
  the Asian Conference on Computer Vision, 2020, pp. 601--618.

\bibitem{t9qk_cn48_21}
L.~Diniz, T.~Santa~Maria, G.~A. Pussente,
  \href{https://dx.doi.org/10.21227/t9qk-cn48}{Power transmission line dataset}
  (2021).
\newblock \href {https://doi.org/10.21227/t9qk-cn48}
  {\path{doi:10.21227/t9qk-cn48}}.
\newline\urlprefix\url{https://dx.doi.org/10.21227/t9qk-cn48}

\bibitem{recognizance_2}
T.~Diwakar, \href{https://kaggle.com/competitions/recognizance-2}{(
  recognizance - 2 ) power lines detection} (2021).
\newline\urlprefix\url{https://kaggle.com/competitions/recognizance-2}

\bibitem{9643100}
A.~L.~B. Vieira-e Silva, H.~de~Castro~Felix, T.~de~Menezes~Chaves, F.~P.~M.
  Simões, V.~Teichrieb, M.~M. dos Santos, H.~da~Cunha~Santiago, V.~A.~C.
  Sgotti, H.~B. D. T.~L. Neto, Stn plad: A dataset for multi-size power line
  assets detection in high-resolution uav images, in: 2021 34th SIBGRAPI
  Conference on Graphics, Patterns and Images (SIBGRAPI), 2021, pp. 215--222.
\newblock \href {https://doi.org/10.1109/SIBGRAPI54419.2021.00037}
  {\path{doi:10.1109/SIBGRAPI54419.2021.00037}}.

\bibitem{antonis_savva_2023_7781388}
A.~Savva, R.~Makrigiorgis, P.~Kolios, C.~Kyrkou,
  \href{https://doi.org/10.5281/zenodo.7781388}{Aerial power infrastructure
  detection dataset} (2023).
\newblock \href {https://doi.org/10.5281/zenodo.7781388}
  {\path{doi:10.5281/zenodo.7781388}}.
\newline\urlprefix\url{https://doi.org/10.5281/zenodo.7781388}

\bibitem{drones7020125}
F.~Shuang, S.~Han, Y.~Li, T.~Lu,
  \href{https://www.mdpi.com/2504-446X/7/2/125}{Rsin-dataset: An uav-based
  insulator detection aerial images dataset and benchmark}, Drones 7~(2)
  (2023).
\newblock \href {https://doi.org/10.3390/drones7020125}
  {\path{doi:10.3390/drones7020125}}.
\newline\urlprefix\url{https://www.mdpi.com/2504-446X/7/2/125}

\bibitem{sundaram_deep_2021}
K.~M. Sundaram, A.~Hussain, P.~Sanjeevikumar, J.~B. Holm-Nielsen, V.~K.
  Kaliappan, B.~K. Santhoshi, Deep {Learning} for {Fault} {Diagnostics} in
  {Bearings}, {Insulators}, {PV} {Panels}, {Power} {Lines}, and {Electric}
  {Vehicle} {Applications}—{The} {State}-of-the-{Art} {Approaches}, IEEE
  Access 9 (2021) 41246--41260.
\newblock \href {https://doi.org/10.1109/ACCESS.2021.3064360}
  {\path{doi:10.1109/ACCESS.2021.3064360}}.

\bibitem{redmon_yolo9000_2016}
J.~Redmon, A.~Farhadi, {YOLO9000}: {Better}, {Faster}, {Stronger} (Dec. 2016).
\newblock \href {https://doi.org/10.48550/arXiv.1612.08242}
  {\path{doi:10.48550/arXiv.1612.08242}}.

\bibitem{redmon_yolov3_2018}
J.~Redmon, A.~Farhadi, {YOLOv3}: {An} {Incremental} {Improvement} (Apr. 2018).
\newblock \href {https://doi.org/10.48550/arXiv.1804.02767}
  {\path{doi:10.48550/arXiv.1804.02767}}.

\bibitem{bochkovskiy_yolov4_2020}
A.~Bochkovskiy, C.-Y. Wang, H.-Y.~M. Liao, {YOLOv4}: {Optimal} {Speed} and
  {Accuracy} of {Object} {Detection} (Apr. 2020).
\newblock \href {https://doi.org/10.48550/arXiv.2004.10934}
  {\path{doi:10.48550/arXiv.2004.10934}}.

\bibitem{jocher_yolo_2023}
G.~Jocher, A.~Chaurasia, J.~Qiu, {YOLO} by {Ultralytics} (Jan. 2023).

\bibitem{girshick_fast_2015}
R.~Girshick, Fast {R}-{CNN} (Sep. 2015).
\newblock \href {https://doi.org/10.48550/arXiv.1504.08083}
  {\path{doi:10.48550/arXiv.1504.08083}}.

\bibitem{ren_faster_2016}
S.~Ren, K.~He, R.~Girshick, J.~Sun, Faster {R}-{CNN}: {Towards} {Real}-{Time}
  {Object} {Detection} with {Region} {Proposal} {Networks} (Jan. 2016).
\newblock \href {https://doi.org/10.48550/arXiv.1506.01497}
  {\path{doi:10.48550/arXiv.1506.01497}}.

\bibitem{dosovitskiy2020image}
A.~Dosovitskiy, L.~Beyer, A.~Kolesnikov, D.~Weissenborn, X.~Zhai,
  T.~Unterthiner, M.~Dehghani, M.~Minderer, G.~Heigold, S.~Gelly, et~al., An
  image is worth 16x16 words: Transformers for image recognition at scale,
  arXiv preprint arXiv:2010.11929 (2020).

\bibitem{liu2021swintransformerhierarchicalvision}
Z.~Liu, Y.~Lin, Y.~Cao, H.~Hu, Y.~Wei, Z.~Zhang, S.~Lin, B.~Guo,
  \href{https://arxiv.org/abs/2103.14030}{Swin transformer: Hierarchical vision
  transformer using shifted windows} (2021).
\newblock \href {http://arxiv.org/abs/2103.14030} {\path{arXiv:2103.14030}}.
\newline\urlprefix\url{https://arxiv.org/abs/2103.14030}

\bibitem{carion2020end}
N.~Carion, F.~Massa, G.~Synnaeve, N.~Usunier, A.~Kirillov, S.~Zagoruyko,
  End-to-end object detection with transformers, in: European conference on
  computer vision, Springer, 2020, pp. 213--229.

\bibitem{sadykova2019yolo}
D.~Sadykova, D.~Pernebayeva, M.~Bagheri, A.~James, In-yolo: Real-time detection
  of outdoor high voltage insulators using uav imaging, IEEE Transactions on
  Power Delivery 35~(3) (2019) 1599--1601.

\bibitem{singh_2023_interpretable}
G.~Singh, S.~Stefenon, K.-C. Yow, Interpretable visual transmission lines
  inspections using pseudo-prototypical part network, Machine Vision and
  Applications 34~(3) (2023).
\newblock \href {https://doi.org/10.1007/s00138-023-01390-6}
  {\path{doi:10.1007/s00138-023-01390-6}}.

\bibitem{zhang_cloud_edge_2020}
S.~Zhang, J.~Wang, J.~Tong, J.~Zhang, M.~Zhang, Cloud-{Edge} {Fusion} {Based}
  {Abnormal} {Object} {Detection} of {Power} {Transmission} {Lines} {Using}
  {Incremental} {Learning}, IEEE Access 8 (2020) 218694--218701.
\newblock \href {https://doi.org/10.1109/ACCESS.2020.3037172}
  {\path{doi:10.1109/ACCESS.2020.3037172}}.

\bibitem{wei_online_2022}
B.~Wei, Z.~Xie, Y.~Liu, K.~Wen, F.~Deng, P.~Zhang, Online {Monitoring} {Method}
  for {Insulator} {Self}-explosion {Based} on {Edge} {Computing} and {Deep}
  {Learning}, CSEE Journal of Power and Energy Systems 8~(6) (2022) 1684--1696.
\newblock \href {https://doi.org/10.17775/CSEEJPES.2020.05910}
  {\path{doi:10.17775/CSEEJPES.2020.05910}}.

\bibitem{zhai_hybrid_2021}
Y.~Zhai, X.~Yang, Q.~Wang, Z.~Zhao, W.~Zhao, Hybrid {Knowledge} {R}-{CNN} for
  {Transmission} {Line} {Multifitting} {Detection}, IEEE Transactions on
  Instrumentation and Measurement 70 (2021) 1--12.
\newblock \href {https://doi.org/10.1109/TIM.2021.3096600}
  {\path{doi:10.1109/TIM.2021.3096600}}.

\bibitem{rong_intelligent_2021}
S.~Rong, L.~He, L.~Du, Z.~Li, S.~Yu, Intelligent {Detection} of {Vegetation}
  {Encroachment} of {Power} {Lines} {With} {Advanced} {Stereovision}, IEEE
  Transactions on Power Delivery 36~(6) (2021) 3477--3485.
\newblock \href {https://doi.org/10.1109/TPWRD.2020.3043433}
  {\path{doi:10.1109/TPWRD.2020.3043433}}.

\bibitem{miao_insulator_2019}
X.~Miao, X.~Liu, J.~Chen, S.~Zhuang, J.~Fan, H.~Jiang, Insulator {Detection} in
  {Aerial} {Images} for {Transmission} {Line} {Inspection} {Using} {Single}
  {Shot} {Multibox} {Detector}, IEEE Access 7 (2019) 9945--9956.
\newblock \href {https://doi.org/10.1109/ACCESS.2019.2891123}
  {\path{doi:10.1109/ACCESS.2019.2891123}}.

\bibitem{dong_improved_2023}
K.~Dong, Q.~Shen, C.~Wang, Y.~Dong, Q.~Liu, Z.~Lu, Z.~Lu, Improved swin
  transformer-based defect detection method for transmission line patrol
  inspection images, Evolutionary Intelligence (2023).
\newblock \href {https://doi.org/10.1007/s12065-023-00837-z}
  {\path{doi:10.1007/s12065-023-00837-z}}.

\bibitem{zhang_pa_detr_2023}
K.~Zhang, W.~Lou, J.~Wang, R.~Zhou, X.~Guo, Y.~Xiao, C.~Shi, Z.~Zhao,
  {PA}-{DETR}: {End}-to-{End} {Visually} {Indistinguishable} {Bolt} {Defects}
  {Detection} {Method} {Based} on {Transmission} {Line} {Knowledge}
  {Reasoning}, IEEE Transactions on Instrumentation and Measurement 72 (2023)
  1--14.
\newblock \href {https://doi.org/10.1109/TIM.2023.3282302}
  {\path{doi:10.1109/TIM.2023.3282302}}.

\bibitem{jain2024transfer}
N.~Jain, J.~Bedi, A.~Anand, S.~Godara, A transfer learning architecture to
  detect faulty insulators in powerlines, IEEE Transactions on Power Delivery
  (2024).

\bibitem{li_improved_2022}
H.~Li, L.~Liu, J.~Du, F.~Jiang, F.~Guo, Q.~Hu, L.~Fan, An {Improved} {YOLOv3}
  for {Foreign} {Objects} {Detection} of {Transmission} {Lines}, IEEE Access 10
  (2022) 45620--45628.
\newblock \href {https://doi.org/10.1109/ACCESS.2022.3170696}
  {\path{doi:10.1109/ACCESS.2022.3170696}}.

\bibitem{bharati2020deep}
P.~Bharati, A.~Pramanik, Deep learning techniques—r-cnn to mask r-cnn: a
  survey, Computational Intelligence in Pattern Recognition: Proceedings of
  CIPR 2019 (2020) 657--668.

\bibitem{huang2017speed}
J.~Huang, V.~Rathod, C.~Sun, M.~Zhu, A.~Korattikara, A.~Fathi, I.~Fischer,
  Z.~Wojna, Y.~Song, S.~Guadarrama, et~al., Speed/accuracy trade-offs for
  modern convolutional object detectors, in: Proceedings of the IEEE conference
  on computer vision and pattern recognition, 2017, pp. 7310--7311.

\bibitem{han2022survey}
K.~Han, Y.~Wang, H.~Chen, X.~Chen, J.~Guo, Z.~Liu, Y.~Tang, A.~Xiao, C.~Xu,
  Y.~Xu, et~al., A survey on vision transformer, IEEE transactions on pattern
  analysis and machine intelligence 45~(1) (2022) 87--110.

\bibitem{5206848}
J.~Deng, W.~Dong, R.~Socher, L.-J. Li, K.~Li, L.~Fei-Fei, Imagenet: A
  large-scale hierarchical image database, in: 2009 IEEE Conference on Computer
  Vision and Pattern Recognition, 2009, pp. 248--255.
\newblock \href {https://doi.org/10.1109/CVPR.2009.5206848}
  {\path{doi:10.1109/CVPR.2009.5206848}}.

\bibitem{he_2023_deep}
K.~He, X.~Zhang, S.~Ren, J.~Sun, \href{http://arxiv.org/abs/1512.03385}{Deep
  residual learning for image recognition}, CoRR abs/1512.03385 (2015).
\newblock \href {http://arxiv.org/abs/1512.03385} {\path{arXiv:1512.03385}}.
\newline\urlprefix\url{http://arxiv.org/abs/1512.03385}

\bibitem{simonyan2014very}
K.~Simonyan, A.~Zisserman, Very deep convolutional networks for large-scale
  image recognition, arXiv preprint arXiv:1409.1556 (2014).

\bibitem{howard2017mobilenets}
A.~G. Howard, M.~Zhu, B.~Chen, D.~Kalenichenko, W.~Wang, T.~Weyand,
  M.~Andreetto, H.~Adam, Mobilenets: Efficient convolutional neural networks
  for mobile vision applications, arXiv preprint arXiv:1704.04861 (2017).

\bibitem{tan2019efficientnet}
M.~Tan, Q.~Le, Efficientnet: Rethinking model scaling for convolutional neural
  networks, in: International conference on machine learning, PMLR, 2019, pp.
  6105--6114.

\bibitem{cao_accurate_2023}
Y.~Cao, H.~Xu, C.~Su, Q.~Yang, Accurate {Glass} {Insulators} {Defect}
  {Detection} in {Power} {Transmission} {Grids} {Using} {Aerial} {Image}
  {Augmentation}, IEEE Transactions on Power Delivery 38~(2) (2023) 956--965.
\newblock \href {https://doi.org/10.1109/TPWRD.2022.3202958}
  {\path{doi:10.1109/TPWRD.2022.3202958}}.

\bibitem{luo_ultrasmall_2023}
P.~Luo, B.~Wang, H.~Wang, F.~Ma, H.~Ma, L.~Wang, An {Ultrasmall} {Bolt}
  {Defect} {Detection} {Method} for {Transmission} {Line} {Inspection}, IEEE
  Transactions on Instrumentation and Measurement 72 (2023) 1--12.
\newblock \href {https://doi.org/10.1109/TIM.2023.3241994}
  {\path{doi:10.1109/TIM.2023.3241994}}.

\bibitem{stefenon_semi_protopnet_2022}
S.~F. Stefenon, G.~Singh, K.-C. Yow, A.~Cimatti, Semi-{ProtoPNet} {Deep}
  {Neural} {Network} for the {Classification} of {Defective} {Power} {Grid}
  {Distribution} {Structures}, Sensors 22~(13) (2022) 4859, number: 13
  Publisher: Multidisciplinary Digital Publishing Institute.
\newblock \href {https://doi.org/10.3390/s22134859}
  {\path{doi:10.3390/s22134859}}.

\bibitem{qiu_lightweight_2023}
Z.~Qiu, X.~Zhu, C.~Liao, W.~Qu, Y.~Yu, A {Lightweight} {YOLOv4}-{EDAM} {Model}
  for {Accurate} and {Real}-time {Detection} of {Foreign} {Objects} {Suspended}
  on {Power} {Lines}, IEEE Transactions on Power Delivery 38~(2) (2023)
  1329--1340.
\newblock \href {https://doi.org/10.1109/TPWRD.2022.3213598}
  {\path{doi:10.1109/TPWRD.2022.3213598}}.

\bibitem{odo_aerial_2021}
A.~Odo, S.~McKenna, D.~Flynn, J.~B. Vorstius, Aerial {Image} {Analysis} {Using}
  {Deep} {Learning} for {Electrical} {Overhead} {Line} {Network} {Asset}
  {Management}, IEEE Access 9 (2021) 146281--146295.
\newblock \href {https://doi.org/10.1109/ACCESS.2021.3123158}
  {\path{doi:10.1109/ACCESS.2021.3123158}}.

\bibitem{li_pin_2022}
Y.~Li, Z.~Li, Y.~Liu, G.~Sheng, X.~Jiang, Pin {Bolt} {State} {Identification}
  {Using} {Cascaded} {Object} {Detection} {Networks}, Frontiers in Energy
  Research 10 (2022).
\newblock \href {https://doi.org/10.3389/fenrg.2022.813945}
  {\path{doi:10.3389/fenrg.2022.813945}}.

\bibitem{vaswani2017attention}
A.~Vaswani, N.~Shazeer, N.~Parmar, J.~Uszkoreit, L.~Jones, A.~N. Gomez,
  {\L}.~Kaiser, I.~Polosukhin, Attention is all you need, Advances in neural
  information processing systems 30 (2017).

\bibitem{fan_2019_few}
Q.~Fan, W.~Zhuo, Y.~Tai, \href{http://arxiv.org/abs/1908.01998}{Few-shot object
  detection with attention-rpn and multi-relation detector}, CoRR
  abs/1908.01998 (2019).
\newblock \href {http://arxiv.org/abs/1908.01998} {\path{arXiv:1908.01998}}.
\newline\urlprefix\url{http://arxiv.org/abs/1908.01998}

\bibitem{kong_context_2018}
L.~Kong, X.~Zhu, G.~Wang, Context semantics for small target detection in
  large-field images with two cascaded faster r-cnns, in: Journal of Physics:
  Conference Series, Vol. 1069, IOP Publishing, 2018, p. 012138.
\newblock \href {https://doi.org/10.1088/1742-6596/1069/1/012138}
  {\path{doi:10.1088/1742-6596/1069/1/012138}}.

\bibitem{ge_birds_2022}
Z.~Ge, H.~Li, R.~Yang, H.~Liu, S.~Pei, Z.~Jia, Z.~Ma, Bird’s {Nest}
  {Detection} {Algorithm} for {Transmission} {Lines} {Based} on {Deep}
  {Learning}, in: 2022 3rd {International} {Conference} on {Computer} {Vision},
  {Image} and {Deep} {Learning} \& {International} {Conference} on {Computer}
  {Engineering} and {Applications} ({CVIDL} \& {ICCEA}), 2022, pp. 417--420.
\newblock \href {https://doi.org/10.1109/CVIDLICCEA56201.2022.9824057}
  {\path{doi:10.1109/CVIDLICCEA56201.2022.9824057}}.

\bibitem{electronics12153210}
Y.~Chen, H.~Liu, J.~Chen, J.~Hu, E.~Zheng, Insu-yolo: An insulator defect
  detection algorithm based on multiscale feature fusion, Electronics 12~(15)
  (2023).
\newblock \href {https://doi.org/10.3390/electronics12153210}
  {\path{doi:10.3390/electronics12153210}}.

\bibitem{bob_semantic}
M.~Oberweger, A.~Wendel, H.~Bischof, Visual recognition and fault detection for
  power line insulators, in: 19th computer vision winter workshop, 2014, pp.
  1--8.

\bibitem{kang_deep_2019}
G.~Kang, S.~Gao, L.~Yu, D.~Zhang, Deep {Architecture} for {High}-{Speed}
  {Railway} {Insulator} {Surface} {Defect} {Detection}: {Denoising}
  {Autoencoder} {With} {Multitask} {Learning}, IEEE Transactions on
  Instrumentation and Measurement 68~(8) (2019) 2679--2690.
\newblock \href {https://doi.org/10.1109/TIM.2018.2868490}
  {\path{doi:10.1109/TIM.2018.2868490}}.

\bibitem{zhao2020image}
Z.~Zhao, H.~Qi, X.~Fan, G.~Xu, Y.~Qi, Y.~Zhai, K.~Zhang, Image representation
  method based on relative layer entropy for insulator recognition, Entropy
  22~(4) (2020) 419.

\bibitem{zhang_defgan_2021}
D.~Zhang, S.~Gao, L.~Yu, G.~Kang, X.~Wei, D.~Zhan, {DefGAN}: {Defect}
  {Detection} {GANs} {With} {Latent} {Space} {Pitting} for {High}-{Speed}
  {Railway} {Insulator}, IEEE Transactions on Instrumentation and Measurement
  70 (2021) 1--10.
\newblock \href {https://doi.org/10.1109/TIM.2020.3038008}
  {\path{doi:10.1109/TIM.2020.3038008}}.

\bibitem{waleed_drone_based_2021}
D.~Waleed, S.~Mukhopadhyay, U.~Tariq, A.~H. El-Hag, Drone-{Based} {Ceramic}
  {Insulators} {Condition} {Monitoring}, IEEE Transactions on Instrumentation
  and Measurement 70 (2021) 1--12.
\newblock \href {https://doi.org/10.1109/TIM.2021.3078538}
  {\path{doi:10.1109/TIM.2021.3078538}}.

\bibitem{antwi_bekoe_deep_2022}
E.~Antwi-Bekoe, G.~Liu, J.-P. Ainam, G.~Sun, X.~Xie, A deep learning approach
  for insulator instance segmentation and defect detection, Neural Computing
  and Applications 34~(9) (2022) 7253--7269.
\newblock \href {https://doi.org/10.1007/s00521-021-06792-z}
  {\path{doi:10.1007/s00521-021-06792-z}}.

\bibitem{shuang_rsin_dataset_2023}
F.~Shuang, S.~Han, Y.~Li, T.~Lu, {RSIn}-{Dataset}: {An} {UAV}-{Based}
  {Insulator} {Detection} {Aerial} {Images} {Dataset} and {Benchmark}, Drones
  7~(2) (2023) 125, number: 2 Publisher: Multidisciplinary Digital Publishing
  Institute.
\newblock \href {https://doi.org/10.3390/drones7020125}
  {\path{doi:10.3390/drones7020125}}.

\bibitem{zhai_multi_fitting_2022}
Y.~Zhai, Q.~Wang, X.~Yang, Z.~Zhao, W.~Zhao, Multi-{Fitting} {Detection} on
  {Transmission} {Line} {Based} on {Cascade} {Reasoning} {Graph} {Network},
  IEEE Transactions on Power Delivery 37~(6) (2022) 4858--4868.
\newblock \href {https://doi.org/10.1109/TPWRD.2022.3161124}
  {\path{doi:10.1109/TPWRD.2022.3161124}}.

\bibitem{zhang_attention_guided_2022}
H.~Zhang, L.~Wu, Y.~Chen, R.~Chen, S.~Kong, Y.~Wang, J.~Hu, J.~Wu,
  Attention-{Guided} {Multitask} {Convolutional} {Neural} {Network} for {Power}
  {Line} {Parts} {Detection}, IEEE Transactions on Instrumentation and
  Measurement 71 (2022) 1--13.
\newblock \href {https://doi.org/10.1109/TIM.2022.3162615}
  {\path{doi:10.1109/TIM.2022.3162615}}.

\bibitem{huang_structural_2023}
X.~Huang, Y.~Wu, Y.~Zhang, B.~Li, Structural {Defect} {Detection} {Technology}
  of {Transmission} {Line} {Damper} {Based} on {UAV} {Image}, IEEE Transactions
  on Instrumentation and Measurement 72 (2023) 1--14.
\newblock \href {https://doi.org/10.1109/TIM.2022.3228008}
  {\path{doi:10.1109/TIM.2022.3228008}}.

\bibitem{yang_vision_based_2022}
L.~Yang, J.~Fan, S.~Xu, E.~Li, Y.~Liu, Vision-{Based} {Power} {Line}
  {Segmentation} {With} an {Attention} {Fusion} {Network}, IEEE Sensors Journal
  22~(8) (2022) 8196--8205.
\newblock \href {https://doi.org/10.1109/JSEN.2022.3157336}
  {\path{doi:10.1109/JSEN.2022.3157336}}.

\bibitem{yang_dra_net_2023}
L.~Yang, S.~Kong, J.~Deng, H.~Li, Y.~Liu, {DRA}-{Net}: {A} {Dual}-{Branch}
  {Residual} {Attention} {Network} for {Pixelwise} {Power} {Line} {Detection},
  IEEE Transactions on Instrumentation and Measurement 72 (2023) 1--13.
\newblock \href {https://doi.org/10.1109/TIM.2023.3259047}
  {\path{doi:10.1109/TIM.2023.3259047}}.

\bibitem{zhang_multi_scale_2020}
P.~Zhang, Z.~Zhang, Y.~Hao, Z.~Zhou, B.~Luo, T.~Wang, Multi-{Scale} {Feature}
  {Enhanced} {Domain} {Adaptive} {Object} {Detection} {For} {Power}
  {Transmission} {Line} {Inspection}, IEEE Access 8 (2020) 182105--182116.
\newblock \href {https://doi.org/10.1109/ACCESS.2020.3027850}
  {\path{doi:10.1109/ACCESS.2020.3027850}}.

\bibitem{wang_image_2019}
Y.~Wang, Q.~Li, B.~Chen, Image classification towards transmission line fault
  detection via learning deep quality-aware fine-grained categorization,
  Journal of Visual Communication and Image Representation 64 (2019).
\newblock \href {https://doi.org/10.1016/j.jvcir.2019.102647}
  {\path{doi:10.1016/j.jvcir.2019.102647}}.

\bibitem{dong2024transmission}
C.~Dong, K.~Zhang, Z.~Xie, J.~Wang, X.~Guo, C.~Shi, Y.~Xiao, Transmission line
  key components and defects detection based on meta-learning, IEEE
  Transactions on Instrumentation and Measurement (2024).

\bibitem{liu2023tower}
X.~Liu, X.~Miao, H.~Jiang, J.~Chen, M.~Wu, Z.~Chen, Tower masking mim: A
  self-supervised pretraining method for power line inspection, IEEE
  Transactions on Industrial Informatics 20~(1) (2023) 513--523.

\bibitem{chen_research_2019}
H.~Chen, Z.~He, B.~Shi, T.~Zhong, Research on {Recognition} {Method} of
  {Electrical} {Components} {Based} on {YOLO} {V3}, IEEE Access 7 (2019)
  157818--157829.
\newblock \href {https://doi.org/10.1109/ACCESS.2019.2950053}
  {\path{doi:10.1109/ACCESS.2019.2950053}}.

\bibitem{shi2024lskf}
C.~Shi, X.~Zheng, Z.~Zhao, K.~Zhang, Z.~Su, Q.~Lu, Lskf-yolo: Large selective
  kernel feature fusion network for power tower detection in high-resolution
  satellite remote sensing images, IEEE Transactions on Geoscience and Remote
  Sensing (2024).

\bibitem{liu_discrimination_2017}
Y.~Liu, S.~Pei, W.~Fu, K.~Zhang, X.~Ji, Z.~Yin, The discrimination method as
  applied to a deteriorated porcelain insulator used in transmission lines on
  the basis of a convolution neural network, IEEE Transactions on Dielectrics
  and Electrical Insulation 24~(6) (2017) 3559--3566.
\newblock \href {https://doi.org/10.1109/TDEI.2017.006840}
  {\path{doi:10.1109/TDEI.2017.006840}}.

\bibitem{ibrahim_application_2020}
A.~Ibrahim, A.~Dalbah, A.~Abualsaud, U.~Tariq, A.~El-Hag, Application of
  {Machine} {Learning} to {Evaluate} {Insulator} {Surface} {Erosion}, IEEE
  Transactions on Instrumentation and Measurement 69~(2) (2020) 314--316.
\newblock \href {https://doi.org/10.1109/TIM.2019.2956300}
  {\path{doi:10.1109/TIM.2019.2956300}}.

\bibitem{mussina_multi_modal_2020}
D.~Mussina, A.~Irmanova, P.~K. Jamwal, M.~Bagheri, Multi-{Modal} {Data}
  {Fusion} {Using} {Deep} {Neural} {Network} for {Condition} {Monitoring} of
  {High} {Voltage} {Insulator}, IEEE Access 8 (2020) 184486--184496.
\newblock \href {https://doi.org/10.1109/ACCESS.2020.3027825}
  {\path{doi:10.1109/ACCESS.2020.3027825}}.

\bibitem{stefenon_classification_2022}
S.~F. Stefenon, K.-C. Yow, A.~Nied, L.~H. Meyer, Classification of distribution
  power grid structures using inception v3 deep neural network, Electrical
  Engineering 104~(6) (2022) 4557--4569.
\newblock \href {https://doi.org/10.1007/s00202-022-01641-1}
  {\path{doi:10.1007/s00202-022-01641-1}}.

\bibitem{roy_accurate_2023}
S.~S. Roy, A.~Paramane, J.~Singh, S.~Chatterjee, A.~K. Das, Accurate {Sensing}
  of {Insulator} {Surface} {Contamination} {Using} {Customized} {Convolutional}
  {Neural} {Network}, IEEE Sensors Letters 7~(1) (2023) 1--4.
\newblock \href {https://doi.org/10.1109/LSENS.2022.3232506}
  {\path{doi:10.1109/LSENS.2022.3232506}}.

\bibitem{wang_detection_2020}
S.~Wang, Y.~Liu, Y.~Qing, C.~Wang, T.~Lan, R.~Yao, Detection of insulator
  defects with improved {ResNeSt} and region proposal network, IEEE Access 8
  (2020) 184841--184850.
\newblock \href {https://doi.org/10.1109/ACCESS.2020.3029857}
  {\path{doi:10.1109/ACCESS.2020.3029857}}.

\bibitem{fu_small_sized_2023}
Q.~Fu, J.~Liu, X.~Zhang, Y.~Zhang, Y.~Ou, R.~Jiao, C.~Li, G.~Mazzanti, A
  {Small}-{Sized} {Defect} {Detection} {Method} for {Overhead} {Transmission}
  {Lines} {Based} on {Convolutional} {Neural} {Networks}, IEEE Transactions on
  Instrumentation and Measurement 72 (2023) 1--12.
\newblock \href {https://doi.org/10.1109/TIM.2023.3298424}
  {\path{doi:10.1109/TIM.2023.3298424}}.

\bibitem{liu_box_point_2021}
X.~Liu, X.~Miao, H.~Jiang, J.~Chen, Box-{Point} {Detector}: {A} {Diagnosis}
  {Method} for {Insulator} {Faults} in {Power} {Lines} {Using} {Aerial}
  {Images} and {Convolutional} {Neural} {Networks}, IEEE Transactions on Power
  Delivery 36~(6) (2021) 3765--3773.
\newblock \href {https://doi.org/10.1109/TPWRD.2020.3048935}
  {\path{doi:10.1109/TPWRD.2020.3048935}}.

\bibitem{jiang_insulator_2019}
H.~Jiang, X.~Qiu, J.~Chen, X.~Liu, X.~Miao, S.~Zhuang, Insulator {Fault}
  {Detection} in {Aerial} {Images} {Based} on {Ensemble} {Learning} {With}
  {Multi}-{Level} {Perception}, IEEE Access 7 (2019) 61797--61810.
\newblock \href {https://doi.org/10.1109/ACCESS.2019.2915985}
  {\path{doi:10.1109/ACCESS.2019.2915985}}.

\bibitem{tao_detection_2020}
X.~Tao, D.~Zhang, Z.~Wang, X.~Liu, H.~Zhang, D.~Xu, Detection of {Power} {Line}
  {Insulator} {Defects} {Using} {Aerial} {Images} {Analyzed} {With}
  {Convolutional} {Neural} {Networks}, IEEE Transactions on Systems, Man, and
  Cybernetics: Systems 50~(4) (2020) 1486--1498.
\newblock \href {https://doi.org/10.1109/TSMC.2018.2871750}
  {\path{doi:10.1109/TSMC.2018.2871750}}.

\bibitem{zhang_insudet_2021}
X.~Zhang, Y.~Zhang, J.~Liu, C.~Zhang, X.~Xue, H.~Zhang, W.~Zhang, {InsuDet}:
  {A} {Fault} {Detection} {Method} for {Insulators} of {Overhead}
  {Transmission} {Lines} {Using} {Convolutional} {Neural} {Networks}, IEEE
  Transactions on Instrumentation and Measurement 70 (2021) 1--12.
\newblock \href {https://doi.org/10.1109/TIM.2021.3120796}
  {\path{doi:10.1109/TIM.2021.3120796}}.

\bibitem{hao_insulator_2022}
K.~Hao, G.~Chen, L.~Zhao, Z.~Li, Y.~Liu, C.~Wang, An {Insulator} {Defect}
  {Detection} {Model} in {Aerial} {Images} {Based} on {Multiscale} {Feature}
  {Pyramid} {Network}, IEEE Transactions on Instrumentation and Measurement 71
  (2022) 1--12.
\newblock \href {https://doi.org/10.1109/TIM.2022.3200861}
  {\path{doi:10.1109/TIM.2022.3200861}}.

\bibitem{hao2023pkamnet}
S.~Hao, B.~An, X.~Ma, X.~Sun, T.~He, S.~Sun, Pkamnet: a transmission line
  insulator parallel-gap fault detection network based on prior knowledge
  transfer and attention mechanism, IEEE Transactions on Power Delivery 38~(5)
  (2023) 3387--3397.

\bibitem{zhang_multi_objects_2023}
S.~Zhang, C.~Qu, C.~Ru, X.~Wang, Z.~Li, Multi-{Objects} {Recognition} and
  {Self}-{Explosion} {Defect} {Detection} {Method} for {Insulators} {Based} on
  {Lightweight} {GhostNet}-{YOLOV4} {Model} {Deployed} {Onboard} {UAV}, IEEE
  Access 11 (2023) 39713--39725.
\newblock \href {https://doi.org/10.1109/ACCESS.2023.3268708}
  {\path{doi:10.1109/ACCESS.2023.3268708}}.

\bibitem{wang2024mci}
Y.~Wang, X.~Song, L.~Feng, Y.~Zhai, Z.~Zhao, S.~Zhang, Q.~Wang, Mci-gla plug-in
  suitable for yolo series models for transmission line insulator defect
  detection, IEEE Transactions on Instrumentation and Measurement (2024).

\bibitem{yi_intelligent_2022}
Y.~Yi, Z.~Chen, L.~Wang, Intelligent {Aging} {Diagnosis} of {Conductor} in
  {Smart} {Grid} {Using} {Label}-{Distribution} {Deep} {Convolutional} {Neural}
  {Networks}, IEEE Transactions on Instrumentation and Measurement 71 (2022)
  1--8.
\newblock \href {https://doi.org/10.1109/TIM.2022.3141160}
  {\path{doi:10.1109/TIM.2022.3141160}}.

\bibitem{zhao_detection_2020}
Z.~Zhao, H.~Qi, Y.~Qi, K.~Zhang, Y.~Zhai, W.~Zhao, Detection {Method} {Based}
  on {Automatic} {Visual} {Shape} {Clustering} for {Pin}-{Missing} {Defect} in
  {Transmission} {Lines}, IEEE Transactions on Instrumentation and Measurement
  69~(9) (2020) 6080--6091.
\newblock \href {https://doi.org/10.1109/TIM.2020.2969057}
  {\path{doi:10.1109/TIM.2020.2969057}}.

\bibitem{xiao_detection_2021}
Y.~Xiao, Z.~Li, D.~Zhang, L.~Teng, Detection of {Pin} {Defects} in {Aerial}
  {Images} {Based} on {Cascaded} {Convolutional} {Neural} {Network}, IEEE
  Access 9 (2021) 73071--73082.
\newblock \href {https://doi.org/10.1109/ACCESS.2021.3079172}
  {\path{doi:10.1109/ACCESS.2021.3079172}}.

\bibitem{zhao_new_2022}
Z.~Zhao, R.~Wang, Y.~Li, Y.~Zhai, W.~Zhao, K.~Zhang, A {New} {Multilabel}
  {Recognition} {Framework} for {Transmission} {Lines} {Bolt} {Defects} {Based}
  on the {Combination} of {Semantic} {Knowledge} and {Structural} {Knowledge},
  IEEE Transactions on Instrumentation and Measurement 71 (2022) 1--11.
\newblock \href {https://doi.org/10.1109/TIM.2022.3200103}
  {\path{doi:10.1109/TIM.2022.3200103}}.

\bibitem{jiao2023defective}
R.~Jiao, Z.~Fu, Y.~Liu, Y.~Zhang, Y.~Song, A defective bolt detection model
  with attention-based roi fusion and cascaded classification network, IEEE
  Transactions on Instrumentation and Measurement (2023).

\bibitem{song_deformable_2023}
Z.~Song, X.~Huang, C.~Ji, Y.~Zhang, Deformable {YOLOX}: {Detection} and {Rust}
  {Warning} {Method} of {Transmission} {Line} {Connection} {Fittings} {Based}
  on {Image} {Processing} {Technology}, IEEE Transactions on Instrumentation
  and Measurement 72 (2023) 1--21.
\newblock \href {https://doi.org/10.1109/TIM.2023.3238742}
  {\path{doi:10.1109/TIM.2023.3238742}}.

\bibitem{zhang2023dsa}
Y.~Zhang, B.~Li, J.~Shang, X.~Huang, P.~Zhai, C.~Geng, Dsa-net: An
  attention-guided network for real-time defect detection of transmission line
  dampers applied to uav inspections, IEEE Transactions on Instrumentation and
  Measurement (2023).

\bibitem{cortes1995support}
C.~Cortes, V.~Vapnik, Support-vector networks, Machine learning 20 (1995)
  273--297.

\bibitem{liang_detection_2020}
H.~Liang, C.~Zuo, W.~Wei, Detection and {Evaluation} {Method} of {Transmission}
  {Line} {Defects} {Based} on {Deep} {Learning}, IEEE Access 8 (2020)
  38448--38458.
\newblock \href {https://doi.org/10.1109/ACCESS.2020.2974798}
  {\path{doi:10.1109/ACCESS.2020.2974798}}.

\bibitem{liu2022component}
X.~Liu, X.~Miao, H.~Jiang, J.~Chen, M.~Wu, Z.~Chen, Component detection for
  power line inspection using a graph-based relation guiding network, IEEE
  Transactions on Industrial Informatics 19~(9) (2022) 9280--9290.

\bibitem{liu2023fault}
X.~Liu, X.~Miao, H.~Jiang, J.~Chen, Z.~Chen, Fault diagnosis in power line
  inspection using normalized multihierarchy embedding matching, IEEE
  Transactions on Instrumentation and Measurement 72 (2023) 1--10.

\bibitem{yi2023pstl}
J.~Yi, J.~Mao, H.~Zhang, K.~Zeng, Z.~Tao, H.~Zhong, S.~Wang, Y.~Wang, Pstl-net:
  A patchwise self-texture-learning network for transmission line inspection,
  IEEE Transactions on Instrumentation and Measurement (2023).

\bibitem{zhong2024visual}
L.~Zhong, K.~Liu, Visual classification and detection of power inspection
  images based on federated learning, IEEE Transactions on Industry
  Applications (2024).

\bibitem{zhu_deep_2020}
J.~Zhu, Y.~Guo, F.~Yue, H.~Yuan, A.~Yang, X.~Wang, M.~Rong, A {Deep} {Learning}
  {Method} to {Detect} {Foreign} {Objects} for {Inspecting} {Power}
  {Transmission} {Lines}, IEEE Access 8 (2020) 94065--94075.
\newblock \href {https://doi.org/10.1109/ACCESS.2020.2995608}
  {\path{doi:10.1109/ACCESS.2020.2995608}}.

\bibitem{bi_yolox_2023}
Z.~Bi, L.~Jing, C.~Sun, M.~Shan, {YOLOX}++ for {Transmission} {Line} {Abnormal}
  {Target} {Detection}, IEEE Access 11 (2023) 38157--38167.
\newblock \href {https://doi.org/10.1109/ACCESS.2023.3268106}
  {\path{doi:10.1109/ACCESS.2023.3268106}}.

\bibitem{yu_foreign_2023}
C.~Yu, Y.~Liu, W.~Zhang, X.~Zhang, Y.~Zhang, X.~Jiang, Foreign {Objects}
  {Identification} of {Transmission} {Line} {Based} on {Improved} {YOLOv7},
  IEEE Access 11 (2023) 51997--52008.
\newblock \href {https://doi.org/10.1109/ACCESS.2023.3277954}
  {\path{doi:10.1109/ACCESS.2023.3277954}}.

\bibitem{zhang_edge_2023}
J.~Zhang, J.~Wang, R.~Song, G.~Peng, T.~Pu, S.~Zhang, An {Edge} {Visual}
  {Incremental} {Perception} {Framework} {Based} on {Deep} {Semi}-supervised
  {Learning} for {Monitoring} {Power} {Transmission} {Lines}, CSEE Journal of
  Power and Energy Systems 9~(2) (2023) 759--768.
\newblock \href {https://doi.org/10.17775/CSEEJPES.2022.03120}
  {\path{doi:10.17775/CSEEJPES.2022.03120}}.

\bibitem{lin2017feature}
T.-Y. Lin, P.~Doll{\'a}r, R.~Girshick, K.~He, B.~Hariharan, S.~Belongie,
  Feature pyramid networks for object detection, in: Proceedings of the IEEE
  conference on computer vision and pattern recognition, 2017, pp. 2117--2125.

\bibitem{ge2021yolox}
Z.~Ge, S.~Liu, F.~Wang, Z.~Li, J.~Sun, Yolox: Exceeding yolo series in 2021,
  arXiv preprint arXiv:2107.08430 (2021).

\bibitem{everingham2010pascal}
M.~Everingham, L.~Van~Gool, C.~K. Williams, J.~Winn, A.~Zisserman, The pascal
  visual object classes (voc) challenge, International journal of computer
  vision 88 (2010) 303--338.

\bibitem{wang2020convergence}
X.~Wang, Y.~Han, V.~C. Leung, D.~Niyato, X.~Yan, X.~Chen, Convergence of edge
  computing and deep learning: A comprehensive survey, IEEE Communications
  Surveys \& Tutorials 22~(2) (2020) 869--904.

\bibitem{shi2016edge}
W.~Shi, J.~Cao, Q.~Zhang, Y.~Li, L.~Xu, Edge computing: Vision and challenges,
  IEEE internet of things journal 3~(5) (2016) 637--646.

\bibitem{mcmahan2017communication}
B.~McMahan, E.~Moore, D.~Ramage, S.~Hampson, B.~A. y~Arcas,
  Communication-efficient learning of deep networks from decentralized data,
  in: Artificial intelligence and statistics, PMLR, 2017, pp. 1273--1282.

\bibitem{karim2023current}
S.~Karim, G.~Tong, J.~Li, A.~Qadir, U.~Farooq, Y.~Yu, Current advances and
  future perspectives of image fusion: A comprehensive review, Information
  Fusion 90 (2023) 185--217.

\bibitem{meher2019survey}
B.~Meher, S.~Agrawal, R.~Panda, A.~Abraham, A survey on region based image
  fusion methods, Information Fusion 48 (2019) 119--132.

\bibitem{shen2017research}
Z.~Shen-pei, L.~Xi, Q.~Bing-chen, H.~Hui, Research on insulator fault diagnosis
  and remote monitoring system based on infrared images, Procedia Computer
  Science 109 (2017) 1194--1199.

\bibitem{ma2019fusiongan}
J.~Ma, W.~Yu, P.~Liang, C.~Li, J.~Jiang, Fusiongan: A generative adversarial
  network for infrared and visible image fusion, Information fusion 48 (2019)
  11--26.

\bibitem{han2020electrical}
S.~Han, F.~Yang, G.~Yang, B.~Gao, N.~Zhang, D.~Wang, Electrical equipment
  identification in infrared images based on roi-selected cnn method, Electric
  Power Systems Research 188 (2020) 106534.

\bibitem{song2020analysis}
C.~Song, W.~Xu, Z.~Wang, S.~Yu, P.~Zeng, Z.~Ju, Analysis on the impact of data
  augmentation on target recognition for uav-based transmission line
  inspection, Complexity 2020 (2020) 1--11.

\bibitem{goodfellow2020generative}
I.~Goodfellow, J.~Pouget-Abadie, M.~Mirza, B.~Xu, D.~Warde-Farley, S.~Ozair,
  A.~Courville, Y.~Bengio, Generative adversarial networks, Communications of
  the ACM 63~(11) (2020) 139--144.

\bibitem{YE2024100250}
R.~Ye, A.~Boukerche, X.-S. Yu, C.~Zhang, B.~Yan, X.-J. Zhou,
  \href{https://www.sciencedirect.com/science/article/pii/S1674862X24000181}{A
  data augmentation method for insulators based on cycle gan}, Journal of
  Electronic Science and Technology (2024) 100250\href
  {https://doi.org/https://doi.org/10.1016/j.jnlest.2024.100250}
  {\path{doi:https://doi.org/10.1016/j.jnlest.2024.100250}}.
\newline\urlprefix\url{https://www.sciencedirect.com/science/article/pii/S1674862X24000181}

\bibitem{zhu2017unpaired}
J.-Y. Zhu, T.~Park, P.~Isola, A.~A. Efros, Unpaired image-to-image translation
  using cycle-consistent adversarial networks, in: Proceedings of the IEEE
  international conference on computer vision, 2017, pp. 2223--2232.

\bibitem{raina2007self}
R.~Raina, A.~Battle, H.~Lee, B.~Packer, A.~Y. Ng, Self-taught learning:
  transfer learning from unlabeled data, in: Proceedings of the 24th
  international conference on Machine learning, 2007, pp. 759--766.

\bibitem{wang2020generalizing}
Y.~Wang, Q.~Yao, J.~T. Kwok, L.~M. Ni, Generalizing from a few examples: A
  survey on few-shot learning, ACM computing surveys (csur) 53~(3) (2020)
  1--34.

\bibitem{finn2017model}
C.~Finn, P.~Abbeel, S.~Levine, Model-agnostic meta-learning for fast adaptation
  of deep networks, in: International conference on machine learning, PMLR,
  2017, pp. 1126--1135.

\bibitem{dwyer2024roboflow}
B.~Dwyer, J.~Nelson, T.~Hansen, et~al., \href{https://roboflow.com}{Roboflow}
  (2024).
\newline\urlprefix\url{https://roboflow.com}

\bibitem{HumanSignal}
HumanSignal,
  \href{https://github.com/HumanSignal/awesome-data-labeling}{Humansignal/awesome-data-labeling:
  A curated list of awesome data labeling tools} (2024).
\newline\urlprefix\url{https://github.com/HumanSignal/awesome-data-labeling}

\bibitem{labelstudio}
M.~Tkachenko, M.~Malyuk, A.~Holmanyuk, N.~Liubimov,
  \href{https://github.com/heartexlabs/label-studio}{{Label Studio}: Data
  labeling software}, open source software available from
  https://github.com/heartexlabs/label-studio (2020).
\newline\urlprefix\url{https://github.com/heartexlabs/label-studio}

\bibitem{kirillov2023segment}
A.~Kirillov, E.~Mintun, N.~Ravi, H.~Mao, C.~Rolland, L.~Gustafson, T.~Xiao,
  S.~Whitehead, A.~C. Berg, W.-Y. Lo, et~al., Segment anything, in: Proceedings
  of the IEEE/CVF International Conference on Computer Vision, 2023, pp.
  4015--4026.

\bibitem{zhou2018brief}
Z.-H. Zhou, A brief introduction to weakly supervised learning, National
  science review 5~(1) (2018) 44--53.

\bibitem{choi2021weakly}
H.~Choi, G.~Koo, B.~J. Kim, S.~W. Kim, Weakly supervised power line detection
  algorithm using a recursive noisy label update with refined broken line
  segments, Expert Systems with Applications 165 (2021) 113895.

\bibitem{ledig2017photo}
C.~Ledig, L.~Theis, F.~Husz{\'a}r, J.~Caballero, A.~Cunningham, A.~Acosta,
  A.~Aitken, A.~Tejani, J.~Totz, Z.~Wang, et~al., Photo-realistic single image
  super-resolution using a generative adversarial network, in: Proceedings of
  the IEEE conference on computer vision and pattern recognition, 2017, pp.
  4681--4690.

\bibitem{hu2018small}
G.~X. Hu, Z.~Yang, L.~Hu, L.~Huang, J.~M. Han, et~al., Small object detection
  with multiscale features, International Journal of Digital Multimedia
  Broadcasting 2018 (2018).

\bibitem{defard2021padim}
T.~Defard, A.~Setkov, A.~Loesch, R.~Audigier, Padim: a patch distribution
  modeling framework for anomaly detection and localization, in: International
  Conference on Pattern Recognition, Springer, 2021, pp. 475--489.

\bibitem{batzner2024efficientad}
K.~Batzner, L.~Heckler, R.~K{\"o}nig, Efficientad: Accurate visual anomaly
  detection at millisecond-level latencies, in: Proceedings of the IEEE/CVF
  Winter Conference on Applications of Computer Vision, 2024, pp. 128--138.

\bibitem{tsai2021autoencoder}
D.-M. Tsai, P.-H. Jen, Autoencoder-based anomaly detection for surface defect
  inspection, Advanced Engineering Informatics 48 (2021) 101272.

\bibitem{sun2023semisupervised}
S.~Sun, Y.~Liu, X.~Hu, W.~Zhang, A semisupervised autoencoder-based method for
  anomaly detection in cutting tools, Journal of Manufacturing Processes 93
  (2023) 315--327.

\bibitem{roth2022towards}
K.~Roth, L.~Pemula, J.~Zepeda, B.~Sch{\"o}lkopf, T.~Brox, P.~Gehler, Towards
  total recall in industrial anomaly detection, in: Proceedings of the IEEE/CVF
  conference on computer vision and pattern recognition, 2022, pp.
  14318--14328.

\bibitem{yang2023memseg}
M.~Yang, P.~Wu, H.~Feng, Memseg: A semi-supervised method for image surface
  defect detection using differences and commonalities, Engineering
  Applications of Artificial Intelligence 119 (2023) 105835.

\bibitem{zavrtanik2021draem}
V.~Zavrtanik, M.~Kristan, D.~Sko{\v{c}}aj, Draem-a discriminatively trained
  reconstruction embedding for surface anomaly detection, in: Proceedings of
  the IEEE/CVF international conference on computer vision, 2021, pp.
  8330--8339.

\bibitem{ultralyticsFrequentlyAsked}
Ultralytics, {F}requently {A}sked {Q}uestions ({F}{A}{Q}) ---
  docs.ultralytics.com, \url{https://docs.ultralytics.com/help/FAQ/}, [Accessed
  18-02-2024] (2024).

\bibitem{Team}
K.~Team, \href{https://keras.io/api/applications/}{Keras documentation: Keras
  applications} (2024).
\newline\urlprefix\url{https://keras.io/api/applications/}

\bibitem{chollet2017xception}
F.~Chollet, Xception: Deep learning with depthwise separable convolutions, in:
  Proceedings of the IEEE conference on computer vision and pattern
  recognition, 2017, pp. 1251--1258.

\end{thebibliography}
